\documentclass{article}

\usepackage{microtype}
\usepackage{graphicx}
\usepackage{subfig}
\usepackage[dvipsnames, table]{xcolor}
\usepackage{colortbl}
\usepackage{booktabs} 
\definecolor{LightCyan}{rgb}{0.93,1,1}
\definecolor{LightRed}{rgb}{1,0.93,1}
\definecolor{LightGreen}{rgb}{1, 1, 0.88}
\definecolor{Gray}{gray}{0.9}
\newcolumntype{h}{>{\columncolor{LightCyan}}c}
\newcolumntype{a}{>{\columncolor{LightRed}}c}
\newcolumntype{d}{>{\columncolor{LightGreen}}c}
\newcolumntype{g}{>{\columncolor{Gray}}c}

\usepackage{hyperref}

\usepackage[accepted]{icml2023}

\usepackage{amsmath, bm}

\DeclareMathOperator*{\argmin}{arg\,min}
\usepackage{amssymb}
\usepackage{mathtools}
\usepackage{amsthm}
\usepackage{algorithm}
\usepackage{hhline}
\usepackage{multirow, makecell}
\usepackage{xcolor}
\newcommand{\bftab}{\fontseries{b}\selectfont}
\usepackage[capitalize,noabbrev]{cleveref}
\theoremstyle{plain}
\newtheorem{theorem}{Theorem}[section]
\newtheorem{proposition}[theorem]{Proposition}
\newtheorem{lemma}[theorem]{Lemma}
\newtheorem{corollary}[theorem]{Corollary}
\theoremstyle{definition}
\newtheorem{definition}[theorem]{Definition}
\newtheorem{assumption}[theorem]{Assumption}
\newtheorem{remark}[theorem]{Remark}

 \usepackage[textsize=tiny]{todonotes}

\newcommand{\Cosim}{\textsc{cosim}}
\newcommand{\w}{\bm w}
\newcommand{\nw}{\nabla \bm w}
\newcommand{\hw}{\widehat{\bm w}}
\newcommand{\e}{\bm\epsilon}
\newcommand{\maE}{\mathbb E}
\icmltitlerunning{Surrogate Model Extension (SME): A Fast and Accurate Weight Update Attack on Federated Learning}

\begin{document}

\twocolumn[
\icmltitle{Surrogate Model Extension (SME):\\ A Fast and Accurate Weight Update Attack on Federated Learning}




\begin{icmlauthorlist}
\icmlauthor{Junyi Zhu}{yyy}
\icmlauthor{Ruicong Yao}{comp}
\icmlauthor{Matthew B. Blaschko}{yyy}
\end{icmlauthorlist}

\icmlaffiliation{yyy}{ESAT-PSI, KU Leuven, Belgium}
\icmlaffiliation{comp}{Statistics and Data Science, Dept. Mathematics, KU Leuven, Belgium}

\icmlcorrespondingauthor{Junyi Zhu}{Junyi.Zhu@esat.kuleuven.be}

\icmlkeywords{Machine Learning, ICML}

\vskip 0.3in
]



\printAffiliationsAndNotice{}  

\begin{abstract}
In Federated Learning (FL) and many other distributed training frameworks, collaborators can hold their private data locally and only share the network weights trained with the local data after multiple iterations. Gradient inversion is a family of privacy attacks that recovers data from its generated gradients. Seemingly, FL can provide a degree of protection against gradient inversion attacks on weight updates, since the gradient of a single step is concealed by the accumulation of gradients over multiple local iterations. In this work, we propose a principled way to extend gradient inversion attacks to weight updates in FL, thereby better exposing weaknesses in the presumed privacy protection inherent in FL. In particular, we propose a surrogate model method based on the characteristic of two-dimensional gradient flow and low-rank property of local updates.
Our method largely boosts the ability of gradient inversion attacks on weight updates containing many iterations and achieves state-of-the-art (SOTA) performance. 
Additionally, our method runs up to $100\times$ faster than the SOTA baseline in the common FL scenario. Our work re-evaluates and highlights the privacy risk of sharing network weights. Our code is available at~\url{https://github.com/JunyiZhu-AI/surrogate_model_extension}.
\end{abstract}

\section{Introduction}
\begin{figure*}[t]
    \centering
\includegraphics[width=0.95\textwidth]{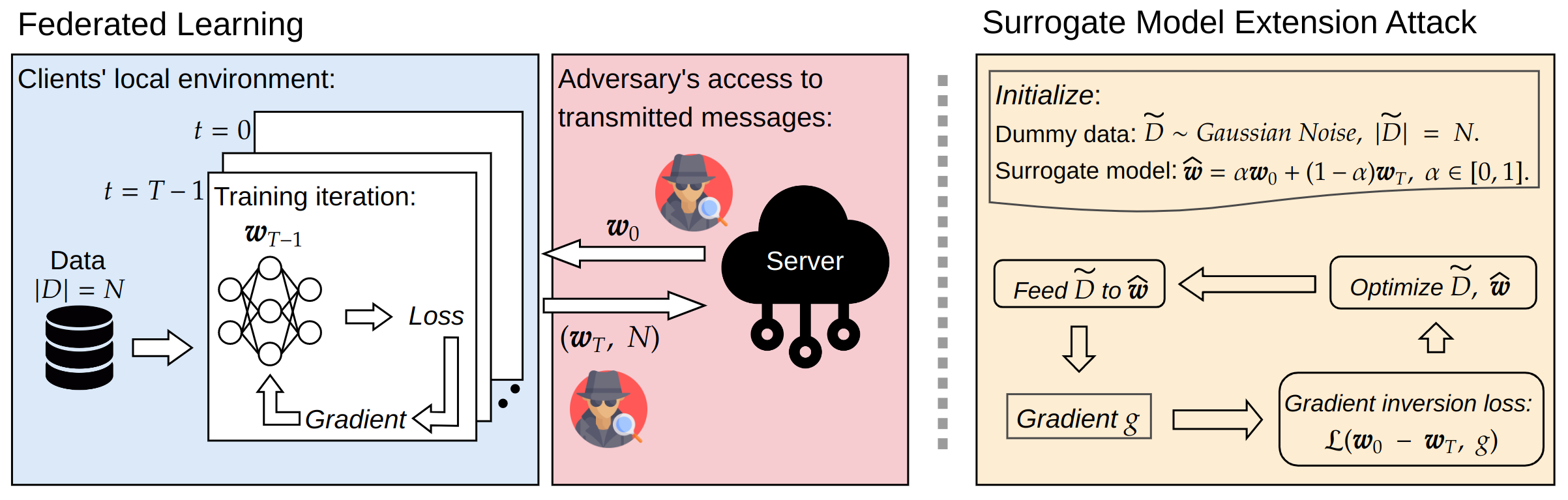}
    \caption{Illustration of the threat model (left) and working pipeline of our surrogate model extension (right). In FL, a client trains the received model $\bm w_0$ for $T$ iterations with local data set $D$ of size $N$, then sends the weights and the number $N$ back to the server. An adversary observes the messages and launches the SME attack through optimization of dummy data $\Tilde{D}$ and surrogate model $\widehat{\bm w}$.}
    \label{fig:illustration}
\end{figure*}
Privacy concerns arise from many areas. Therefore, data-driven technologies, e.g.\ machine learning, cannot solely rely on a large data center, but also adapt to distributed data scenarios. Federated Learning (FL) is proposed to this end~\cite{mcmahan17a_fedavg, kairouz2021}, which is a framework for training a joint model built upon server-client 
communication. A presumed strength of FL from the perspective of privacy is that the network weights instead of data are shared between the server and clients. 
However, training data may still be extracted from the transmitted weights due to their statistical dependence.

Gradient inversion presents a way to reconstruct input data from the gradient~\cite{dlg_zhu, geiping_inverting_gradient}. The reconstruction frequently has high fidelity, thus raising alarms about the privacy risk of sharing network weights. However, such methods assume that weights are shared at every iteration, thus the gradient of a single update can be constructed from server-client communications. In practice, communication overhead is a major bottleneck of collaborative training, especially for a wide range of applications of FL on edge devices. Therefore, FL frameworks mostly require clients to train multiple epochs before sharing their updated weights. The gradient of a single step is thus concealed by the accumulation of local iterations.
An existing attack on weight updates in FL is the simulation method~\cite{dimitrov2022data}, which emulates multiple steps of gradient descent. However, such methods are computationally demanding and may not be possible to apply in practice, providing the illusion of data protection.

In this work, we propose an extension of the gradient inversion method, which attacks weight updates of accumulated local iterations in a principled way. In particular, we introduce a surrogate model, which is a model not necessarily appearing during training. We take the surrogate model as a basis and regard the reversed direction of the weight update as the direction of the gradient of training data computed on this surrogate model. We propose that an effective surrogate model can be found as a linear combination of the weights before and after local training, and identify that this is feasible due to the characteristic of two-dimensional gradient flow and the low-rank property of local steps. Equipped with this Surrogate Model Extension (\verb|SME|), we can then invert the weight update and reconstruct training data using any gradient inversion method. 
\Cref{fig:illustration} illustrates the threat model and working pipeline of our method \verb|SME|.

\textbf{Our contributions are summarized as}: (i) We propose a method that extends gradient inversion to weight updates computed from multiple iterations over the data. (ii) We also provide an analysis for theoretical understanding of the proposed method. (iii) We demonstrate that our method largely strengthens the ability of gradient inversion in attacking weight updates at a negligible additional cost. (iv) Our method achieves SOTA performance in reconstruction and requires substantially fewer computational resources than the current SOTA approach.

\textbf{Paper Organization.} In \Cref{sec:related_works} we discuss the related works. In \Cref{sec:preliminaries} we provide methodological foundations of FL and inversion attacks. In \Cref{sec:theory}, we elaborate on our method with theoretical analysis underpinning its generality. In \Cref{sec:experiments} we present the experimental results. Finally, we conclude in \Cref{sec:conclusion}.
\section{Related Works}
\label{sec:related_works}
The first observation of data leakage from gradients appears to be due to~\cite{phong2017gap}, where they find that the gradient w.r.t.\ weights divided by the gradient w.r.t.\ bias can be used to reconstruct the input of a fully connected layer. Subsequently, \citet{dlg_zhu} propose the first general gradient inversion method to reconstruct the input from the gradient over an arbitrary network architecture. A line of follow-up work intends to improve the reconstruction quality by further constraining the reconstruction through some auxiliary information of the input \cite{geiping_inverting_gradient, yin_cvpr_see_through_gradient, april_lu_cvpr, jeon2021gradient}. Additionally, \citet{zhu2021rgap, fan2020rethinking} propose analytic methods for gradient inversion. \citet{balunovic2022bayesian} propose a Bayesian framework. 

Many works discuss the threat of gradient inversion to FL~\cite{Wei2020AFF, geng_2022_aaai, GradDisagg, Jin_cafe}. Since in FL and many other distributed training schemes, weight updates are in form of multiple local steps instead of a single step, \citet{geiping_inverting_gradient, dimitrov2022data} propose simulation methods which fit the weight updates through mimicking the local training iterations.
We provide a broader discussion of other types of privacy attacks and defense strategies in \Cref{sec:app:related_works}.
\section{Preliminaries}
\label{sec:preliminaries}
This section presents the necessary background of this work. In \Cref{sec:DLG}, we review the mechanism of gradient inversion attacks. In \Cref{sec:FL} we introduce a general training protocol of the FL framework, which we will use as the attack scenario in this work. In \Cref{sec:DLFA} we clarify the simulation method that is specifically designed for attacking weight updates of many iterations.

\subsection{Deep Leakage from Gradients}
\label{sec:DLG}
Data used to train the network can be extracted from the generated gradient. \citet{dlg_zhu} propose the first general gradient inversion method called Deep Leakage from Gradient (\verb|DLG|), which recovers the original data by searching for the data that generates a gradient matching the original one. Denote by $\ell(\cdot)$ the loss function of the network training, and denote by $\bm w$ the weights of the network, $D=\{(x_i, y_i)\}_{i=1}^N$ is the data used to generate the gradient, which consists of $N$ pairs of inputs $x$ and labels $y$. \verb|DLG| optimizes a randomly initialized dummy data $\Tilde{D}$ w.r.t.\ the Euclidean distance between the original gradient and dummy gradient. Thus, the reconstruction loss used in \verb|DLG| can be defined as: $\| \nabla_{\bm w}\ell(\bm w, D) -\nabla_{\bm w}\ell(\bm w, \Tilde{D}) \|$.

Based on \verb|DLG|, another work Inverting Gradient (\verb|IG|) 
proposes to replace the Euclidean distance with cosine similarity loss and incorporate the total variation (TV) of the pixel values as a prior for image reconstruction ~\cite{geiping_inverting_gradient}. The reconstruction loss $\mathcal{L}$ of \verb|IG| can thus be described as:
\begin{equation}
    \label{eq:IG}
    \mathcal{L} = \overbrace{1-\frac{\langle\nabla_{\bm w}\ell(\bm w, D),\nabla_{\bm w}\ell(\bm w, \Tilde{D})\rangle}{\|\nabla_{\bm w}\ell(\bm w, D)\|\|\nabla_{\bm w}\ell(\bm w, \Tilde{D})\|}}^{\mathcal{L}_{sim}(\nabla_{\bm w}\ell(\bm w, D),\nabla_{\bm w}\ell(\bm w, \Tilde{D}))} + \lambda\underbrace{ \operatorname{TV}(\Tilde{D})}_{\mathcal{L}_{prior}},
\end{equation}
where $\lambda$ is a hyperparameter tuning the strength of the prior. Later, we will use $\mathcal{L}_{sim}$ to denote the cosine similarity loss.

\verb|IG| has been proven to be more stable and efficient than \verb|DLG| in practice, and its objective is relatively simple and general.
In this work, we will use the image reconstruction attack as a proof of concept and rely on the reconstruction loss defined in \Cref{eq:IG}. Our method \verb|SME| focuses on adapting the core term in \Cref{eq:IG}, i.e.\ the gradient similarity loss $\mathcal{L}_{sim}$, to the weight update of many iterations.

\subsection{Communication in Federated Learning}
\label{sec:FL}
In this work we consider a classical FL framework: Federated Averaging (\verb|FedAvg|)~\cite{mcmahan17a_fedavg}. The procedure of \verb|FedAvg| is given in \Cref{alg:fedavg} in \Cref{sec:app:experiment}. For each communication round in \verb|FedAvg|, clients train the local model for $E$ epochs before sending back the optimized weights. While for each epoch, the network will be updated for $\lceil N/B \rceil$ steps, where $B$ is the batch size and $N$ is the local data size. Generally, $B$ is fixed by the training protocol and $N$ depends on how much data a client can collect. 

Therefore, for one round of communication, the network weights that could be intercepted by an adversary at the communication channel side, or observed by the server, are $\bm w_0$ and $\bm w_T$. The weight update $\bm w_T - \bm w_0$ accumulates $T = E\lceil N/B \rceil$ local steps of gradient descent, thus the gradient of every single step is obfuscated.

Weight updates of multiple steps can still be plugged into the reconstruction loss of gradient inversion \cite{Wei2020AFF, geng_2022_aaai}. Such technologies usually normalize the reversed weight update of $T$ steps, i.e.\ $(\bm w_0 - \bm w_T)/\eta T$, where $\eta$ is the learning rate. Then they take the normalized gradient as an approximation of the gradient of the local data set $D$ computed on $\bm w_0$, i.e.\ $\nabla_{\bm w_0} \ell(\bm w_0, D)$, and apply gradient inversion to reconstruct the local data set $D$. However, as the number of local steps $T$ increases, the normalized gradient incrementally mismatches the gradient at $\bm w_0$, which induces intrinsic objective error defined as the cosine similarity loss with the optimal reconstruction $\Tilde{D} = D$:
\begin{equation}
\label{eq:error}
    \delta_{error} := \mathcal{L}_{sim}(\bm w_0 - \bm w_T, \nabla_{\bm w_0} \ell(\bm w_0, D)),
\end{equation}
since $1-\mathcal{L}_{sim}$ is a cosine similarity, it is invariant to the rescaling $1/\eta T$.
As a result, when the direction of weight updates deviates from $\nabla_{\bm w_0} \ell(\bm w_0, D)$, this error increases. It has been observed that the reconstruction quality of gradient inversion degrades with the number of local steps $T$~\cite{ Wei2020AFF, dimitrov2022data}.

\subsection{Inverting Weight Updates through Simulation}
\label{sec:DLFA}
An existing privacy attack targeting weight updates is the simulation method, for which Data Leakage from Federated Averaging (\verb|DLFA|) achieves SOTA performance in terms of the reconstruction quality \cite{dimitrov2022data}. Simulation methods reconstruct data by fitting internal states of the local training. That is, the adversary mimics the local training steps by optimizing $\bm w_0$ with dummy data $\Tilde{D}$ and obtains dummy weights $\{\tilde{\bm w}_t\}_{t=1}^T$. Then the adversary minimizes the difference between the weight updates, i.e. $\Tilde{\bm w}_T - \bm w_0 $ and $\bm w_T - \bm w_0$. The simulation loss $\mathcal{L}$ can thus be defined as:
\begin{align}
    \nonumber
    \mathcal{L} &= \|\tilde{\bm w}_T - \bm w_0 - (\bm w_T - \bm w_0)\|\\
    \nonumber
    &= \|\eta(\nabla \bm w_0 + \nabla \Tilde{\bm w}_1 + \dots + \nabla \Tilde{\bm w}_{T-1}  ) - \bm w_T + \bm w_0\|,
\end{align}
where we define $\nabla \bm w = \nabla_{\bm w}\ell(\bm w, \tilde{D})$. The Euclidean distance can also be replaced by a cosine similarity loss.

Simulation methods circumvent the intrinsic objective error of the gradient inversion. However, a major issue of the simulation methods is their computational complexity. Consider a simulation for only two local steps, the reconstruction gradient w.r.t.\ the dummy data can be derived as:\footnote{In the original work \cite{dimitrov2022data}, the input could be batches of the dummy data. Here we consider the full $\Tilde{D}$ as input for clarity. This does not change our discussion.}
\begin{align}
    \nonumber
    \nabla_{\Tilde{D}} \mathcal{L} = &\frac{\partial \mathcal{L}}{\partial \nabla \Tilde{\bm w_1}}\left(\frac{\partial \nabla_{\bm w}\ell(\bm w, \Tilde{D})}{\partial \Tilde{D}}|_{\bm w = \Tilde{\bm w_1}} - \eta\frac{\partial \nabla \Tilde{\bm w_1}}{\partial \Tilde{\bm w}_1}\frac{\partial \nabla \bm w_0}{\partial \Tilde{D}}\right)\\
    \label{eq:SIM_grad}
    &+\frac{\partial \mathcal{L}}{\partial \nabla \bm w_0}\frac{\partial \nabla \bm w_0}{\partial \Tilde{D}}.
\end{align}
A full derivation of $\nabla_{\Tilde{D}}\mathcal{L}$ for $T$ local steps is deferred to \Cref{sec:app:sim_grad}. According to \Cref{eq:SIM_grad}, we can see that the reconstruction gradient $\nabla_{\Tilde{D}} \mathcal{L}$ of simulation methods has a complicated form, which involves the computation of multiple second-order derivatives and their products. When there are many local steps, a long chain of forward and backward propagation will be induced, which leads to very large computational overhead and memory footprint.
\section{Inverting Weight Updates through a Surrogate Model}\label{sec:theory}
In this section, we elaborate on our surrogate model extension \verb|SME|. Denote $\widehat{\bm w}$ as the surrogate model, we give the general objective of surrogate model extension attack:
\begin{equation}
    \label{eq:general_obj}
    \argmin_{\Tilde{D}\in\mathcal{D}} \min_{\widehat{\bm w} \in \mathcal{W}} \mathcal{L}_{sim}(\bm w_0-\bm w_T, \nabla_{\widehat{\bm w}}\ell(\widehat{\bm w}, \Tilde{D})).
\end{equation}
When we fix $\widehat{\bm w}$ to $\bm w_0$, this objective reverts to the vanilla gradient inversion and contains an intrinsic objective error as shown in \Cref{eq:error}.
It is worth noting that $\widehat{\bm w}$ generally has many more parameters than $\Tilde{D}$. If we do not constrain the feasible set $\mathcal{W}$, this objective still cannot give us a good reconstruction. 

Our key insight is that the linear combinations of $\bm w_0$ and $\bm w_T$ is a good choice of $\mathcal{W}$, i.e.\ $\mathcal{W} = \{\alpha\bm w_0 + (1-\alpha)\bm w_T\mid\alpha\in[0,1]\}$.
We identify that if the parameter space is two-dimensional (2D), then there exists a $\widehat{\bm w}$ on the connected line between $\bm w_0$ and $\bm w_T$, 
such that $\nabla_{\widehat{\bm w}}\ell(\widehat{\bm w}, D)$ is parallel to $\bm w_0 - \bm w_T$, denoted as $\nabla_{\widehat{\bm w}}\ell(\widehat{\bm w}, D) \parallel \bm w_0 - \bm w_T$. Thus a minimizer $\widehat{\bm w}$ corresponding to the original data $D$ can be found on the connected line, which leads to $\mathcal{L}_{sim}(\bm w_0-\bm w_T, \nabla_{\widehat{\bm w}}\ell(\widehat{\bm w}, D)) = 0$. To keep the notation uncluttered, we will slightly abuse $\nabla \w$ and denote $\nabla \w := \nabla \ell(\w, D)$.

For a general high-dimensional parameter space, we further observe that the local update steps of a network basically transit within a low-dimensional subspace due to the low-rank property of the gradients. Therefore, for a high-dimensional network, it is still possible to find a surrogate model $\widehat{\bm w}$ as a linear combination of $\bm w_0$ and $\bm w_T$, such that the direction of $\nabla \widehat{\bm w}$ and $\bm w_0 - \bm w_T$ are close. 
The low-rank property of gradients has also been observed and utilized by a wide range of works~\cite{low_rank_gradient1, low_rank_gradient2, low_rank_gradient3, zhou2021bypassing, li2022when}.
Additionally, as we can reparamterize the surrogate model with only a scalar $\alpha$, simultaneously optimizing the reconstructed data $\Tilde{D}$ and the surrogate model $\widehat{\bm w}$ becomes tractable. After incorporating the image prior TV, we define the reconstruction loss $\mathcal{L}$ of our \verb|SME| as:
\begin{align}
    \nonumber
     \mathcal{L}(\Tilde{D}, \widehat{\bm w}) &=\overbrace{1-\frac{\langle \bm w_0 - \bm w_T,\nabla_{\widehat{\bm w}}\ell(\widehat{\bm w}, \Tilde{D})\rangle}{\|\bm w_0 - \bm w_T\|\|\nabla_{\widehat{\bm w}}\ell(\widehat{\bm w}, \Tilde{D})\|}}^{\mathcal{L}_{sim}(\bm w_0 - \bm w_T, \nabla_{\widehat{\bm w}}\ell(\widehat{\bm w}, \Tilde{D}))} + \lambda\underbrace{\operatorname{TV}(\Tilde{D})}_{\mathcal{L}_{prior}},\\
    s.t.\quad
    \widehat{\bm w} &\in\{\alpha\bm w_0 + (1-\alpha)\bm w_T\mid\alpha\in[0, 1]\}.
    \label{eq:sme}
\end{align}
Next, in \Cref{sec:method:2D} we first prove that for a 2D parameter space, there exists  a minimizer of $\hw$ for the original data on the connected line between the starting point $\bm w_0$ and end point $\bm w_T$ of gradient flow (gradient descent (GD) with infinitesimal step sizes). Then in \Cref{sec:method:general}, we extend our analysis to more general cases. Finally, in \Cref{sec:method:impl} we present the implementation of our surrogate model extension. All proofs are deferred to \Cref{sec:app:proof}.

\subsection{Gradient Flow in a 2-Dimensional Parameter Space}
\label{sec:method:2D}
\begin{figure}[t]
    \centering
\includegraphics[width=0.8\linewidth]{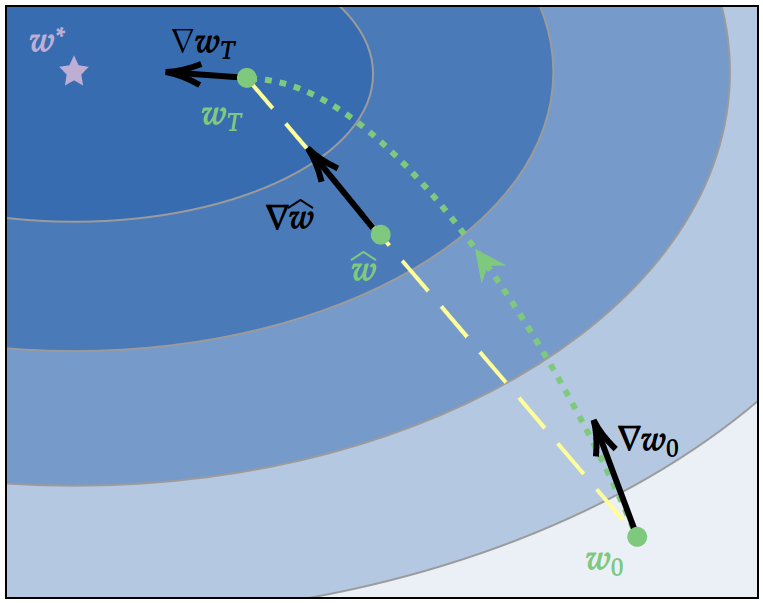}
    \caption{Illustration of gradient flow in a quadratic model. The green dotted line represents the gradient flow, and $\bm w^*$ denotes the global optimum.}
    \label{fig:quadratic_model}
\end{figure}

Figure \ref{fig:quadratic_model} illustrates a gradient flow in a quadratic model and motivates our surrogate model extension. 
The key observation from this figure is that the gradient $\nabla \bm w_0$ points to the right hand side of $\bm w_T-\bm w_0$, while $\nabla \bm w_T$ points to the other side. Since the gradient is continuous, a surrogate model whose gradient is parallel to $\bm w_T-\bm w_0$ must exist.  Although other configurations of the gradient flow exist, we can prove the following proposition holds generally:


\begin{proposition}\label{prop:2dspace}
For a 2D gradient flow $\w_t$ based on the the loss $\ell(\w)\in \mathcal{C}^2$, there exists a surrogate model $\widehat\w$ on the connected line between $\bm w_0$ and $\bm w_T$ satisfying $\langle \nabla\widehat\w,\w_0-\w_T\rangle\in\{\pm1\}$, provided that $\langle\nw_t,\w_0-\w_T\rangle\ge0$, $\forall t\in[0,T]$. If 
we further assume that $\ell(\w)$ is convex, then $\nabla\widehat {\bm w}\parallel \w_0-\w_T$. 
\end{proposition}

\subsection{Cosine Similarity between the High-Dimensional Surrogate Gradient and Weight Update}
\label{sec:method:general}
In the 2D gradient flow case, it has been shown that there exists a value of $\alpha$ that achieves a cosine similarity of exactly 1. This may fail in higher dimensional spaces. Consider a gradient flow in $\mathbb R^3$  parameterized by $(\cos t-1, \sin t, t)$, $t\in \mathbb R$, such that the origin is the starting point and $(0,0,2\pi)$ is the end point. We further assume that the gradient field in $\mathbb R^3$ is parameterized by $(\cos t-1-s, \sin t-r, t)$, $r,s,t\in \mathbb R$ (which is reasonable since the gradient flows can be similar to each other locally). Then we know that the gradient of the weights on the connected line would be $(-\sin t, \cos t, 1)$. This indicates that their cosine similarity with $(0,0,2\pi)$ would always be $1/\sqrt{2}$ which deviates from 1.

However, in the above artificial example, the gradient flow curls in three dimensions. In practice, we observe that there exists a 2D subspace that absorbs a majority of the magnitude of the gradients of local steps. Using this low-rank property of the local steps, we show that there exists a surrogate model $\widehat{\bm w}$ on the connected line, s.t.\ $\mathcal{L}_{sim}(\bm w_0 - \bm w_T, \nabla \widehat\w)\approx0$, where $1-\mathcal{L}_{sim}$ is known as the cosine similarity. Additionally, we consider (stochastic) gradient descent with step size $\eta$.
Our assumption and first result are:
\begin{assumption}\label{ass:1}
     We assume that the true loss $\ell(\w):=\maE_D[\ell(\w,D)]$ is Lipschitz with constant $L$ and $\gamma-$strongly convex, the true gradient $\nabla \bm w:=\maE_D[\nabla_{\w}\ell(\w,D)]$ is Lipschitz with constant $\beta$, and there exists $0<G_2\ll1$ and a projection matrix $P_2$ from $\mathbb R^p$ to some 2D subspace $V_2$ s.t.
     \begin{equation}\label{eq:ass1}
      ||P_2\nabla \bm w||^2\ge (1-G_2^2)||\nabla \bm w||^2.   
     \end{equation}
\end{assumption}

\vspace{-2pt}

In the following, the gradient flow path is denoted as $\w(t),t\in\mathbb R^+$ to be distinguished with the GD path $\{\w_t\}_{t=1}^T$ with step size $\eta$. In addition, we assume that it satisfies: 
     \begin{equation}
         1-\mathcal{L}_{sim}(\nw(t),\w(0)-\w(T\eta))\ge G_2.
     \end{equation}
\begin{theorem}\label{thm:GD}
    Let $\{\w_t\}_{t=0}^{T}$ result from the application of GD in $\mathbb R^{p}$. Under \Cref{ass:1} and that $\eta\beta<1$, $\eta<1$, we have for $G(\eta,\beta) := \sqrt{\frac{G_2^2\eta}{1-\eta\beta/2}}$, there exists a surrogate model $\widehat\w$ on the connected line between $\w_0$ and $\w_T$:
    \begin{equation}\label{eq:thm4.3}
    \small{\mathcal{L}_{sim}(\bm w_0 - \bm w_T, \nabla\hw)\le \left(G_2+\sqrt{{\frac{TG^2(\eta,\beta)L^2}{\ell(\w_0)-\ell(\w_T)}}}\right)^2}
    \end{equation}
    up to a $\Delta_{trun}$ (see \cref{eq:trun}) due to discretization.
\end{theorem}
The proof is based on the identity $\langle\nabla\widehat \w,\bm w_0-\bm w_T \rangle = \langle P_2\nabla\widehat \w,P_2(\bm w_0-\bm w_T) \rangle+\langle P_2^\perp\nabla\widehat \w,P_2^\perp(\bm w_0-\bm w_T) \rangle$ ($P_2^\perp$ is the projection onto $V_2^\perp$) which allows us to estimate the quantity $\mathcal{L}_{sim}$ on two orthogonal subspaces. Here $\Delta_{trun}$ stems from the discretization (gradient flow to gradient descent path). The idea in the proof of \cref{thm:SGD} also adapts to its estimation and we argue that $\Delta_{trun}$ is negligible as long as GD approximates the gradient flow well, see \cref{sec:app:proof:GD}. Thus we omit it in \cref{thm:SGD} for brevity.

\begin{remark}
\Cref{thm:GD} shows that small step size $\eta$ and high concentration of the gradient on $V_2$ (thus $G_2$ is small) result in low $\mathcal{L}_{sim}$. It also indicates that there is a `scaling effect' of $T/(\ell(\bm w_0) - \ell(\bm w_T))$.
This explains why a good surrogate model can be found for long training steps when the drop of loss is adequate, given that $G_2$ is small. 
\end{remark}

For SGD, again we first deal with the 2D case. Since a fully random path does not have the property we want, we assume that the magnitude of the gradient noise is moderate so that we can approximate SGD $\{\w_t\}_{t=0}^T$ by GD $\{\w_t'\}_{t=0}^T$.  We first show that with high probability, the angle $\theta_1$ between $\w_T-\w_0
$ and $\w_T'-\w_0$ is small. 
Then we can find a surrogate model $\widehat \w'$ on the connected line between $\bm w_0$ and $\bm w_T'$ by Proposition \ref{prop:2dspace}. We then try to locate a surrogate model $\widehat \w$ on the connected line between $\w_0$ and $\w_T$ such that the angle $\theta_2$ between $\nabla\widehat \w$ and $\nabla\widehat\w'$ is small. This is done by upper bounding $||\nabla\widehat \w-\nabla\widehat \w'||$ and lower bounding $||\nabla\widehat \w'||$, where we use strong convexity.

For high-dimensional spaces, we use the aforementioned identity and show that $||P_2^\perp(\bm w_T-\bm w_0)||/||\bm w_T-\bm w_0||$ is small. Thus the weight update is mainly in $V_2$, and $\mathcal{L}_{sim}$ would be similar to the 2D case. The final result is summarized in the following theorem: 
\begin{theorem}\label{thm:SGD}
 Let $\{\w_t\}_{t=0}^T$ result from the application of SGD in $\mathbb R^{p}$ and $\{\w_t'\}_{t=0}^T$ result from GD where $\w_0=\w_0'$. Assume the same assumptions as \Cref{thm:GD} and that $C_{GD}:=\sqrt{\frac{TG^2(\eta,\beta)L^2}{\ell(\w_0)-\ell(\w_T)}}<1$. Let $c:=\frac{1}{\sqrt{\eta}(1+\eta\beta)^{T-1}}$, $E_{max}$ be the maximum $L^2$ norm of the gradient noise and define:
\begin{equation}
\label{eq:r}
 \small{r:=\min\left\{\frac{c(\ell(\w_0)-\ell(\w_T'))^2}{4L^2TE^2_{max}}, \frac{c(\ell(\w_0)-\ell(\w_T'))}{4LE_{max}}\right\}}.
 \end{equation}
 Let $C_{\eta}=\frac{1}{1-\sqrt{\eta}}$. Then with probability at least $1-3T\exp(-r)$ there exists a surrogate model $\widehat\w$ on the connected line between $\w_0$ and $\w_T$:
    {\small
  \begin{align}
       &\mathcal{L}_{sim}(\bm w_0 - \bm w_T, \nabla\widehat \w)\le2\eta+\frac{2\eta^2\beta^2T}{\gamma(1-G_2^2)}\frac{\ell(\w_0)-\ell(\w_T')}{\ell(\w_T')-\ell(\w^*)}\nonumber\\
    &{+ \frac{C_\eta(1+G_2)}{1-C_{GD}}\left(C_{GD}+\frac{LTE_{max}\eta(1+\eta\beta)^{T-1}}{\ell(\w_0)-\ell(\w_T')}\right)+G_2^2},
  \end{align}
}%
    where $\w^*$ is the global minimum.
\end{theorem}
\begin{remark}
The scaling effect in \Cref{thm:GD} also exists in SGD due to $C_{GD}$. Note that the process from $\w_0$ to $\w_T$ is just local training steps (compared to the fully trained model), it's natural to expect that $\ell(\w_0)-\ell(\w_T')$ is much less than $\ell(\w_T')-\ell(\w^*)$ in the second term. The ratio w.r.t.\ $E_{max}$ 
in the third term indicates that a small $\mathcal{L}_{sim}$ may not be achieved when gradient noise is large while loss reduction $\ell(\w_0)-\ell(\w_T')$ is small.
\end{remark}
\begin{algorithm}[tb]
   \caption{Surrogate Model Extension}
   \label{alg:sme}
\begin{algorithmic}[1]
   \STATE {\bfseries Input:} Victim's weights $\bm w_0, \bm w_T$; Local data size $N$; Iterations $K$; Learning rate $\eta_{\Tilde{D}}$ for the dummy data and $\eta_\alpha$ for $\alpha$; Loss function $\mathcal{L}$.
   \STATE Initialize $\Tilde{D}_0; \alpha_0 \leftarrow 0.5$.
   \FOR{each step $k=0\dots K-1$}
   \STATE $\widehat{\bm w} = \alpha_k\bm w_0 + (1-\alpha_k)\bm w_T$
   \STATE $\Tilde{D}_{k+1} = \Tilde{D}_k - \eta_{\tilde{D}}\nabla_{\Tilde{D}_k}\mathcal{L}(\Tilde{D}_k, \widehat{\bm w})$
   \STATE $\alpha_{k+1} = \alpha_k - \eta_\alpha \nabla_{\alpha_k}\mathcal{L}(\Tilde{D}_k, \widehat{\bm w})$
   \ENDFOR
   \STATE {\bfseries Output:} Reconstructed data $\Tilde{D}_K$.
\end{algorithmic}
\end{algorithm}
\begin{figure*}[t!]
    \centering
    \subfloat[\centering $E$=10, $B$=50, $R$=1]{\includegraphics[width=0.21\textwidth]{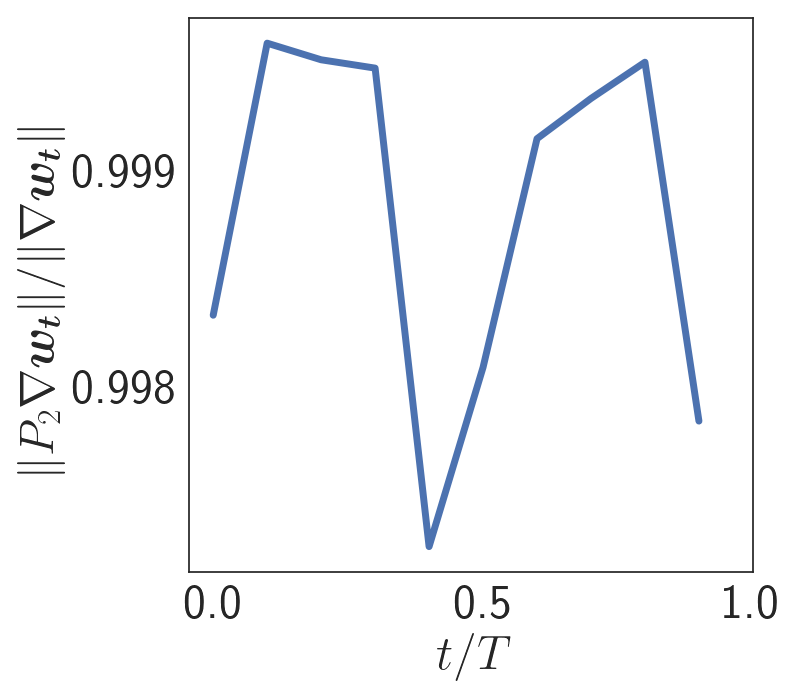}}%
    \subfloat[\centering $E$=50, $B$=50, $R$=1\label{fig:g2:50e50b1r}]{\includegraphics[width=0.21\textwidth]{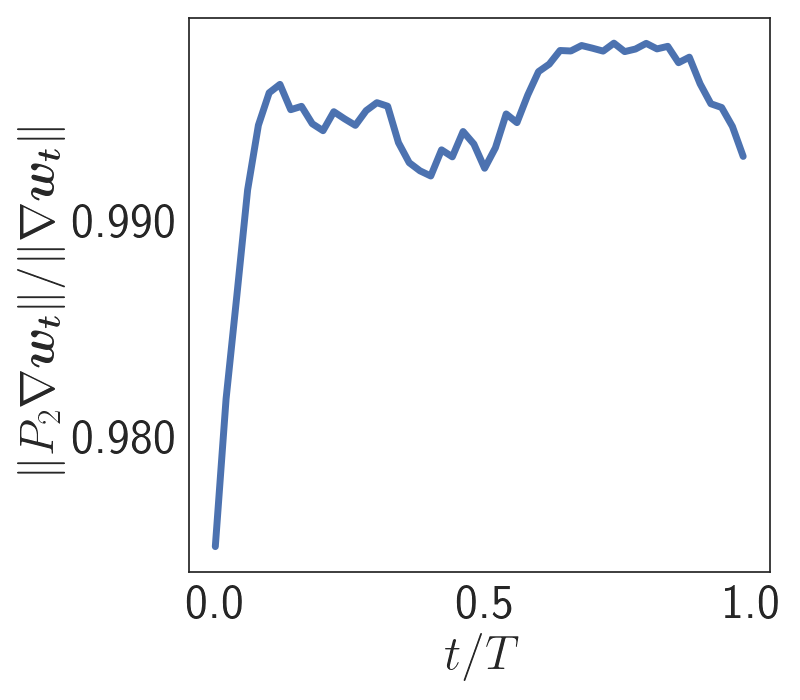}}%
    \subfloat[\centering $E$=10, $B$=10, $R$=1\label{fig:g2:10e10b1r}]{\includegraphics[width=0.21\textwidth]{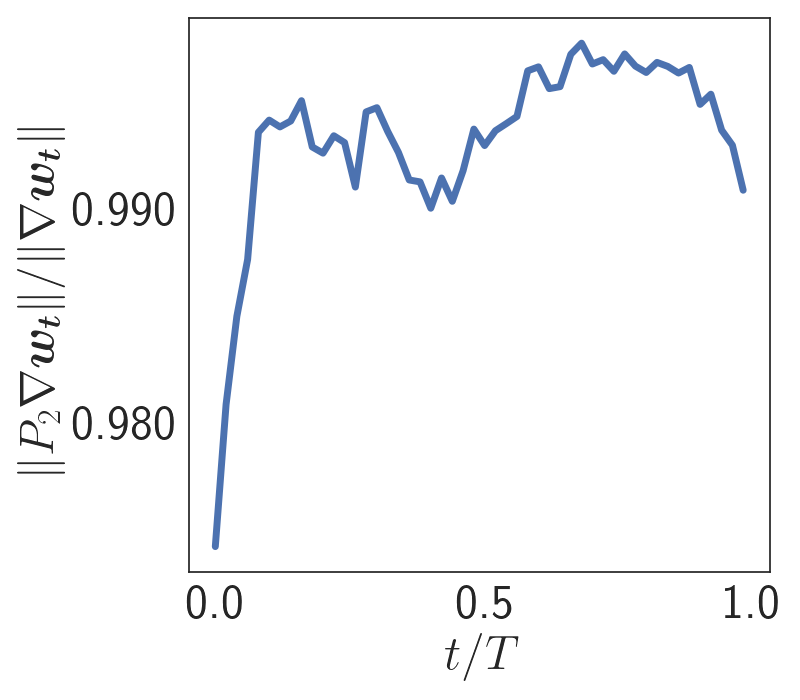}}%
    \subfloat[\centering $E$=10, $B$=50, $R$=20]{\includegraphics[width=0.21\textwidth]{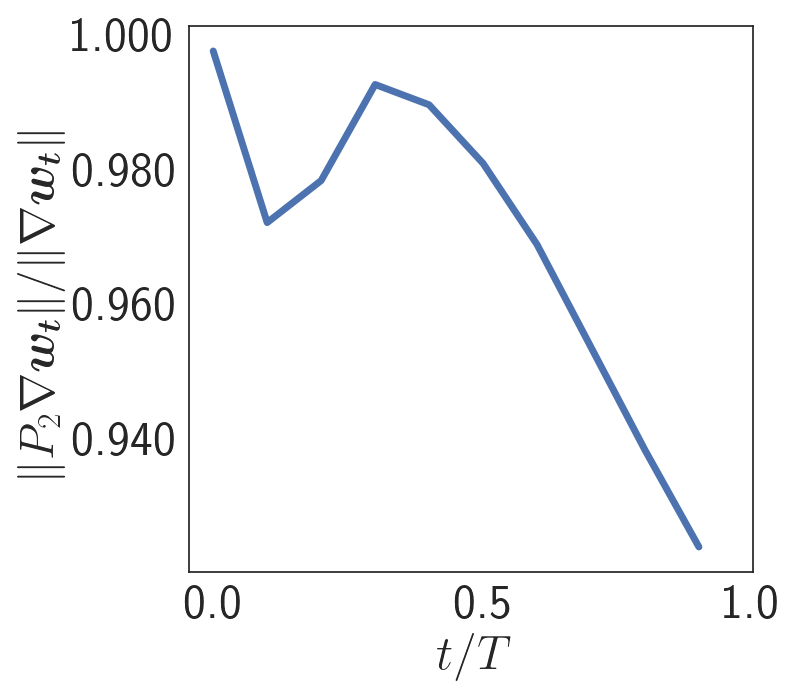}}%
    \caption{Gradient norm ratio $\|P_2\nabla \bm w_t\|/\|\nabla \bm w_t\|$ along the local steps $t=0\dots T-1$. We consider a client with $N = 50$ data points and in (a) we train a regular number of epochs $E=10$ with full gradient descent due to $B\geq N$. For comparison, we consider three other settings: (b) a large number of epochs $E=50$; (c) stochastic mini-batch gradient descent with $N/B = 5$; (d) training on a network that has been optimized for $R=20$ communication rounds. Note that (b) and (c) have the same local steps $T=50$, but (c) optimizes with SGD.}
    \label{fig:g2}
\end{figure*}
\begin{figure*}[t!]
    \centering
    \subfloat[\centering $E$=10, $B$=50, $R$=1]{\includegraphics[width=0.21\textwidth]{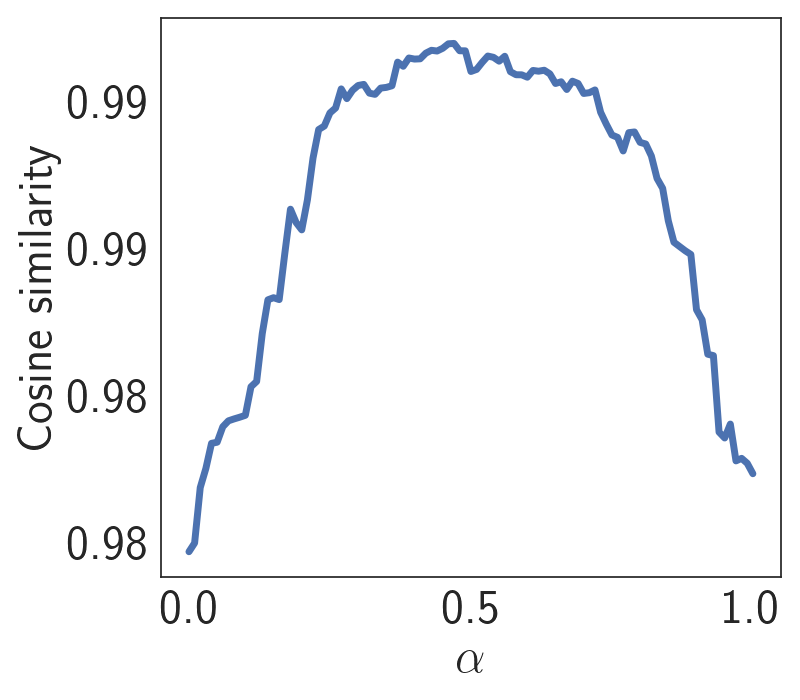}}%
    \subfloat[\centering $E$=50, $B$=50, $R$=1]{\includegraphics[width=0.21\textwidth]{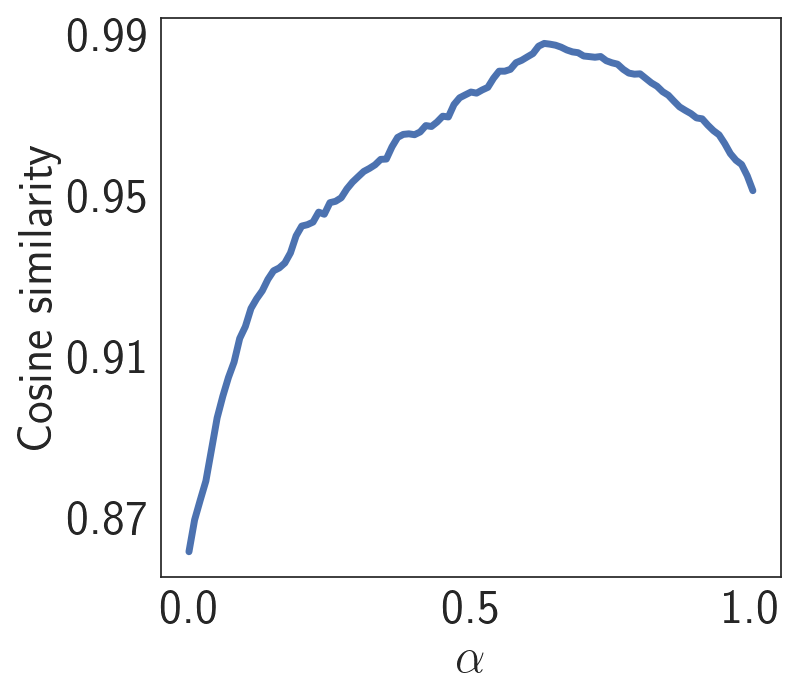}}%
    \subfloat[\centering $E$=10, $B$=10, $R$=1]{\includegraphics[width=0.21\textwidth]{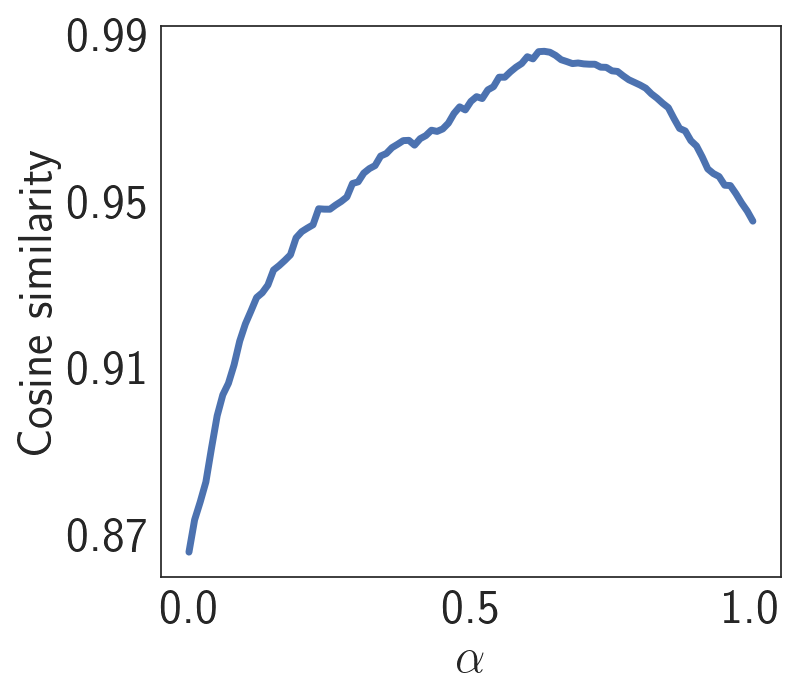}}%
    \subfloat[\centering $E$=10, $B$=50, $R$=20]{\includegraphics[width=0.21\textwidth]{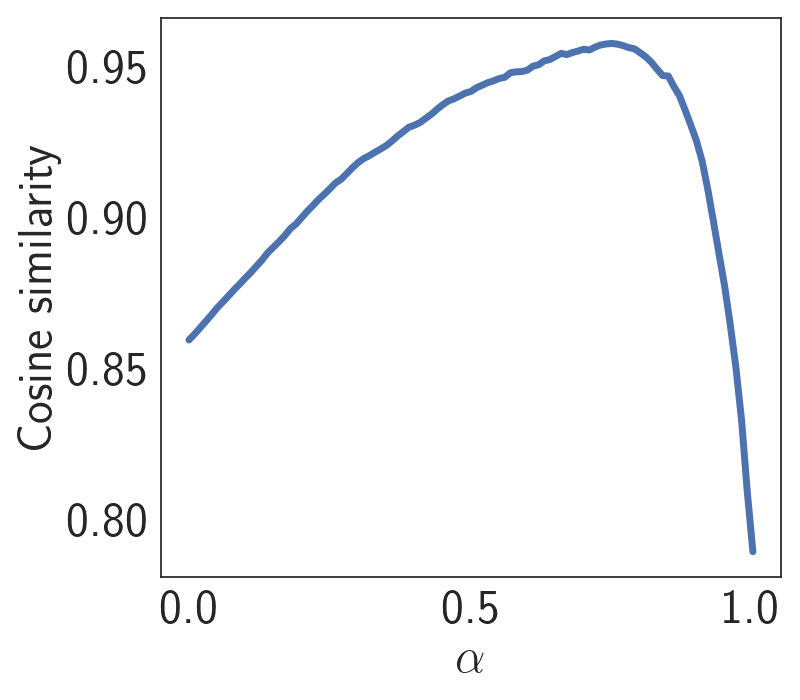}}%
    \caption{Cosine similarity between $\bm w_0 - \bm w_T$ and $\nabla\widehat \w$ vs.\ $\alpha$. The settings are aligned with \Cref{fig:g2}, please refer to the caption above.}
    \label{fig:cos}
\end{figure*}
\subsection{Implementation of the Surrogate Model Extension}
\label{sec:method:impl}
Our method \verb|SME| is easy to implement. Based on the objective of \verb|SME| in \Cref{eq:sme}, in addition to the gradient inversion steps, we only need to update the surrogate model's weights w.r.t.\ $\alpha$ before the forward propagation and optimize $\alpha$ after the backward propagation. \Cref{alg:sme} presents the optimization steps of our approach.

It is worth emphasizing that the additional computation caused by \verb|SME| due to the optimization of $\alpha$ is negligible. Since reconstruction loss $\mathcal{L}$ only depends on $\alpha$ through $\hw$ due to $\hw = \alpha \bm w_0 + (1-\alpha)\bm w_T$, we have:
\begin{align}
    \nabla_\alpha\mathcal{L}(\bm w_0 - \bm w_T, \nabla \hw)
    = \frac{\partial \mathcal{L}}{\partial \hw}\cdot(\bm w_T - \bm w_0),
\end{align}
which is simply an inner product of the gradient and weight update. Compared with the simulation methods as discussed in \Cref{sec:DLFA}, \verb|SME| is expected to run much faster.

\section{Experiments}
\label{sec:experiments}
In this section, we present empirical evidence for our analysis and validate the performance of our method \verb|SME|. We compare \verb|SME| with the vanilla gradient inversion method \verb|IG| \ \cite{geiping_inverting_gradient}, which our extension is based on, and the simulation method \verb|DLFA| \ \cite{dimitrov2022data}, which achieves the current SOTA weight update attack. 

\textbf{Setup:} We conduct our experiments on two image classification datasets: FEMNIST~\cite{cohen2017emnist, caldas2018leaf} and CIFAR100~\cite{cifar10}, and investigate the influence of local training for different attack approaches w.r.t.\ different numbers of epochs $E$, batch sizes $B$, and local data size $N$. In particular, we mainly investigate the privacy risk of small clients with $N\leq 50$. No existing reconstruction attack recovers large data sizes (e.g.\ a few hundred). On the other hand, small clients, e.g.\ edge devices, are common in FL and may have lower amounts of data. We also evaluate on larger $N$ to demonstrate the superiority of our method in a broad range of scenarios. Additionally, we consider clients joining the FL group at different communication rounds $R$, such that the received network $\w_0$ is at different training stages. Following \citet{dimitrov2022data} we conduct attacks on two CNNs for the respective datasets. The networks consist of two convolutional layers followed by two fully-connected layers. We also provide results on other architectures in \Cref{sec:app:other_network}.
A detailed description of the datasets, federated learning, and hyperparameters for different attack approaches is given in \Cref{sec:app:experiment}.
\begin{table*}[t!]
\begin{center}
\begin{small}
\begin{tabular}{ccccc}
\toprule
     &(E=10, B=50, R=1) 
&(E=50, B=50, R=1)  &(E=10, B=10, R=1)& (E=10, B=50, R=20)\\
\hhline{~----}
$\maE [\min_t \frac{\|P_2\nabla \bm w_t\|}{\|\bm w_t\|}]$
&.996$\pm$.002 &.977$\pm$.009 &.977$\pm$.008 &.960$\pm$.033\\
\midrule
$\maE[\Cosim|\alpha=0]$
&.984$\pm$.008 &.867$\pm$.041 &.872$\pm$.037 &.825$\pm$.055\\
\midrule
$\maE[\max_{\alpha\in[0,1]}\Cosim]$
&.997$\pm$.002 &.980$\pm$.007 &.979$\pm$.007 &.973$\pm$.016\\
\bottomrule
\end{tabular}
\end{small}
\caption{Statistical evaluations corresponding to \Cref{fig:g2,fig:cos}. We repeat the experiment 100 times to compute the mean and standard deviation. $\Cosim$ is a shorthand for the cosine similarity between $\w_0 - \w_T$ and $\nabla \hw$.}
\label{tab:stats}
\end{center}
\end{table*}
\begin{table*}[t!]
\begin{center}
\resizebox{\textwidth}{!}{%
\begin{small}
\begin{tabular}{cccchahaha|d}
\toprule
\multirow{2}{*}{Dataset}     & & & &\multicolumn{2}{c}{DLFA} 
&\multicolumn{2}{c}{IG}  &\multicolumn{2}{c}{SME (ours)}&\multicolumn{1}{c}{}\\
\hhline{~~~~--||--||---}
  &$E$ &$N$ &$T$ &$\mathcal{L}_{sim}\downarrow$     &PSNR $\uparrow$     &$\mathcal{L}_{sim}\downarrow$     &PSNR $\uparrow$ &$\mathcal{L}_{sim}\downarrow$     &PSNR $\uparrow$&$\bm \Delta$ PSNR\\
\midrule
\multirow{6}{*}{FEMNIST} & 10 &10 &10 &$.021\pm.001$ &$24.9\pm0.2$ &$.044\pm.002$ &$27.8\pm0.3$ &$.019\pm.001$ &\bftab 30.3 $\pm$ \bftab0.3& $+2.5$\\
& 20 &10 &20 &$.019\pm.001$ &$25.4\pm0.2$ &$.090\pm.003$ &$24.2\pm0.3$ &$.019\pm.001$ &\bftab 28.6 $\pm$ \bftab0.3& $+3.2$\\
& 50 &10 &50 &$.016\pm.002$ &\bftab 26.4 $\pm$ \bftab 0.3 &$.202\pm.005$ &$19.5\pm0.2$ &$.050\pm.004$ &$25.7\pm0.3$&$-0.7$\\
& 10 &50 &50 &$.015\pm.001$ &$21.7\pm0.1$ &$.091\pm.003$ &$19.1\pm0.2$ &$.027\pm.001$ &\bftab 22.2 $\pm$ \bftab 0.2&$+0.5$\\
& 20 &50 &100 &$.014\pm.001$ &\bftab 21.7 $\pm$ \bftab 0.2 &$.176\pm.011$ &$17.0\pm0.2$ &$.037\pm.001$ &$21.5\pm0.2$&$-0.2$\\
& 50 &50 &250 &$.015\pm.001$ &\bftab 20.3 $\pm$ \bftab0.2 &$.322\pm.016$ &$15.0\pm0.1$ &$.065\pm.002$ &$20.1\pm0.2$&$-0.2$\\
\midrule
\multirow{6}{*}{CIFAR100} & 10 &10 &10 &$.017\pm.001$ &$26.7\pm0.2$ &$.050\pm.001$ &$26.0\pm0.1$ &$.024\pm.001$ &\bftab 28.5 $\pm$ \bftab 0.1&$+1.8$\\
&20 &10 &20 &$.013\pm.000$ &$26.0\pm0.2$ &$.102\pm.001$ &$22.3\pm0.1$ &$.040\pm.001$ &\bftab 26.9 $\pm$ \bftab 0.1&$+0.9$\\
& 50 &10 &50 &$.011\pm.000$ &\bftab 24.3 $\pm$ \bftab 0.2 &$.225\pm.002$ &$16.4\pm0.2$ &$.056\pm.001$ &$24.2\pm0.1$&$-0.1$\\
& 10 &50 &50 &$.023\pm.001$ &$20.3\pm0.1$ &$.094\pm.001$ &$17.9\pm0.1$ &$.035\pm.000$ &\bftab 23.5 $\pm$ \bftab 0.1&$+3.2$\\
& 20 &50 &100 &$.018\pm.000$ &$19.5\pm0.1$ &$.154\pm.002$ &$14.8\pm0.1$ &$.047\pm.001$ &\bftab 21.7 $\pm$ \bftab 0.1& $+2.2$\\
& 50 &50 &250 &N/A &N/A &$.280\pm.002$ &$12.1\pm0.0$ &$.056\pm.001$ &\bftab 18.3 $\pm$ \bftab 0.1&N/A\\
\bottomrule
\end{tabular}
\end{small}
}
\caption{Average reconstructed image quality measured by PSNR and similarity loss of the reconstruction objective $\mathcal{L}_{sim}$ on FEMNIST and CIFAR100. For clarity, we set batch size $B=10$ and change local data sizes $N$ and epochs $E$. Local steps $T=E\lceil N / B\rceil$. The best reconstruction results are bold. The difference of PSNRs between SME and the best baseline is given in the last column. We remark that for $\text{PSNR} < 18$ the reconstruction will be visually corrupted. Also refer to \Cref{fig:visual} for visualization.
Results of DLFA in the last row is not available, as it needs to allocate 102 Gigabytes of GPU memory, which we cannot support.
}
\label{tab:psnr}
\end{center}
\end{table*}

\textbf{Threat Model:} To implement our method \verb|SME| and the gradient inversion method \verb|IG|, the adversary only needs to know the model weights $\bm w_0$, $\bm w_T$ of a victim, and its local data size $N$. We note that clients in FL also need to transmit $N$ to the server due to weighted aggregation (Line 8, \Cref{alg:fedavg}). Therefore, any adversary intercepting the messages or an honest-and-curious server has the ingredients to conduct \verb|SME| and \verb|IG|. Whereas, \verb|DLFA| also needs to know the number of epochs $E$, batch size $B$, and learning rate $\eta$, so that the simulation can be implemented. This may limit the adversary to the server, which has knowledge of the training protocol. 
In our comparison experiment, we still grant \verb|DLFA| this additional required information. 

\subsection{Low-Rank Property of Local Steps}
In \Cref{sec:method:general} we show that if the magnitude of high-dimensional gradient mostly concentrates on a 2D plane $V_2$ then we are able to find a proper surrogate model $\hw$ on the connected line between $\bm w_0$ and $\bm w_T$ achieving high cosine similarity between $\nabla \hw$ and $\w_0 - \w_T$. To support \Cref{ass:1} empirically, we collect the true gradients of the local training steps $\{\nabla \bm w_t\}_{t=0}^{T-1}$ and compute the two principal components with the largest two singular values. We let $P_2$ consist of these two principal components. Then we project the gradients $\{\nabla \bm w_t\}_{t=0}^{T-1}$ onto $V_2$ using $P_2$ and compute the norm ratio $\{\|P_2\nabla \bm w_t\|/\|\nabla \bm w_t\|\}_{t=0}^{T-1}$. Furthermore, we measure the cosine similarity between the connected line $\bm w_0 - \bm w_T$ and surrogate gradient $\nabla \widehat{\bm w}$ along the line.

We sample $N = 50$ data points from FEMNIST and investigate different training situations in terms of epochs $E$, batch size $B$, and communication rounds $R$.
\Cref{fig:g2} shows that in different situations, more than 0.9 of the gradient magnitude concentrates on $V_2$. This implies that the low-rank property 
generally holds for the local updates of small clients in FL.
\Cref{fig:cos} shows that there exists a $\nabla \widehat{\bm w}$ with much higher cosine similarity than $\nabla \bm w_0$ some where in the middle of the connected line, while the cosine similarity is always positive for any $\nabla \hw$. Especially, we observe that the cosine similarity along $\alpha$ has a reversed U-shape which indicates that it is adequate to use an optimization method to find the best $\alpha$. Statistical evaluations of multiple samplings are presented in \Cref{tab:stats}. We give more results of other settings in \Cref{sec:app:evidence}.
\begin{figure*}[t!]
\begin{center}
\centerline{\includegraphics[width=0.95\textwidth]{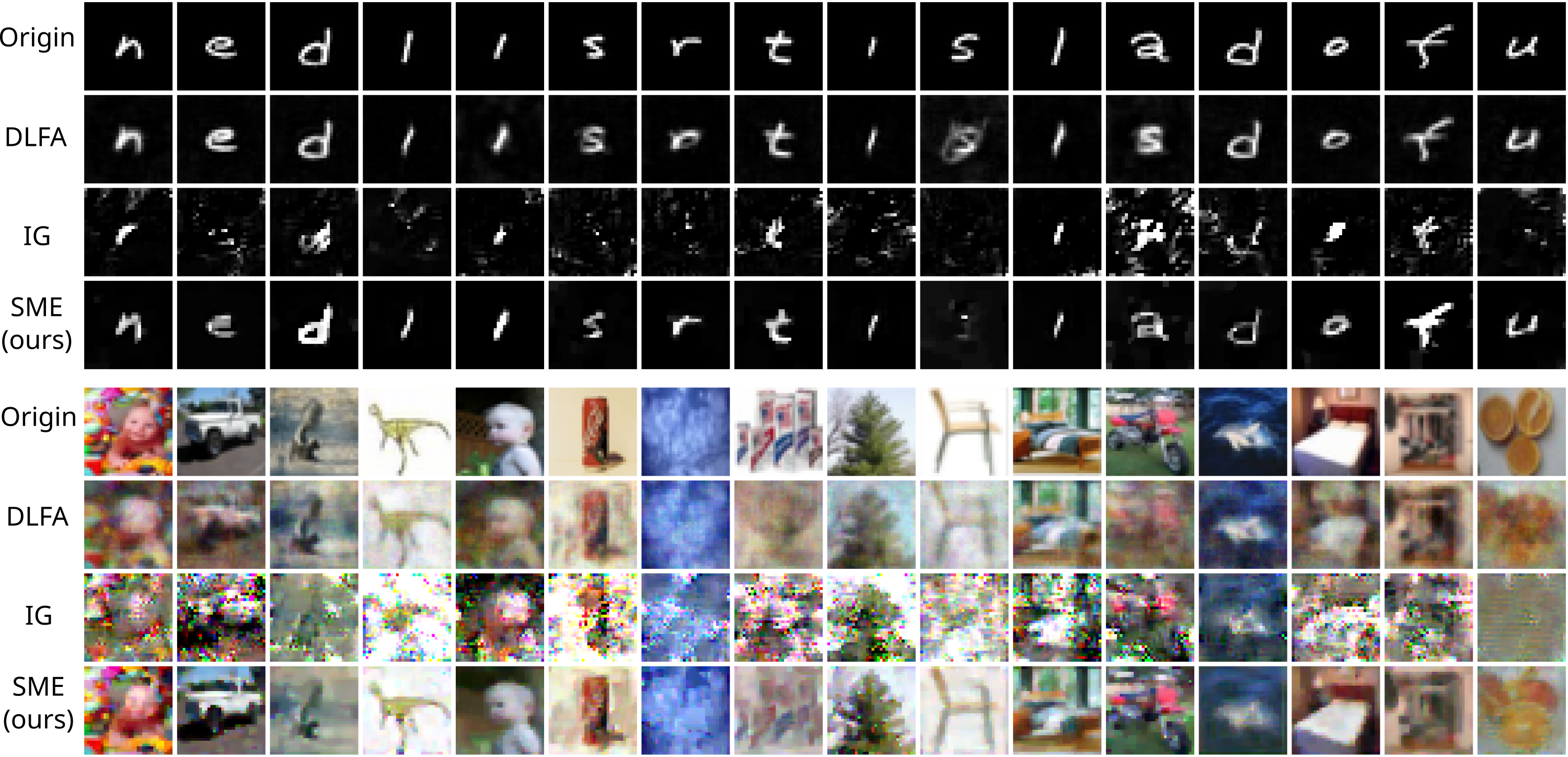}}
\caption{Visualization of the reconstructed images. The results are drawn from the setting ($E=20$, $N=50$, $T=100$) in \Cref{tab:psnr}. The reconstructed images are paired with the original images through \textit{linear sum assignment}. We randomly sample 16 out of 50 images of one reconstruction.}
\label{fig:visual}
\end{center}
\end{figure*}

\subsection{Reconstruction Performance}
In this part, we compare the quality of the reconstructed data across different approaches. In previous works, it has been proposed that the label information can be independently recovered with high accuracy by inspecting the weight update of the last linear classifier layer~\cite{zhao2020idlg, geng_2022_aaai, dimitrov2022data}. Thus, we assume the labels have been recovered, and compare the reconstructed images, which is arguably the primary interest of the adversary. To measure the quality of the reconstructed images, we use the \textit{Peak Signal-to-Noise Ratio} (PSNR). As the reconstructed images may have a different ordering from the original images, \textit{linear sum assignment} is used to find the optimal pairing before computing PSNR. We also present the cosine similarity loss $\mathcal{L}_{sim}$ between the original weight update and dummy gradient (or dummy weight update). For clarity, we set batch size $B = 10$ and conduct experiments with different epochs $E\in\{10, 20, 50\}$ and local data size $N\in\{10, 50\}$, such that when $N=10$ the full gradient is used, and when $N=50$ stochastic mini-batch gradient is computed. We execute attacks at the first communication round of FL when the network is randomly initialized. We also investigate the attacks at other training stages in~\Cref{sec:app:trained_network}. For each setting, we run 100 experiments, then compute the mean and standard error of PSNR and $\mathcal{L}_{sim}$.

\begin{figure}[t!]
\begin{center}
\centerline{\includegraphics[width=0.9\linewidth]{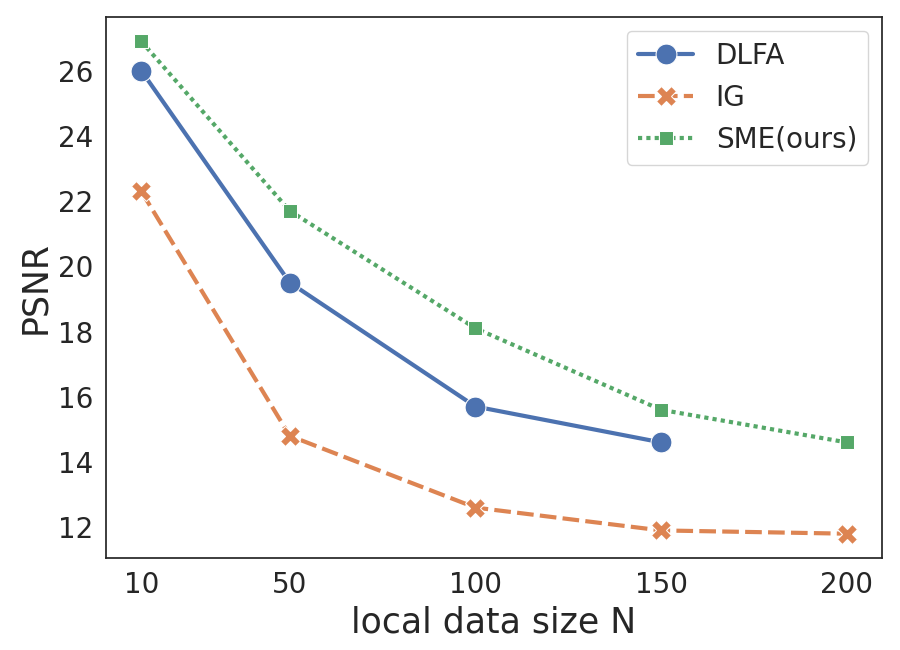}}
\caption{Average reconstructed image quality measured by PSNR vs. local data size $N$. The setting is (E = 20, B=10) on CIFAR100 from \Cref{tab:psnr}, we extend $N$ from 50 to 200. Experiments are repeated 100 times then we compute the mean. DLFA requires more than 100 Gigabytes of GPU memory for $N=200$, which we cannot support.}
\label{fig:largeN}
\end{center}
\end{figure}

Based on \Cref{tab:psnr}, we observe that: (i) Compared with the base method \verb|IG|, our method \verb|SME| achieves significantly lower similarity loss $\mathcal{L}_{sim}$. As a result, \verb|SME| improves the reconstruction quality measured with PSNR by a large margin. The benefit of \verb|SME| is consistent across different settings, which proves that \verb|SME| is an efficient and robust extension to the gradient inversion method in attacking weight update of many iterations; (ii) Compared with \verb|DLFA|, our \verb|SME| obtains competitive results and outperforms \verb|DLFA| in several settings. 
Additionally, we note that \verb|DLFA| usually achieves lower $\mathcal{L}_{sim}$ than \verb|SME|, even with lower reconstruction quality. This may indicate that there are multiple possible transitions from $\bm w_0$ to $\bm w_T$ with different data. Fitting weight updates with simulation does not necessarily reconstruct the original data.

\Cref{fig:visual} shows the reconstruction samples of the setting ($E=20$, $N=50$, $T=100$). As we can see, \verb|IG| fails in reconstruction when there are 100 local steps, as the main features of the images are corrupted and the reconstruction contains many artifacts. Whereas, our \verb|SME| revives the gradient inversion method and delivers high-quality reconstruction which is competitive to or better than \verb|DLFA|. We provide more visualizations in \Cref{sec:app:visual}.

\begin{table*}[t!]
\begin{center}
\begin{small}
\begin{tabular}{ccchahaha|dd}
\toprule
\multirow{2}{*}{Dataset}     & & &\multicolumn{2}{c}{DLFA} 
&\multicolumn{2}{c}{IG}  &\multicolumn{2}{c}{SME (ours)}& \multicolumn{2}{c}{Ratio}\\
\hhline{~~~--||--||----}
  &$E$ &$T$ &Mem.    & Time   &Mem.    & Time &Mem.     & Time  &Mem.     & Time \\
\midrule
\multirow{3}{*}{FEMNIST}
& 10 &50 &2.1 &5.9 &1.8 &.29 &1.8 &.29 &$1.2\times$& $20\times$\\
& 20 &100 &2.4 &11.8 &1.8 &.29 &1.8 &.29&$1.3\times$ &$41\times$\\
& 50 &100 &3.9 &30.1 &1.8 &.29 &1.8 &.29&$2.2\times$&$103\times$\\
\midrule
\multirow{3}{*}{CIFAR100}
& 10 &50 &11.1 &33.2 &2.8 &1.6 &2.8 &1.6&$7\times$&$21\times$\\
& 20 &100 &29.9 & 67.6 &2.8 &1.6 &2.8 &1.6&$19\times$&$42\times$\\
& 50 &250 &$102^*$ &N/A &2.8 &1.6 &2.8 &1.6&$64\times$&N/A\\
\bottomrule
\end{tabular}
\end{small}
\caption{Computation resource required by different approaches. We present the GPU memory footprint in Gigabytes and computational time in Hours for 100 times of attack executed on a V100 GPU card. All approaches are implemented in \textit{JAX}~\cite{jax2018github}. We disable the preallocation behavior and exclude the compilation time. We present the measurements of the experiments in \Cref{tab:psnr} with local data size $N=50$. Note that SME and IG run $2.5\times$ reconstruction iterations than DLFA, but still, run up to $100\times$ faster. The runtime of DLFA in the last row is not available as it requires 102 Gigabytes of GPU memory estimated by \textit{JAX}, which we cannot support.
}
\label{tab:cost}
\end{center}
\end{table*}

To evaluate the effectiveness of \verb|SME| on larger local dataset, we conduct experiments with $N$ up to 200. \Cref{fig:largeN} shows that: (i) as N increases, the reconstruction of all methods becomes worse as the search space expands. (ii) SME consistently improves over the vanilla gradient inversion method and obtains better results compared to the SOTA baseline DLFA.

\subsection{Computational Efficiency}
\label{sec:experiments:computation}
As discussed in \Cref{sec:DLFA}, the algorithm of \verb|DLFA| demands a large amount of computational resources as the number of local steps increases, while both \verb|SME| and \verb|IG| do not involve a long chain of simulation. We also show in \Cref{sec:method:impl} that the cost of optimizing the surrogate model in \verb|SME| is negligible. To demonstrate this empirically, we present the runtime and memory footprint of different approaches. The results are shown in \Cref{tab:cost}. We see both \verb|SME| and \verb|IG| have a consistent and small runtime across different settings. In particular, our method \verb|SME| achieves the SOTA performance on reconstruction while running up to $100\times$ faster than the SOTA baseline \verb|DLFA|. We remark that the runtime of \verb|SME| is nearly the same as \verb|IG|. More precisely, the increase due to surrogate model optimization is less than $1\%$ of the cost. Additionally, \verb|SME| and \verb|IG| both have moderate and consistent (w.r.t.\ local steps $T$) memory footprint. While GPU memory allocated by \verb|DLFA| increases dramatically when the dimension of data grows or the number of local steps increases. This demand of \verb|DLFA| may limit its application to some simple tasks. In contrast, \verb|SME| is applicable on typical hardware and achieves the SOTA performance while running up to $100\times$ faster than \verb|DLFA|, which shows the potential to launch large-scale attacks.
\section{Discussion and Conclusion}
\label{sec:conclusion}
In this work, we propose a surrogate model extension for gradient inversion. Our method is based on our key insights of the characteristic of 2D gradient flow and the low-rank property of local training steps. We analyze our method theoretically and empirically verify that our method largely strengthens the ability of gradient inversion in attacking weight updates of many training iterations. Compared with the SOTA baseline \verb|DLFA|, our method \verb|SME| achieves the SOTA performance of reconstruction while running up to $100\times$ faster and demanding less GPU memory. These indicate that adversaries can in fact launch effective attacks using low-end devices.

\textbf{Limitations and Future Work.}
In this work, we consider that federated optimization is performed with a standard SGD optimizer. Some adaptive optimizer, e.g.\  Adam~\citep{kingma2014adam}, may lead to different optimization properties and thus interfere with \verb|SME|. This needs further investigation. On the other side, incorporating a more informative prior may empower \verb|SME| to reconstruct larger datasets or input, which has succeeded in vanilla gradient inversion scenarios \citep{yin_cvpr_see_through_gradient,jeon2021gradient}. Moreover, rather than interpolation of $\w_0$ and $\w_T$, there may exist other construction of $\mathcal{W}$ that makes the general objective of \verb|SME| effective, i.e.\  \Cref{eq:general_obj}. We encourage further study in these directions.

\textbf{Societal Impact.}
We believe our effort in studying adversarial attack and presenting it to the open community is beneficial to society. We emphasize that our study does not attempt to claim that federated learning is completely insecure, since it certainly provides a layer of privacy protection by avoiding data collection. Instead, our work intends to evaluate the true risk hiding in the weight sharing in collaborative learning schemes. We hope our research will motivate further research of defense against privacy attacks.

\textbf{Acknowledgements.}
This research received funding from the Flemish Government (AI Research Program) and the Research Foundation - Flanders (FWO) through project number G0G2921N. {Ruicong Yao would like to thank the guidance and encouragement from Tim Verdonck and Jakob Raymaekers in this research process.}

\bibliography{example_paper}
\bibliographystyle{icml2023}

\appendix
\onecolumn
\section{Related Works}
\label{sec:app:related_works}
In addition to gradient inversion methods, another line of work aims to extract training data from a trained network \cite{Fredrikson_model_inversion, Yang2019, He2019_model_inversion, Carlini_secret_sharer, carlini_large_language_model, haim2022reconstructing}.
Many other works focus on extracting partial information from the data, e.g. label-only inference and membership inference~\cite{membership_inference, zhao2020idlg, carlini2021membership, Choquette-Choo_label_only}.

On the other hand, recently many defense strategies and privacy-enhancing methods have emerged. Differential privacy provides a rigorous definition of privacy based on a probabilistic perspective \cite{DP2, DP3, private_erm}. A long line of works adopts it in deep learning to alleviate the privacy leakage from the model or intermediate training products like weights, to name a few~\cite{DP-SGD, tempered_sigmoid, handcrafted-sgd}. However, differential privacy requires injecting extensive Gaussian noise into the gradients and thus impedes the model performance~\cite{private_erm, non_convex_risk}.
Moreover, many works exploit encryption technology~\cite{Gilad-Bachrach_cryptonet, Bonawitz_secure_aggregation, so_secure_aggregation, FHE}, such that messages communicated between participants are in cipher text, thus adversaries cannot extract the sensitive information directly. However, these methods in general give rise to a significant additional computational overhead.
\section{Reconstruction Gradient of Simulation Methods}
\label{sec:app:sim_grad}
Suppose that for each simulation step, $\Tilde{D}$ is used to train the network, and consider the following loss function of a simulation method:
\begin{align}
    \label{eq:SIM}
    \mathcal{L} &= \|\tilde{\bm w}_T - \bm w_0 - (\bm w_T - \bm w_0)\|\\
    &= \|\eta(\nabla \bm w_0 + \nabla \Tilde{\bm w}_1 + \dots + \nabla \Tilde{\bm w}_{T-1}  ) - \bm w_T + \bm w_0\|,
\end{align}
the reconstruction gradient $\nabla_{\Tilde{D}}\mathcal{L}$ can be derived as:
\begin{equation}
    \label{eq:vnmn}
    \nabla_{\Tilde{D}}\mathcal{L} = \frac{\partial \mathcal{L}}{\partial \nabla \bm w_0}\frac{\partial \nabla \bm w_0}{\partial \Tilde{D}} + \frac{\partial \mathcal{L}}{\partial \nabla \Tilde{\bm w}_1}\frac{\partial \nabla \Tilde{\bm w}_1}{\partial \Tilde{D}} + \dots + \frac{\partial \mathcal{L}}{\partial \nabla \Tilde{\bm w}_{T-1}}\frac{\partial \nabla \Tilde{\bm w}_{T-1}}{\partial \Tilde{D}}.
\end{equation}
Using the chain rule, we have for $\frac{\partial \nabla \Tilde{\bm w}_1}{\partial \Tilde{D}}$:
\begin{align}
    \frac{\partial \nabla \Tilde{\bm w}_1}{\partial \Tilde{D}} &= \frac{\partial \nabla_{\bm w}\ell(\bm w, \Tilde{D})}{\partial \Tilde{D}}|_{\bm w = \Tilde{\bm w}_1} + \frac{\partial \nabla \Tilde{\bm w}_1}{\partial \Tilde{\bm w}_1}\frac{\partial \Tilde{\bm w}_1}{\partial \nabla \Tilde{\bm w}_0}\frac{\partial \nabla \bm w_0}{\partial \Tilde{D}}\\
    \label{eq:fdgd}
    &= \frac{\partial \nabla_{\bm w}\ell(\bm w, \Tilde{D})}{\partial \Tilde{D}}|_{\bm w = \Tilde{\bm w}_1} - \eta\frac{\partial \Tilde{\bm w}_1}{\partial \nabla \Tilde{\bm w}_1}\frac{\partial \nabla \bm w_0}{\partial \Tilde{D}},
\end{align}
where \Cref{eq:fdgd} uses the fact that $\Tilde{\bm w}_1 = \bm w_0 - \eta\nabla \bm w_0$. Then for the reconstruction gradient led by $\nabla \bm w_t$, considering that  $\Tilde{\bm w}_t = \bm w_0 - \eta(\nabla\bm w_0 + \dots + \nabla \Tilde{\bm w}_{t-1})$, we have:
\begin{equation}
    \label{eq:noip}
    \frac{\partial \nabla \Tilde{\bm w}_t}{\partial \Tilde{D}} = \frac{\partial \nabla\ell_{\bm w}(\bm w, \Tilde{D})}{\partial \Tilde{D}}|_{\bm w = \Tilde{\bm w}_t} - \eta\frac{\partial \nabla\Tilde{\bm w}_t}{\partial \Tilde{\bm w}_t}(\frac{\partial \nabla\bm w_0}{\partial \Tilde{D}} + \frac{\partial \nabla\Tilde{\bm w}_1}{\partial \Tilde{D}} + \dots + \frac{\partial \nabla\Tilde{\bm w}_{t-1}}{\partial \Tilde{D}})
\end{equation}
Substituting \Cref{eq:noip} into \Cref{eq:vnmn} for each $t=1\dots T-1$ and rearranging, we have:
\begin{align}
    \label{eq:nnmm}
    \begin{split}
        \nabla_{\tilde{D}} \mathcal{L} = &-\eta\frac{\partial \mathcal{L}}{\partial \Tilde{\bm w}_{T-1}}(\frac{\partial \nabla\bm w_0}{\partial \Tilde{D}} + \frac{\partial \nabla\Tilde{\bm w}_1}{\partial \Tilde{D}} + \dots + \frac{\partial \nabla\Tilde{\bm w}_{T-3}}{\partial \Tilde{D}} + \frac{\partial \nabla\Tilde{\bm w}_{T-2}}{\partial \Tilde{D}})\\
        &-\eta \frac{\partial \mathcal{L}}{\partial \Tilde{\bm w}_{T-2}}(\frac{\partial \nabla\bm w_0}{\partial \Tilde{D}} + \frac{\partial \nabla\Tilde{\bm w}_1}{\partial \Tilde{D}} + \dots + \frac{\partial \nabla\Tilde{\bm w}_{T-3}}{\partial \Tilde{D}})\\
        &\vdots\\
        &-\eta \frac{\partial \mathcal{L}}{\partial \Tilde{\bm w}_1}\frac{\partial \nabla\bm w_0}{\partial \Tilde{D}}\\
        &+ \frac{\partial \mathcal{L}}{\partial \nabla \Tilde{\bm w}_0}\frac{\partial \nabla\ell_{\bm w}(\bm w, \Tilde{D})}{\partial \Tilde{D}}|_{\bm w = \Tilde{\bm w}_0} + \frac{\partial \mathcal{L}}{\partial \nabla \Tilde{\bm w}_1}\frac{\partial \nabla\ell_{\bm w}(\bm w, \Tilde{D})}{\partial \Tilde{D}}|_{\bm w = \Tilde{\bm w}_1} + \dots + \frac{\partial \mathcal{L}}{\partial \nabla \Tilde{\bm w}_{T-1}}\frac{\partial \nabla\ell_{\bm w}(\bm w, \Tilde{D})}{\partial \Tilde{D}}|_{\bm w = \Tilde{\bm w}_{T-1}}.
    \end{split}
\end{align}
\Cref{eq:nnmm} shows that the reconstruction gradient $\nabla_{\Tilde{D}}\mathcal{L}$ in the simulation method has a complexity that increases with the number of local steps $T$.

\newpage
\section{Proofs }
\label{sec:app:proof}
In this section we give the proofs of all the theoretical results in \Cref{sec:theory}. Recall \Cref{ass:1} in \Cref{sec:theory}, we say a function $\ell(\w):\mathbb R^{p}\rightarrow\mathbb R^{q}$ is $L$-Lipschitz if for $\forall \w, \bm v\in\mathbb R^p$, it holds that
\begin{equation}
    ||\ell(\w)-\ell(\bm v)||\le L||\w-\bm v||.
\end{equation}
We say a function $\ell:\mathbb R^p\rightarrow \mathbb R$ is $\gamma$-strongly convex if for $\forall \w, \bm v\in\mathbb R^p$, it holds that,
\begin{equation}\label{def:sc}
 \ell(\w)\ge \ell(\bm v)+\langle \nabla\bm v, \w-\bm v\rangle+\frac{\gamma}{2}||\w-\bm v||^2.
\end{equation}
A well-known property of strongly convex functions is that (cf.\ \cite{Bottou2018}) for $\forall \w\in \mathbb R^p$ and global minimum $\w^*$, it holds that
\begin{equation}\label{}
2\gamma(\ell(\w)-\ell(\w^*))\le||\nabla \ell(\w)||^2.
\end{equation}

\subsection{Proof of Proposition \ref{prop:2dspace}}\label{sec:app:proof:4.1}
The basic idea has been given in \Cref{sec:method:2D}. Here, we further illustrate how we prove the case when both $\nw_0,\nw_T$ point to the right hand side in \Cref{fig:illustration}. Since $\bm w_T$ is the end point of the gradient flow, we can argue that there exists a model $\bm w_{t'}$ on the gradient flow near $\bm w_T$
    located at the left hand side of $\bm w_T-\bm w_0$. Therefore, the intersection of the gradient flow and the connected line is non-empty. We can further deduce that there exists at least one surrogate model $\widehat\w' $ on the connected line such that its gradient $\nabla\widehat\w'$ points to the left hand side of $\bm w_T-\bm w_0$ or is parallel to $\bm w_T-\bm w_0$. Thus in both situations, we can find some $\widehat\w$ on the connected line 
between $\widehat \w'$ and $\bm w_0$
such that its gradient is parallel to $\w_T-\w_0$. A mathematical proof is given in the following.
\begin{proof}[Proof of Proposition \ref{prop:2dspace}]
    To be aligned with the illustration in Section \ref{sec:method:2D}, we prove for the gradient ascent case from $\w_0$ to $\w_T$. Here, the condition would instead be $\langle\nw_t,\w_T-\w_0\rangle\ge0$, $\forall t\in[0,T]$ (we reverse $w_0$ and $w_T$) and the second claim would be $\nabla\widehat \w\parallel \w_T-\w_0$ assuming convexity. The proof for the gradient descent flow is similar. We do not consider the case that there is a $\w^*$ on the connected line such that $\nw^*=0$, which is typically unusual. Let $\bm w^\perp$ be a vector which is orthogonal to $\bm w_T-\bm w_0$. Our first claim immediately follows if 
    \begin{equation}
    \langle\nabla\bm w_0,\bm w^\perp\rangle\cdot \langle\nabla\bm w_T,\bm w^\perp\rangle\le0,  
    \end{equation}
    since $\nabla \widehat\w$ is continuous on the connected line so that the inner product attains zero for some choice of $\alpha$. 
    
    Otherwise, we assume without loss of generality that both inner products are positive. Mathematically it means that there exists a neighbourhood $\mathcal O_{\bm w_T}$ of $\bm w_T$ such that $\langle\nabla\bm w_{t},\bm w^\perp\rangle>0$ for all $\bm w_{t}\in\mathcal O_{\bm w_T}$. Moreover for such $\w_t$,
    \begin{equation}\label{proof4.1:1}
        \langle \bm w_{t}-\bm w_0,\bm w^\perp\rangle = \langle \bm w_{T}-\bm w_0,\bm w^\perp\rangle-\int_{t}^T\langle\nabla\bm w_{s},\bm w^\perp \rangle ds<0. 
    \end{equation}
    On the other hand, we can find a neighbour of $\mathcal{O}_{\w_0}$ of $\w_0$ such that $\langle\nw_t,\w^\perp\rangle>0$ and $\langle \w_t-\w_0,\w^\perp\rangle>0$ for $\forall \w_t\in\mathcal{O}_{\w_0}$. Combining these two facts, we know by continuity that there exists a non-empty set $A$ of all $\alpha$ s.t.\ {$\alpha\in\mathbb R$} and $\widehat\w_\alpha:=\alpha \bm w_0+(1-\alpha)\bm w_T$ is on the gradient flow. By the assumption that $\langle\nw_t,\w_T-\w_0\rangle\ge0$, $\forall t\in[0,T]$ (in the gradient ascent flow case), the projection of $\w_t$ onto $\w_T-\w_0$ is always an interpolation of $\w_0$ and $\w_T$. Thus we have $A\subset(0,1)$.
    
    We now assume that $\langle\nabla\widehat\w_\alpha,\w^\perp\rangle>0$ for $\forall \alpha\in A$ and prove by contradiction. Let $\alpha_{\inf}$ be the infimum of $A$ and $t_\alpha$ be such that $\w_{t_\alpha}=\widehat\w_\alpha$. By continuity we have either {$\langle{\nabla \widehat\w_{\alpha_{\inf}}},\w^\perp\rangle = 0$} or ${\alpha_{\inf}} \in A$. We continue with the latter. If $\alpha_{\inf}>0$, then by a similar argument as before we can show that the gradient flow and the connected line intersects with each other for some $0<t'<t_{\alpha_{\inf}}$, which is a contradiction. If otherwise $\alpha_{\inf}=0$ we can find an $\alpha'\in A$ s.t.\ $\w_{t_{\alpha'}}\in\mathcal{O}_{\w_0}$. Thus by our assumption on $\nabla\widehat\w_\alpha$, there exists a $0<t'<t_{\alpha'}$ s.t.\ $\langle \w_{t'}-\w_0,\w^\perp\rangle<0$ by a similar argument to \cref{proof4.1:1}, which contradicts the property of $\mathcal{O}_{\w_0}$. 
    
    Therefore in any case, there exists an $\alpha$ s.t.\ $\langle\nabla\widehat\w_\alpha,\w^\perp\rangle\le0$ where $\w_\alpha$ is on the connected line and the gradient flow. Then, the problem is reduced to the basic one mentioned at the beginning. 

    {Finally, we show $\nabla{\widehat \w}_\alpha\parallel \w_T-\w_0$ under convexity and gradient ascent flow. Otherwise, we fix an $\alpha$ s.t. $\langle \nabla{\widehat \w}_\alpha, \w_T-\w_0\rangle=-1$. We can thus identify an $\alpha'\in(0,\alpha)$ s.t.\ $\ell(\widehat\w_\alpha)<\ell(\widehat\w_{\alpha'})$ due to
    \begin{equation}
    \ell(\widehat\w_\alpha)-\ell(\widehat\w_{\alpha'})=\int_{0}^1\langle \nabla((1-s)\widehat\w_{\alpha'}+s\widehat\w_{\alpha}),\widehat\w_\alpha-\widehat\w_{\alpha'}\rangle ds 
    \end{equation}
    and the continuity of the gradient. On the other hand, since $\langle\nw_0,\w_T-\w_0\rangle\ge0$, the loss along the connected line first increases. Therefore, we can identify $0\le\alpha_1\le\alpha_2\le\alpha_3\le1$ such that $\ell(\widehat\w_{\alpha_2})>\ell(\widehat\w_{\alpha_1})$ and $\ell(\widehat\w_{\alpha_2})>\ell(\widehat\w_{\alpha_3})$, while $\widehat\w_{\alpha_2}$ is an interpolation of $\widehat\w_{\alpha_1}$ and $\widehat\w_{\alpha_3}$ This contradicts the convexity.}
\end{proof}
{By a similar argument to the last paragraph of the proof, we can show a property of gradient flow on convex objects:}

\begin{corollary}\label{cor:convexity}
    {Suppose $\w_t$, $t\in[0,T]$ is a gradient flow based on a convex loss $\ell(\w)$, then it holds that $\langle \nabla\widehat\w_\alpha,\w_0-\w_T\rangle\ge 0$, where $\widehat\w_\alpha:=\alpha \bm w_0+(1-\alpha)\bm w_T$, $\alpha\in[0,1]$.}
\end{corollary}

\subsection{Proof of Theorem \ref{thm:GD}}\label{sec:app:proof:GD}

{We first state a simple lemma which helps the proof of the theorem.}
\begin{lemma}\label{lem:c.2}
{Let $\nw \in\mathbb R^p$, $p\ge 2$ and $P_2$ satisfy \cref{eq:ass1} in Assumption \ref{ass:1}. Let $\bm v\in \mathbb R^d$ be another vector. It holds that
\begin{equation}
    \langle P_2\nw, P_2\bm v\rangle=\langle\nw,\bm v\rangle - \langle P_2^\perp\nw, P_2^\perp\bm v\rangle\ge(\Cosim(\nw,\bm v)-G_2)||\nw||||\bm v||.
\end{equation}
Here $\Cosim$ stands for the cosine similarity between the two vectors which is equal to $1-\mathcal{L}_{sim}$.}
\end{lemma}
{The proof of the lemma is straightforward. Now we prove the theorem. We begin by first identifying a surrogate model $\widehat \w$ that exhibits a large cosine similarity $\Cosim(P_2\widehat\w,P_2(\w_0-\w_T))$. Subsequently, we establish a lower bound for $\Cosim(\widehat\w,\w_0-\w_T)$ (or equivalently an upper bound for $\mathcal{L}_{sim}(\widehat\w,\w_0-\w_T))$) in high-dimensional space.}

\begin{proof}[Proof of Theorem \ref{thm:GD}]
{Since Proposition \ref{prop:2dspace} only works for 2D continuous gradient flow, we shall discuss how to adapt it to gradient descent method in high-dimensional space and estimate the error term $\Delta_{trun}$ related to \cref{eq:thm4.3}.}

{We first try to apply Proposition \ref{prop:2dspace} to $P_2\w(t)$, which is the projected gradient flow from $P_2\w(0)=P_2\w_0$, so as to identify an intermediate surrogate model $\widehat \w'$. Let $\w(T\eta)$ be the end point of the gradient flow. By \cref{lem:c.2} and the condition that $\Cosim(\nw(t),\w(0)-\w(T\eta))>G_2$, we have $ \langle P_2\nw(t), P_2(\w(0)-\w(T\eta))\rangle>0$. Thus \cref{prop:2dspace} shows that there exists a surrogate model $\widehat \w'$ satisfying $ \langle P_2\widehat\w', P_2(\w(0)-\w(T\eta))\rangle\in\{\pm1\}$. Moreover, by \cref{ass:1} (convexity and low rank property), \cref{cor:convexity} and \cref{lem:c.2}, the surrogate model satisfies $\Cosim(P_2\widehat\w',P_2(\w(0)-\w(T\eta)))>-G_2>-1$. Therefore the only possibility is
\begin{equation}
\langle P_2\widehat\w', P_2(\w(0)-\w(T\eta))\rangle=1,
\end{equation}
which makes $P_2\widehat\w'$ an ideal surrogate model for the projected gradient flow.
} 

{Next we identify a surrogate model $\widehat\w$ for the discrete GD path which is related to $\widehat \w'$. According to the truncation error of the Euler method \citep{atkinson1991}, it holds that $||P_2\w(T\eta)-P_2\w_T||=O(\eta)$. Thus we expect that the GD curve can well approximate the gradient flow. We can then upper bound the angle $\theta_1$ between $P_2(\w_T-\w_0)$ and $P_2(\w(T\eta)-\w_0)$ by $\arcsin\left(\frac{||P_2(\w(T\eta)-\w_T)||}{||P_2(\w_T-\w_0)||}\right)$. Finally, we can identify a $\widehat \w$ as an interpolation of $\w_0$ and $\w_T$ and estimate $\Cosim(P_2\widehat \w,P_2(\w_0-\w_T))$ using $\widehat \w'$ and our assumptions (smoothness, strong convexity). This defines the error 
\begin{equation}\label{eq:trun}
\Delta_{trun}:=1-\Cosim(P_2\widehat \w,P_2(\w_0-\w_T))\ge0,
\end{equation}
which is a function of $G_2,\eta,\beta,L$ and the loss. This quantity is expected to be close to 0 if the GD curve approximates the gradient flow well. Since this final step is very similar to the proof of Theorem \ref{thm:SGD} and no extra parameters would be required, we do not provide the details for brevity.}

Now we estimate $\mathcal{L}_{sim}(\widehat\w,\w_{0}-\w_T)$. By the identity $\langle\nabla\widehat \w,\bm w_T-\bm w_0 \rangle = \langle P_2\nabla\widehat \w,P_2(\bm w_T-\bm w_0) \rangle+\langle P_2^\perp\nabla\widehat \w,P_2^\perp(\bm w_T-\bm w_0) \rangle$ and assumption \ref{ass:1}, it holds that
\begin{align}
     1-\mathcal{L}_{sim}(\widehat\w,\w_{0}-\w_T)&=\frac{\langle \nabla\widehat \w,\w_0-\w_T\rangle}{||\nabla\widehat \w||||\w_T-\w_0||}\\
     &\ge\frac{|\langle P_2\nabla\widehat \w,P_2(\bm w_T-\bm w_0) \rangle|}{||\nabla\widehat \w||||\w_T-\w_0||}-\frac{|\langle P_2^\perp\nabla\widehat \w,P_2^\perp(\bm w_T-\bm w_0) \rangle|}{||\nabla\widehat \w||||\w_t-\w_0||}\\
      &\overset{}{\ge} (1-\Delta_{trun})\frac{(1-G_2^2)||\nabla\widehat \w||||P_2(\w_T-\w_0)||}{||\nabla\widehat \w||||\w_T-\w_0||}-\frac{||P_2^
     \perp\nabla\widehat \w||}{||\nabla\widehat \w||}\frac{||P_2^\perp(\w_T-\w_0)||}{||\w_T-\w_0||} \\
     &= (1-\Delta_{trun})(1-G_2^2)\frac{||P_2(\w_T-\w_0)||}{||\w_T-\w_0||}-G_2\frac{||P_2^\perp(\w_T-\w_0)||}{||\w_T-\w_0||}\\
     &\ge (1-G_2^2)\frac{||P_2(\w_T-\w_0)||}{||\w_T-\w_0||}-G_2\frac{||P_2^\perp(\w_T-\w_0)||}{||\w_T-\w_0||} - \Delta_{trun}
\end{align}
Next we control the magnitude of $||P_2(\w_T-\w_0)||$ and $||P_2^\perp(\w_T-\w_0)||$. By Equation (4.3) of \citet{Bottou2018}, the Lipchitzness of the gradient (also known as $\beta$-smoothness of the loss) implies that for $\forall \bm w, \bm v \in \mathbb R^p$, it holds that
{{\begin{equation}\label{property:betasmooth}
   \ell(\w)\le \ell(\bm v)+\langle \nabla\bm v, \w-\bm v\rangle+\frac{\beta}{2}||\w-\bm v||^2.
\end{equation}}}
If we let $\w=\w_{t+1}$ and $\bm v = \w_t$ and use the fact that $||\w_{t+1}-\w_t||^2=\eta^2||\nw_t||^2$, we have
\begin{equation}\label{property:betasmooth2}
     0\le\eta\Big(1-\frac{\beta\eta}{2}\Big)||\nw_t||^2\le \ell(\w_t)-\ell(\w_{t+1}).
\end{equation}
Then, by definition we have for $\forall\w_t,\w_{t+1}$, $0\le t\le T-1$,
\begin{align}
  {||P_2^\perp(\w_{t+1}-\w_t)||^2} = {\eta^2||P_2^\perp\nw_t||^2}\le{\eta^2G_2^2||\nw_t||^2}\le {G^2(\eta,\beta)(\ell(\w_t)-\ell(\w_{t+1}))},
\end{align}
so that by triangular and Cauchy-Schwartz inequality we have
\begin{align}
    \frac{||P_2^\perp(\w_T-\w_0)||^2}{||\w_T-\w_0||^2}&\le  \frac{(\sum_{t=0}^{T-1}||P_2^\perp(\w_{t+1}-\w_t)||)^2}{||\w_T-\w_0||^2}\\
    &\le \frac{T\sum_{t=0}^{T-1}||P_2^\perp(\w_{t+1}-\w_t)||^2}{||\w_T-\w_0||^2}\\
    &\le \frac{TG^2(\eta,\beta)(\ell(\w_0)-\ell(\w_T))}{||\w_T-\w_0||^2}\\
    &= \frac{TG^2(\eta,\beta)(\ell(\w_0)-\ell(\w_T))^2}{||\w_T-\w_0||^2(\ell(\w_0)-\ell(\w_T))}\\
    &\le\frac{TG^2(\eta,\beta)L^2}{\ell(\w_0)-\ell(\w_T)}\label{eq:p2perp_ratio}.
\end{align}
Therefore we have
\begin{equation}\label{eq:p2_ratio}
\frac{||P_2(\w_T-\w_0)||}{||\w_T-\w_0||}=\sqrt{1-\frac{||P_2^\perp(\w_T-\w_0)||^2}{||\w_T-\w_0||^2}}\ge 1-{\frac{TG^2(\eta,\beta)L^2}{\ell(\w_0)-\ell(\w_T)}},
\end{equation}
{regardless of the magnitude of \cref{eq:p2perp_ratio}}. Summing up everything we finally conclude that
\begin{align}
1-\mathcal{L}_{sim}(\w_{0}-\w_T,\widehat\w)&\ge (1-G_2^2)\left(1-{\frac{TG^2(\eta,\beta)L^2}{\ell(\w_0)-\ell(\w_T)}}\right)-G_2\sqrt{\frac{TG^2(\eta,\beta)L^2}{\ell(\w_0)-\ell(\w_T)}}-\Delta_{trun}\\
&\ge1-G_2^2-2G_2\sqrt{{\frac{TG^2(\eta,\beta)L^2}{\ell(\w_0)-\ell(\w_T)}}}{-\frac{TG^2(\eta,\beta)L^2}{\ell(\w_0)-\ell(\w_T)}-\Delta_{trun}}\\
&{\ge 1-\left(G_2+\sqrt{{\frac{TG^2(\eta,\beta)L^2}{\ell(\w_0)-\ell(\w_T)}}}\right)^2-\Delta_{trun}}.
\end{align}
\end{proof}
\subsection{Proof of Theorem \ref{thm:SGD}}\label{sec:app:proof:SGD}
Now we try to prove a general result for SGD. The proof consists of 2 steps. In the first step we extend the Proposition \ref{prop:2dspace} to the SGD case. Although we may not be able to find a surrogate model that is parallel to the model update due to stochasticity and discretization, we show that the cosine similarity could be large. Then for high-dimensional space, we control $||P_2^\perp(\w_T-\w_0)||^2/||\w_T-\w_0||^2$ so that the model update on $V_2$ is dominating. As a result, the overall cosine similarity is large. 

In the following, we let the stochastic gradient at step $t$ be denoted as $g_t:=\nw_t+\bm\epsilon_t$, where $\bm\epsilon_t$ is the gradient noise which has mean zero. For notational simplicity, we write $\ell(\w)=\maE_D[\ell(\w,D)]$. All the assumptions in \Cref{thm:SGD} are applied here.

\textbf{Step 1:} 

We first control the distance between the weights achieved by GD and SGD (both start from $\w_0$) via the gradient noises.
\begin{lemma}\label{lem:distance}
Consider SGD $\{\w_t\}$ and GD $\{\w_t'\}$ on some $\beta$-smooth loss $\ell$. If both paths have step size $\eta$, $T$ steps and start from $\w_0$, we have
\begin{equation}
||\w_T-\w_{T}'||\le\eta\left|\left|\sum_{t=0}^{T-1}\e_{t}\right|\right|+\sum_{s=0}^{T-2}\eta^2\beta(1+\eta\beta)^{T-2-s}\left|\left|\sum_{t=0}^s\bm\epsilon_t\right|\right|.    
\end{equation}

\end{lemma}
\begin{proof}
By definition we have $\w_1-\w_1'=-\eta(\bm g_0-\nw_0)=-\eta\e_0$ and more generally,
\begin{align}
\w_{t+1}-\w_{t+1}'&=\w_{t}-\w_{t}'-\eta(\bm g_{t}-\nw_{t}')\\
&=\w_{t}-\w_{t}'-\eta(\bm g_{t}-\nw_{t})-\eta(\nw_{t}-\nw_{t}')\\
&=\w_{t}-\w_{t}'-\eta\bm\epsilon_{t}-\eta(\nw_{t}-\nw_{t}')
\end{align}
Thus, by summing the equations for $t=0$ and $t=1$ and using the smoothness condition, we have
\begin{align}
\w_2-\w_2'&=-\sum_{t=0}^1\eta\bm\epsilon_{t}-\eta(\nw_1-\nw_1')\\
\Rightarrow ||\w_2-\w_2'||&\le\eta\left|\left|\sum_{t=0}^1\bm\epsilon_{t}\right|\right|+\eta\beta||\w_1-\w_1'||\\
||\w_2-\w_2'||&\le\eta\left|\left|\sum_{t=0}^1\bm\epsilon_{t}\right|\right|+\eta^2\beta||\bm\epsilon_0||
\end{align}
Therefore by induction, we finally conclude that
\begin{align}
||\w_T-\w_T'||&\le\eta\left|\left|\sum_{t=0}^{T-1}\e_{t}\right|\right|+\sum_{t=0}^{T-1}\eta\beta||\w_t-\w_t'||\\
&\le \eta\left|\left|\sum_{t=0}^{T-1}\e_{t}\right|\right|+{\eta^2\beta}\left|\left|\sum_{t=0}^{T-2}\e_{t}\right|\right|+\cdots+\eta^2\beta(1+\eta\beta)^{T-2}\left|\left|\e_0\right|\right|\\
&\le\eta\left|\left|\sum_{t=0}^{T-1}\e_{t}\right|\right|+\sum_{s=0}^{T-2}\eta^2\beta(1+\eta\beta)^{T-2-s}\left|\left|\sum_{t=0}^s\bm\epsilon_t\right|\right|.
\end{align}
\end{proof}

Since the angle $\theta_1$ between $\w_T-\w_0$ and $\w_T'-\w_0$ satisfies 
\begin{equation}
\sin(\theta_1)\le \frac{||\w_T-\w_T'||}{||\w_T-\w_0||},  
\end{equation}
the key is to control $||\sum_{t=0}^s\e_t||/||\w_T-\w_0||$ . Thanks to the matrix Bernstein inequality below, we can infer with how much probability, $||\sum_{t=0}^s\e_t||/||\w_T-\w_0||$ is small. With a slight abuse of notation, we also write $||A||$ to denote the matrix norm of a $d_1\times d_2$ matrix $A$, i.e.\ for any $\bm x\in \mathbb R^{d_2}$,
\begin{equation}
||A||:=\max_{||\bm x||=1}\frac{||A\bm x||}{||\bm x||}.
\end{equation}

\begin{theorem}[Theorem 1.6.2 of \citet{matrix_ineq}]\label{matrixBernstein}
Let $\bm S_1,\dots,\boldsymbol S_n$ be independent, centered random matrices with common dimension $d_1\times d_2$, and assume that $\maE[\boldsymbol S_i]=0$ and $||\boldsymbol S_i||\le A$. Let $\boldsymbol Z = \sum_{k=1}^n\boldsymbol S_i$ and define
\begin{equation}
v(\boldsymbol Z)=\max\left\{\left|\left|\sum_{i=1}^n\maE[\boldsymbol S_i\boldsymbol S_i^\top]\right|\right|,\left|\left|\sum_{i=1}^n\maE[\boldsymbol S_i^\top\boldsymbol S_i]\right|\right|\right\}.   
\end{equation}

Then
\begin{equation}
P[||\bm Z||\ge t]\le(d_1+d_2)\exp\Big(\frac{-t^2/2}{v(\boldsymbol Z)+At/3}\Big)\;\;\;\text{ for all }t\ge 0.   
\end{equation}

\end{theorem}

The result is the following.
\begin{lemma}\label{lem:probability}
Consider a stochastic gradient path $\{\w_t\}$ and gradient descent path $\{\w_t'\}$ with $T$ steps on an $L$-Lipschitz loss $\ell$ on $\mathbb R^2$. For any $c>0$ (e.g.\ the choice of $c$ in \Cref{thm:SGD}), we define 
\begin{equation}
r:=\min\left\{\frac{c^2(\ell(\w_0)-\ell(\w_T'))^2}{4TL^2E_{max}^2}, \frac{c(\ell(\w_0)-\ell(\w_T'))}{4LE_{max}}\right\},    
\end{equation}

Then, with probability at least $1-3\exp(-r)$, we have
\begin{equation}\label{eq:noiseub}
\left|\left|\sum_{t=0}^{T-1}\e_t\right|\right|\le \frac{c}{L}(\ell(\w_0)-\ell(\w_T'))\le c||\w_T'-\w_0||.
\end{equation}

\end{lemma}
\begin{proof}
Since the gradient flow path is deterministic, we can always compare the accumulative noise with $\{\w_t'\}$. It's clear that $\e_t$ satisfies the conditions in Theorem \ref{matrixBernstein}. Moreover, we have in the notation of their theorem that 
\begin{equation}
v\left(\sum_{t=0}^{T-1}\e_t\right)\le T\max_{t=0,\dots,T-1}\{||\e_t\e_t^\top||,||\e_t^\top\e_t||\}\le TE_{max}^2.  
\end{equation}

Therefore, by the matrix Bernstein inequality we have that
\begin{align}
P\left(\left|\left|\sum_{t=0}^{T-1}\e_t\right|\right|\ge c||\w_T'-\w_0||\right)&\le P\left(\left|\left|\sum_{t=0}^{T-1}\e_t\right|\right|\ge \frac{c}{L}(\ell(\w_0)-\ell(\w_T'))\right)\\
&\le 3\exp\left(-\frac{c^2(\ell(\w_0)-\ell(\w_T'))^2}{2L^2(TE^2_{max}+cE_{max}(\ell(\w_0)-\ell(\w_T'))/3L)}\right)\\
&\le 3\exp(-r).
\end{align}
\end{proof}
\begin{remark}\label{remark:noise}
In high dimensional case, the $E_{max}$ here turns to be $\max||P_2\e||$. Thus, if $||P_2\e||\ll||\e||$, e.g.\ isotropic noise, the probability is close to 1. On the other hand, if $\e_t$ are mainly concentrated on some $k$-dimensional subspace of $\mathbb R^p$, i.e.\ $||P_k\e||\approx||\e||$ for some projection matrix $P_k$, we can still prove \cref{eq:noiseub} (up to some small constant) for $\e_t\in\mathbb R^p$ with probability at least $1-(k+1)\exp(-r)$.
\end{remark}

Now, combining all the previous lemmas we have
\begin{corollary}\label{cor:theta1}
Under the conditions of Lemma \ref{lem:distance} and Lemma \ref{lem:probability}, if $c<{1}/{\eta(1+\eta\beta)^{T-1}}$, we have with probability at least $1-3T\exp(-r)$ that 
\begin{equation}
\theta_1\le\text{arcsin}\left(c\eta(1+\eta\beta)^{T-1}\right),  
\end{equation}
moreover, we have for the SGD and GD that
\begin{equation}\label{eq:GDvsSGDnorm}
\left(1-c\eta(1+\eta\beta)^{T-1}\right)||\w_T'-\w_0||\le||\w_T-\w_0||\le\left(1+c\eta(1+\eta\beta)^{T-1}\right)||\w_T'-\w_0||.
\end{equation}

\end{corollary}
\begin{proof}
By a uniform bound and Lemma \ref{lem:probability} we have that
\begin{align}
 P\left(\forall0\le  s\le T-1, ||\sum_{t=0}^s\e_t||\ge c||\w_T'-\w_0||\right)&\le\sum_{s=0}^{T-1}P\left(||\sum_{t=0}^s\e_t||\ge c||\w_T'-\w_0||\right)\\
 &\le 1-3T\exp(-r)
\end{align}
By Lemma \ref{lem:distance} we have that with probability at least $1-3T\exp(-r)$,
\begin{align}
||\w_T-\w_T'||&\le\eta\left|\left|\sum_{t=0}^{T-1}\e_{t}\right|\right|+\sum_{s=0}^{T-2}\eta^2\beta(1+\eta\beta)^{T-2-s}\left|\left|\sum_{t=0}^s\bm\epsilon_t\right|\right|\\
&\le \eta c||\w_T'-\w_0||+\sum_{s=0}^{T-2}\eta^2\beta(1+\eta\beta)^{T-2-s}c||\w_T'-\w_0||\\
&\le c||\w_T'-\w_0||\left(\eta+\eta^2\beta\frac{(1+\eta\beta)^{T-1}-1}{\eta\beta}\right)\\
&=c\eta(1+\eta\beta)^{T-1}||\w_T'-\w_0||.
\end{align}
The second claim follows from the triangle inequality.
\end{proof}
Intuitively, \Cref{cor:theta1} shows that if the gradient noise has a small magnitude, there is a large probability that $\theta_1$ is small. 

Now, by Proposition \ref{prop:2dspace} {and the proof of \cref{thm:GD}}, we can find a surrogate model $\widehat\w'$ on the connected line between $\w_T'$ and $\w_0$ such that {$\Cosim(\nabla\widehat \w' \w_0-\w_T')=1-\Delta_{trun}\approx1$. As discussed in the main text, we omit $\Delta_{trun}$ in this theorem.} It's clear that we can find a surrogate model $\widehat\w$ on the connected line between $\w_T$ and $\w_0$ such that $||\widehat\w-\widehat\w'||\le||\w_T-\w_T'||$ and therefore by the smoothness of the true loss function we have,
\begin{equation}
||\nabla \widehat\w-\nabla\widehat \w'||\le\beta||\widehat\w-\widehat\w'||\le\beta||\w_T-\w_T'||\le c\eta\beta(1+\eta\beta)^{T-1}||\w_T'-\w_0||
\end{equation}
with high probability. Let $\theta_2$ be the angle between $\nabla\widehat\w$ and $\nabla\widehat \w'$, we have that with probability at least $1-3T\exp(-r)$ that 
\begin{align}
\sin^2(\theta_2)&\le\frac{||\nabla\widehat\w-\nabla\widehat \w'||^2}{||\nabla\widehat \w'||^2}\le\left(c\eta\beta(1+\eta\beta)^{T-1}\right)^2\frac{||\w_T'-\w_0||^2}{||\nabla\widehat \w'||^2}\\
&\overset{(i)}{\le}\left(c\eta\beta(1+\eta\beta)^{T-1}\right)^2\frac{(\sum_{t=0}^{T-1}||\w_{t+1}'-\w_t'||)^2}{||\nabla\widehat \w'||^2}\\
&\overset{(ii)}{\le}\left(c\eta\beta(1+\eta\beta)^{T-1}\right)^2\frac{\eta^2T\sum_{t=0}^{T-1}||\nw_t'||^2}{||\nabla\widehat \w'||^2}\\
&\le \left(c\eta^2\beta(1+\eta\beta)^{T-1}\sqrt{T}\right)^2\frac{\sum_{t=0}^{T-1}||\nw_t'||^2}{||\nabla\widehat \w'||^2}\\
&\overset{(iii)}{\le} \left(c\eta^2\beta(1+\eta\beta)^{T-1}\sqrt{T}\right)^2\frac{2(\ell(\w_0)-\ell(\w_T'))}{\eta||\nabla\widehat \w'||^2}\label{eq:sgd2d1}\\
&\overset{}{\le} \left(\frac{c\eta^{3/2}\beta(1+\eta\beta)^{T-1}\sqrt{T}}{\sqrt{\gamma}}\right)^2\frac{\ell(\w_0)-\ell(\w_T')}{\ell(\w_T')-\ell(\w^*)}\label{eq:sgd2d2}
\end{align}
where (i) is the triangular inequality, (ii) is the Cauchy-Schwartz inequality, (iii) follows from Theorem 4.8 of \citet{Bottou2018} and the last inequality follows from {the definition of $\gamma$-strongly convex and \cref{cor:convexity}, where $\w^*$ is the global minimum}. 

Let $\theta$ be the angle between $\nabla \widehat\w$ and $\w_0-\w_T$. Without loss of generality, we assume both $\theta_1,\theta_2>0$, as otherwise $\theta$ is bounded by $\theta_1$ or $\theta_2$ and the result is trivial. Then we have with probability at least $1-3T\exp(-r)$
\begin{align}
\sin^2(\theta)&=\sin^2(\theta_1+\theta_2)\le({|\sin(\theta_1)|}+{|\sin(\theta_2)|})^2\\
&\le2\sin^2(\theta_1)+2\sin^2(\theta_2)\\
&\le\left(\frac{c\eta^{3/2}\beta(1+\eta\beta)^{T-1}\sqrt{2T}}{\sqrt{\gamma}}\right)^2\frac{\ell(\w_0)-\ell(\w_T')}{\ell(\w_T')-\ell(\w^*)}+\left(\sqrt{2}c\eta(1+\eta\beta)^{T-1}\right)^2,
\end{align}
So that
\begin{align}
    \mathcal{L}_{sim}(\w_0-\w_T,\nabla\widehat\w)&=1-\cos(\theta)\le1-\cos^2(\theta)\nonumber\\
    &\le\sin^2(\theta)\le\left(\frac{c\eta^{3/2}\beta(1+\eta\beta)^{T-1}\sqrt{2T}}{\sqrt{\gamma}}\right)^2\frac{\ell(\w_0)-\ell(\w_T')}{\ell(\w_T')-\ell(\w^*)}+\left(\sqrt{2}c\eta(1+\eta\beta)^{T-1}\right)^2.\label{eq:2dtheta}
\end{align}
The above equation finishes the estimates in the 2D case.

\allowdisplaybreaks
\textbf{Step 2:}
Finally, we prove Theorem \ref{thm:SGD}.
\begin{proof}[Proof of Theorem \ref{thm:SGD}]
{We shall apply the preceding results to the projected SGD on $V_2$. The only difference is that we have another factor $\frac{1}{1-G^2_2}$ in \cref{eq:sgd2d2} since the denominator in \cref{eq:sgd2d1} is replaced by $||P_2\nabla\widehat\w'||$.}
Thus we have with probability at least $1-3T\exp(-r)$ that $||\sum_{t=0}^{T-1}P_2\e_t||\le c||P_2(\w_T'-\w_0)||$ (or {even higher probability} since it could happen that $||P_2\e_t||$ is much less than $||\e_t||$, see \Cref{remark:noise}), so that on this event,
\begin{align}
    \frac{||P_2^\perp(\w_T-\w_0)||}{||\w_T-\w_0||}
    &\le \frac{||P_2^\perp(\w_T'-\w_0)||}{||\w_T-\w_0||}+ \frac{||P_2^\perp(\w_T-\w_T')||}{||\w_T-\w_0||}\\
    &\le  \frac{||P_2^\perp(\w_T'-\w_0)||}{||P_2(\w_T-\w_0)||}+ \frac{||P_2^\perp(\w_T-\w_T')||}{||\w_T-\w_0||}\\
    &\le  C_\eta\frac{||P_2^\perp(\w_T'-\w_0)||}{||P_2(\w_T'-\w_0)||}+ \frac{||\w_T-\w_T'||}{||P_2(\w_T-\w_0)||}\\
    &\le C_\eta\frac{C_{GD}}{1-C_{GD}}+ C_\eta\frac{||\w_T-\w_T'||}{||P_2(\w_T'-\w_0)||}\\
    &\le  C_\eta\frac{C_{GD}}{1-C_{GD}}+ \frac{C_\eta}{1-C_{GD}}\frac{||\w_T-\w_T'||}{||\w_T'-\w_0||}\\
    &\le\frac{C_\eta}{1-C_{GD}}\left(C_{GD}+\frac{L||\w_T-\w_T'||}{\ell(\w_0)-\ell(\w_T')}\right)\\
    &\le\frac{C_\eta}{1-C_{GD}}\left(C_{GD}+\frac{LTE_{max}\eta(1+\eta\beta)^{T-1}}{\ell(\w_0)-\ell(\w_T')}\right).
\end{align}
Here, $C_\eta=\frac{1}{1-\sqrt{\eta}}$ is the constant follows from \cref{eq:GDvsSGDnorm} and the choice of $c$ in the theorem, which is close to 1. The 4th and 5th inequalities follow from \cref{eq:p2perp_ratio} and \cref{eq:p2_ratio}. The last inequality follows from \Cref{lem:distance} and union bound.

In total, let $\widehat w$ be the surrogate model such that the angle $\theta$ between $P_2\nabla\widehat \w$ and $P_2(\w_0-\w_T)$ satisfies \cref{eq:2dtheta}, we have with probability at least $1-3T\exp(-r)$ that,
\begin{align}
\mathcal{L}_{sim}(\w_{0}-\w_T,\widehat\w)&=1-\frac{\langle \nabla\widehat \w,\w_T-\w_0\rangle}{||\nabla\widehat \w||||\w_T-\w_0||}\\
     &\le1-\left(\frac{|\langle P_2\nabla\widehat \w,P_2(\bm w_0-\bm w_T) \rangle|}{||\nabla\widehat \w||||\w_T-\w_0||}-\frac{|\langle P_2^\perp\nabla\widehat \w,P_2^\perp(\bm w_T-\bm w_0) \rangle|}{||\nabla\widehat \w||||\w_T-\w_0||}\right)\\
     &\le1-\left(\cos(\theta)\frac{||P_2\nabla\widehat \w||||P_2(\w_T-\w_0)||}{||\nabla\widehat \w||||\w_T-\w_0||}-\frac{||P_2^
     \perp\nabla\widehat \w||}{||\nabla\widehat \w||}\frac{||P_2^\perp(\w_T-\w_0)||}{||\w_T-\w_0||}\right) \\
     &\le1-\left[(1-G_2^2)\cos(\theta)\frac{||P_2(\w_T-\w_0)||}{||\w_T-\w_0||}-G_2\frac{||P_2^\perp(\w_T-\w_0)||}{||\w_T-\w_0||}\right] \\
     &\le1-\left[(1-G_2^2)\cos(\theta)\left(1-\frac{||P_2^\perp(\w_T-\w_0)||}{||\w_t-\w_0||}\right)-G_2\frac{||P_2^\perp(\w_T-\w_0)||}{||\w_T-\w_0||}\right] \\
     &\le1-(1-G_2^2)\cos^2(\theta)+[(1-G_2^2)\cos(\theta)+G_2]\frac{||P_2^\perp(\w_T-\w_0)||}{||\w_T-\w_0||}
     \\
    &\le1-(1-G_2^2)\cos^2(\theta)+(1+G_2)\frac{||P_2^\perp(\w_T-\w_0)||}{||\w_T-\w_0||}
     \\&\le G_2^2+\sin^2(\theta)+(1+G_2)\frac{||P_2^\perp(\w_t-\w_0)||}{||\w_T-\w_0||}\\
     &\le G_2^2+{\frac{1}{1-G^2_2}}\left(\frac{c\eta^{3/2}\beta(1+\eta\beta)^{T-1}\sqrt{2T}}{\sqrt{\gamma}}\right)^2\frac{\ell(\w_0)-\ell(\w_T')}{\ell(\w_T')-\ell(\w^*)}+\left(\sqrt{2}c\eta(1+\eta\beta)^{T-1}\right)^2\\
     &\quad+ \frac{C_\eta(1+G_2)}{1-C_{GD}}\left(C_{GD}+\frac{LTE_{max}\eta(1+\eta\beta)^{T-1}}{\ell(\w_0)-\ell(\w_T')}\right).
\end{align}
By letting $c={1/\sqrt{\eta}(1+\eta\beta)^{T-1}}$ we prove the desired result.
\end{proof}

\vfill
\newpage
\section{Experiment Setup}
\label{sec:app:experiment}
\paragraph{Dataset.} FEMNIST is proposed in the FL bechmark \textit{LEAF}~\cite{caldas2018leaf}. This dataset is adapted from EMNIST~\cite{cohen2017emnist}. which consists of $28 \times 28$ grayscale images with 62 classes of hand-written digits and characters. CIFAR100~\cite{cifar10} consists of $32 \times 32$ RGB images with 100 classes of natural images. 

\paragraph{Federated Learning.} In this work, we consider a typical FL framework: Federated Averaging (\verb|FedAvg|)~\cite{mcmahan17a_fedavg} as the attack scenario. The algorithm of \verb|FedAvg| is given in \Cref{alg:fedavg}. To consider the different situations of local training in FL, we need to change the number of training epochs $E$, batch size $B$, local data size $N$, and learning rate $\eta$. To improve the clarity of comparison, we follow the previous work~\cite{dimitrov2022data} and fix the learning rate  $\eta = 0.004$ for the two CNNs. This learning rate is also suggested by the benchmark \textit{LEAF} for a good training result in FL using the same network. Then in \Cref{tab:psnr}, we conduct experiments with different batch sizes $B$, epochs $E$, and local data size $N$ to investigate the influence of local training for different attack approaches. 

In \Cref{fig:g2}, \Cref{fig:cos}, \Cref{fig:g2_trained}, \Cref{fig:cos_trained}, and \Cref{tab:trained_psnr}, we also consider clients joining the FL group at different communication rounds, such that the network sent to the clients is at different training stages. To obtain the trained FL network, we generate the FL clients using the data splitting tool provided by \textit{LEAF}. The dataset is FEMNIST. Local data is non-identically distributed across clients and the FL group has approximately 40.000 data points in total. During FL, every client trains the network locally with batch size $B=50$, epochs $E=10$, and learning rate $\eta=0.004$. To simulate the straggler issue in FL, every client has the possibility of 0.1 sending the trained weights back to the server. We save the trained network at the end of communication rounds 5, 10, and 20 to conduct attack experiments. These networks can achieve test accuracy: $\sim19\%$, $\sim28\%$, and $\sim40\%$.
\begin{algorithm}[h]
   \caption{Federated Averaging~\cite{mcmahan17a_fedavg}}
   \label{alg:fedavg}
\begin{algorithmic}[1]
   \STATE {\bfseries Notations:} The $J$ clients are indexed by $j$; $B$ is the batch size, $E$ is the number of local epochs, and $\eta$ is the learning rate.
   \item[]
   \STATE {\bfseries Sever executes:}
   \STATE Initialize $\bm w^0$.
   \FOR{each round $r=0\dots R$}
   \FOR{each client $j=1\dots J$}
   \STATE $\bm w_j^{r+1} \leftarrow \operatorname{ClientUpdate}(\bm w^r)$
   \ENDFOR
   \STATE $\bm w^{r+1} = n_j \bm w_j^{r+1}/ \sum_{j=1}^J n_j$ \label{alg:fedavg:aggregation}
   \ENDFOR
   \STATE {\bfseries ClientUpdate($\bm w_0$):}
   \STATE Set $t=0$.
   \FOR{each epoch $e=0\dots E$}
    \STATE $\mathcal{B} \leftarrow$ Split local data $D$ into batches of size $B$.
   \FOR{batch $b\in\mathcal{B}$}
   \STATE $\bm w_{t+1} = \bm w_t - \eta \nabla \ell(\bm w_t; b)$
   \ENDFOR
   \ENDFOR
\end{algorithmic}
\end{algorithm}
\paragraph{Hyperparameters for attack approaches.}
\verb|DLFA|~\cite{dimitrov2022data} has several variants, we use the one that is reported to always give the best reconstruction results, i.e.\ the full input-reconstruction method including the order-invariant prior $\mathcal{L}_{prior}$. {We adopt the officially released code and recommended hypeparameters that is available at \url{https://github.com/eth-sri/fedavg_leakage}.}

For our \verb|SME| and the base method \verb|IG|, we use the same hyperparameters and adapt them from the work of \verb|IG|~\cite{geiping_inverting_gradient}. We note that due to the similar mechanisms, these two methods achieve the best performance with the same hyperparameters. In particular, we set learning rate $\eta_{\Tilde{D}}=1$ for the optimization of reconstructed data. and learning rate $\eta_\alpha=0.001$ for the optimization of $\alpha$. We set $\lambda = 0.01$ for the total variation prior, and use the Adam optimizer~\cite{kingma2014adam}. Additionally, we bound the reconstructed pixel values within the valid range of image pixel values, this has also been implemented in previous works including \verb|DLFA|. We let the reconstruction run for 1000 iterations.
\vfill

\section{Additional Empirical Evidence}
\label{sec:app:evidence}
According to \Cref{thm:GD} and \Cref{thm:SGD}, in order to find a surrogate model $\hw$ on the connected line achieving small $\mathcal{L}_{sim}(\w_0 - \w_T, \nabla \hw)$, there are two necessary conditions. First, $G_2$ needs to be small such that the low-rank property holds. Second, the magnitude of gradient noise $\{\bm \epsilon_t\}_{t=0}^{T-1}$ needs to be moderate compared with the loss reduction $\ell(\w_0) - \ell(\w_T')$. In \Cref{fig:g2} and \Cref{fig:cos} we have shown for many situations we can find such a proper surrogate model. In this section, we further investigate this problem in more challenging situations.

\begin{figure*}[h!]
    \centering
    \subfloat[\centering $E$=50, $B$=50, $T$=50]{\includegraphics[width=0.23\textwidth]{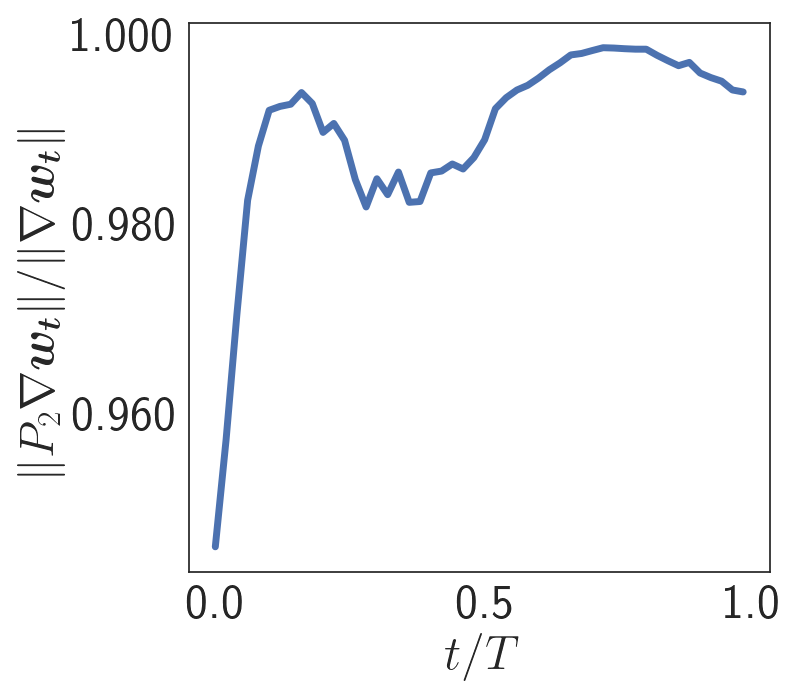}}%
    \subfloat[\centering $E$=25, $B$=25, $T$=50]{\includegraphics[width=0.23\textwidth]{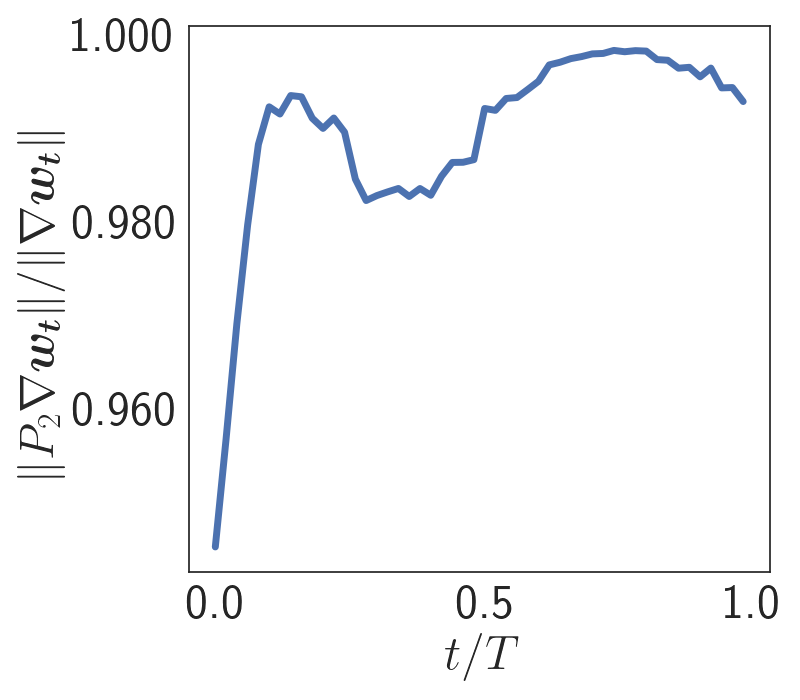}}%
    \subfloat[\centering $E$=10, $B$=10, $T$=50]{\includegraphics[width=0.23\textwidth]{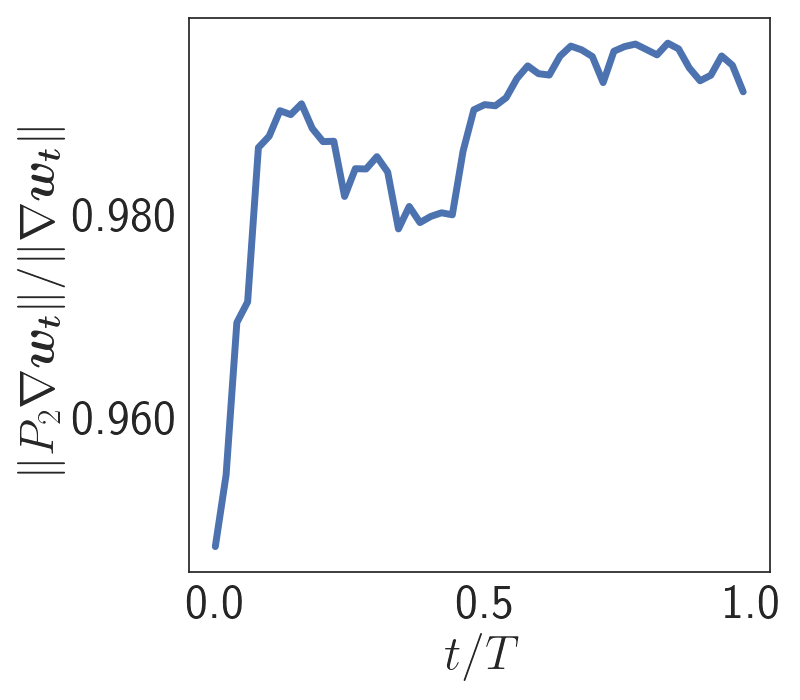}}%
    \subfloat[\centering $E$=5, $B$=5, $T$=50]{\includegraphics[width=0.23\textwidth]{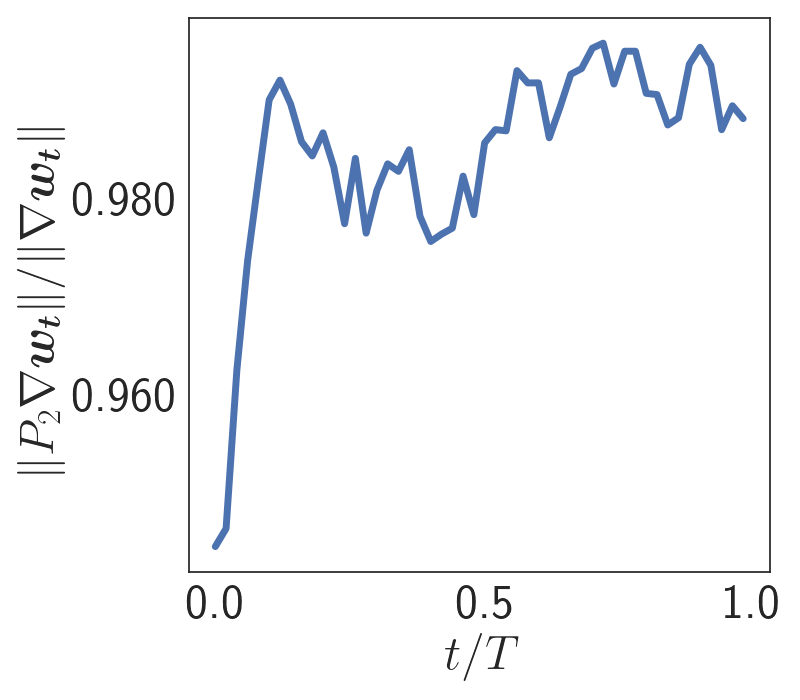}}%
    \caption{Gradient norm ratio $\|P_2\nabla \bm w_t\|/\|\nabla \bm w_t\|$ along the local steps $t=0\dots T-1$. We sample a client with $N = 50$ data points and change the batch size $B$ and epochs $E$, such that there are always $T=50$ steps.}
    \label{fig:g2_sgd}
\end{figure*}

\begin{figure*}[h!]
    \centering
    \subfloat[\centering $E$=50, $B$=50, $T$=50]{\includegraphics[width=0.23\textwidth]{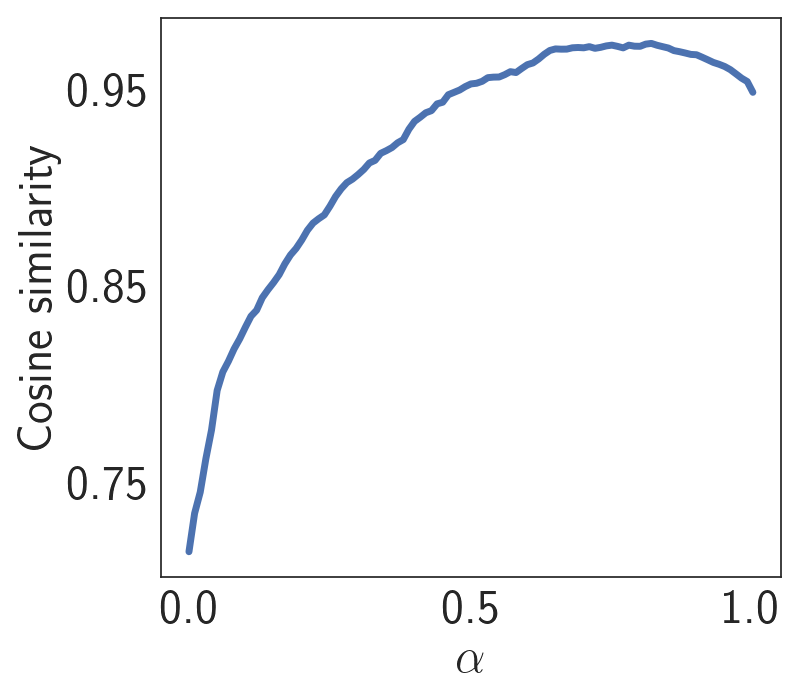}}%
    \subfloat[\centering $E$=25, $B$=25, $T$=50]{\includegraphics[width=0.23\textwidth]{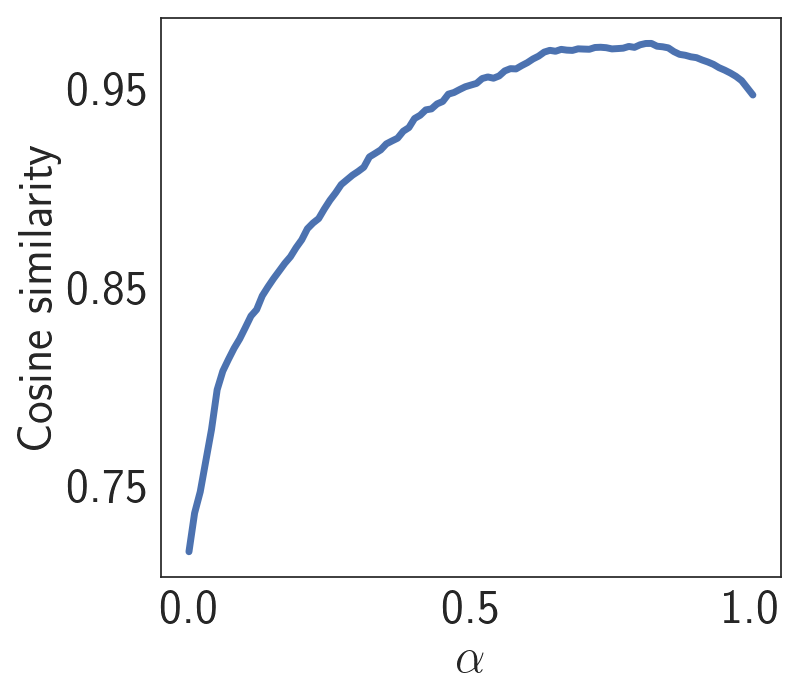}}%
    \subfloat[\centering $E$=10, $B$=10, $T$=50]{\includegraphics[width=0.23\textwidth]{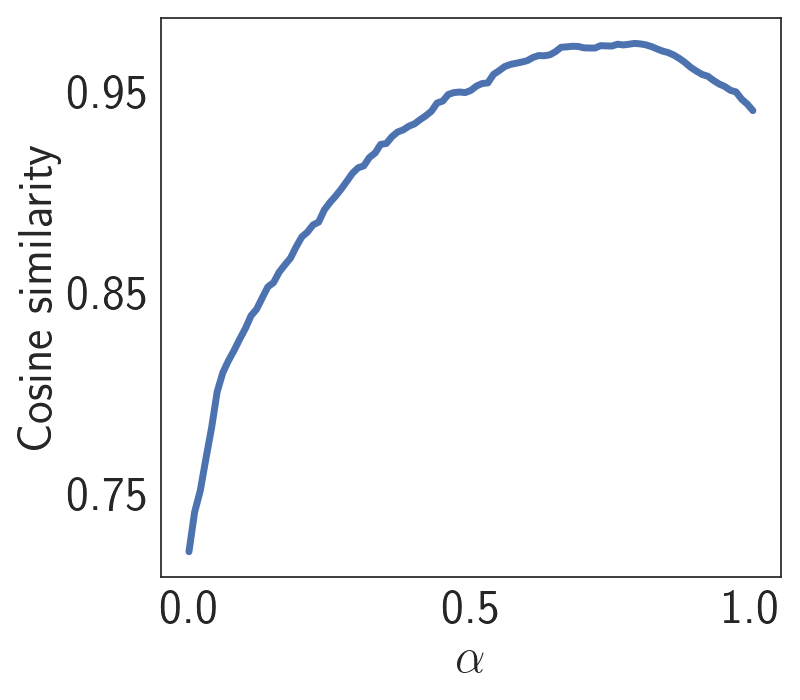}}%
    \subfloat[\centering $E$=5, $B$=5, $T$=50]{\includegraphics[width=0.23\textwidth]{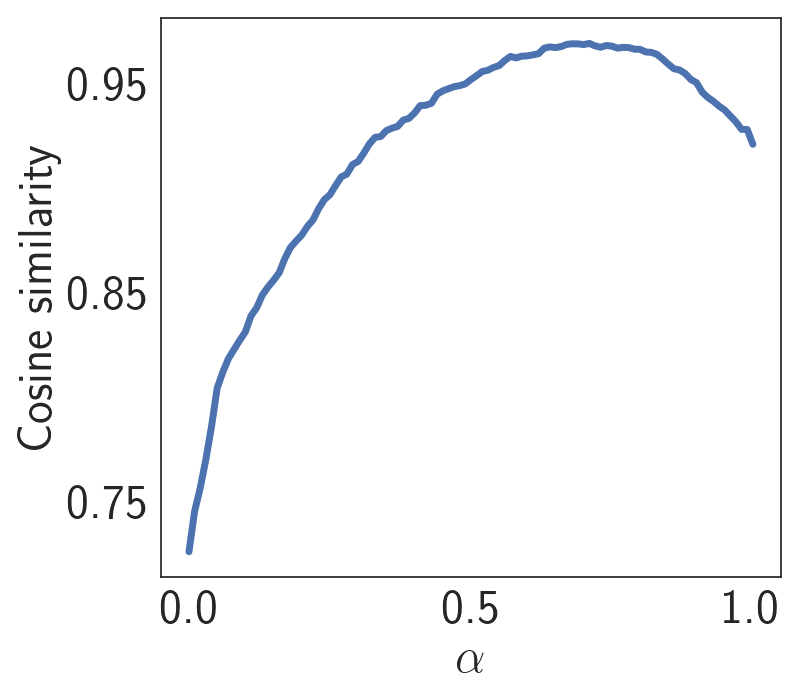}}%
    \caption{Cosine similarity between $\bm w_0 - \bm w_T$ and $\nabla\widehat \w$ vs.\ $\alpha$. The settings are aligned with \Cref{fig:g2_sgd}, please refer to the caption above.}
    \label{fig:cos_sgd}
\end{figure*}

First, we investigate the influence of gradient noise $\bm \epsilon_t$. Since gradient noise is induced by stochastic mini-batch gradient, in \Cref{fig:g2_sgd} and \Cref{fig:cos_sgd} we consider the same local steps $T$ but different batch size $B$. As we can see, when $B=5$ the fluctuation in the gradient norm ratio is more obvious. However, overall the plots of the gradient norm ratio and the cosine similarity are similar. This indicates that the stochastic gradient flow oscillates around the true gradient flow. And the impact of gradient noise $\bm \epsilon_t$ to \verb|SME| is marginal. {Statistical evaluations of multiple samplings are presented in \Cref{tab:stats_sgd}.}
\begin{table*}[h!]
\begin{center}
\begin{small}
\begin{tabular}{ccccc}
\toprule 
&(E=50, B=50, T=50)  &(E=25, B=25, T=50)& (E=10, B=10, T=50) &(E=5, B=5, T=50)\\
\hhline{~----}
$\maE [\min_t \frac{\|P_2\nabla \bm w_t\|}{\|\bm w_t\|}]$
&.977$\pm$.009 &.978$\pm$.007 &.977$\pm$.008 &.973$\pm$.009\\
\midrule
$\maE[\Cosim|\alpha=0]$
&.867$\pm$.041 &.869$\pm$.049 &.872$\pm$.037 &.876$\pm$.041\\
\midrule
$\maE[\max_{\alpha\in[0,1]}\Cosim]$
&.980$\pm$.007 &.978$\pm$.026 &.979$\pm$.007 &.977$\pm$.008\\
\bottomrule
\end{tabular}
\end{small}
\caption{{Statistical evaluations corresponding to \Cref{fig:g2_sgd,fig:cos_sgd}. We repeat the experiment 100 times to compute the mean and standard deviation. $\Cosim$ is a shorthand for the cosine similarity between $\w_0 - \w_T$ and $\nabla \hw$}.}
\label{tab:stats_sgd}
\end{center}
\end{table*}

\clearpage

\begin{figure*}[h!]
    \centering
    \subfloat[\centering $E$=10, $B$=10, $R$=1]{\includegraphics[width=0.23\textwidth]{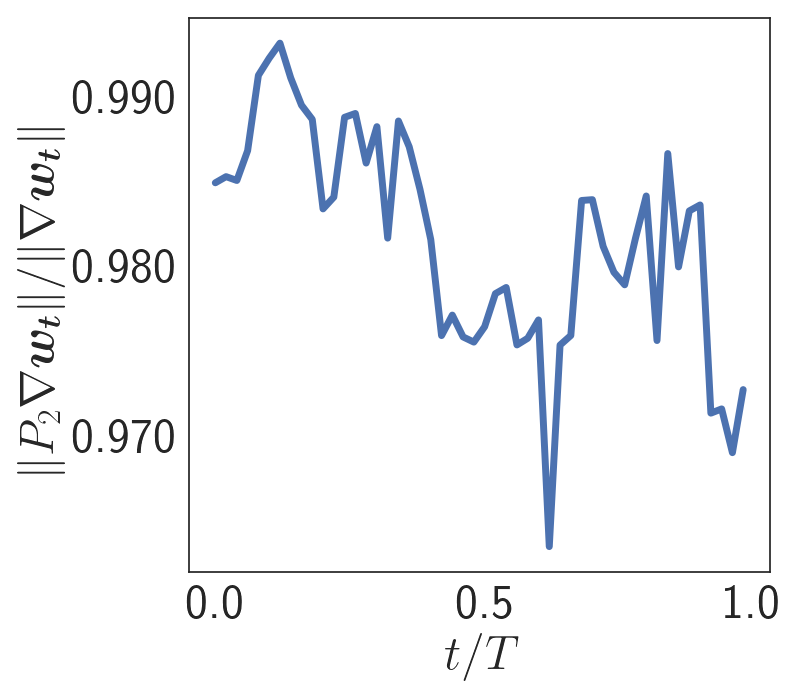}}%
    \subfloat[\centering $E$=10, $B$=10, $R$=5]{\includegraphics[width=0.23\textwidth]{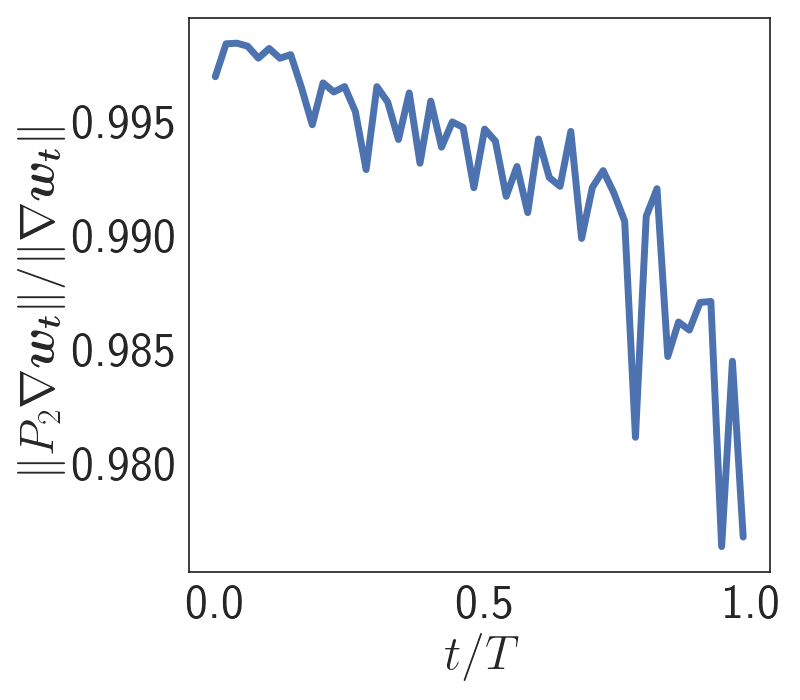}}%
    \subfloat[\centering $E$=10, $B$=10, $R$=10]{\includegraphics[width=0.23\textwidth]{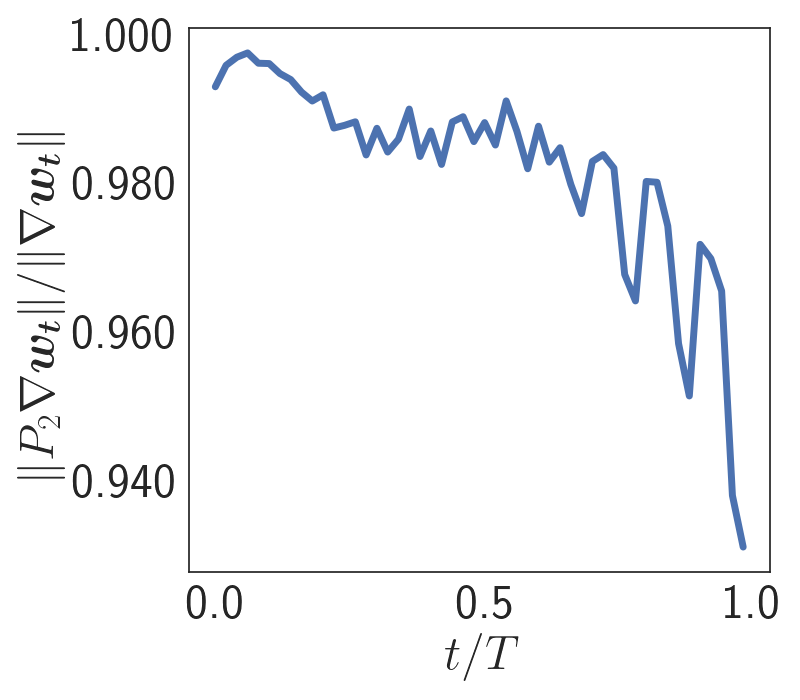}}%
    \subfloat[\centering $E$=10, $B$=10, $R$=20]{\includegraphics[width=0.23\textwidth]{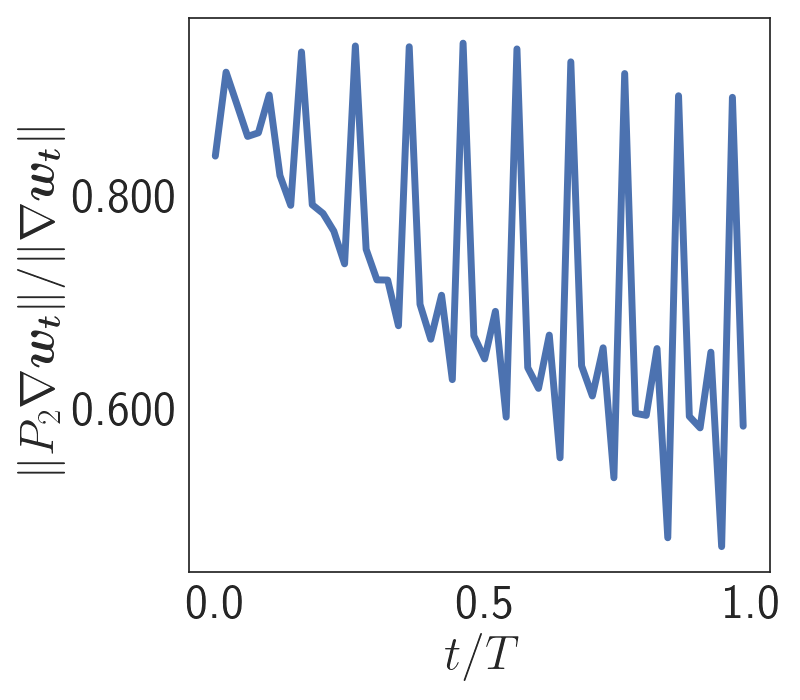}}%
    \caption{Gradient norm ratio $\|P_2\nabla \bm w_t\|/\|\nabla \bm w_t\|$ along the local steps $t=0\dots T-1$. We sample a client with $N = 50$ data points, and set $E=10$, $B=10$, such that for each epoch there are five SGD steps and in total $T=50$ local steps. We change the number of communication rounds $R$, such that the network $\w_0$ received by the clients is at different training stage.}
    \label{fig:g2_trained}
\end{figure*}

\begin{figure*}[h!]
    \centering
    \subfloat[\centering $E$=10, $B$=10, $R$=1]{\includegraphics[width=0.23\textwidth]{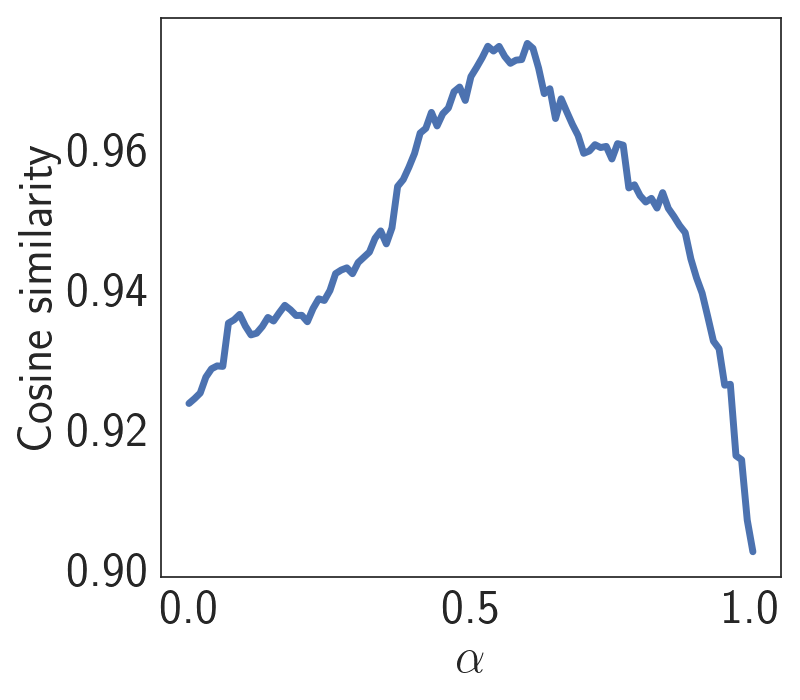}}%
    \subfloat[\centering $E$=10, $B$=10, $R$=5]{\includegraphics[width=0.23\textwidth]{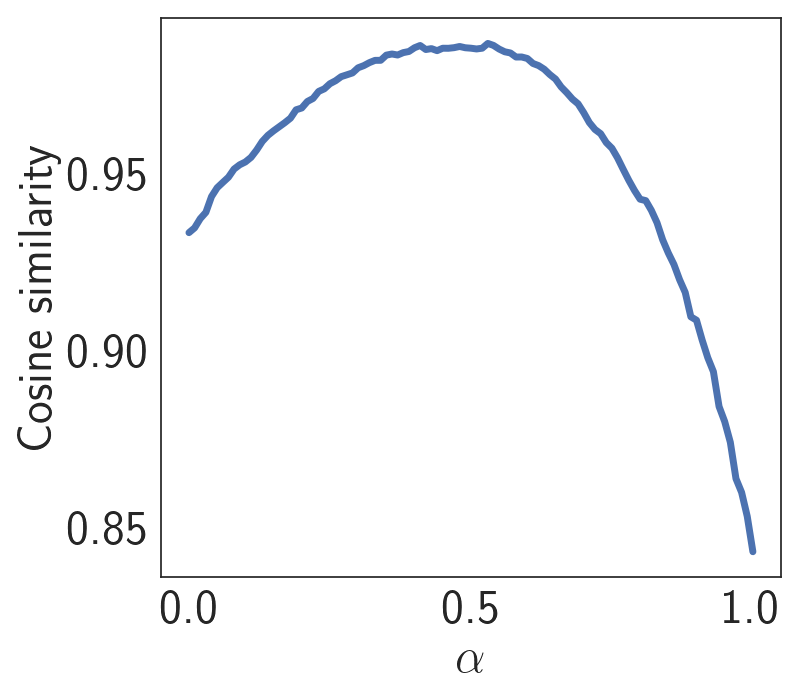}}%
    \subfloat[\centering $E$=10, $B$=10, $R$=10]{\includegraphics[width=0.23\textwidth]{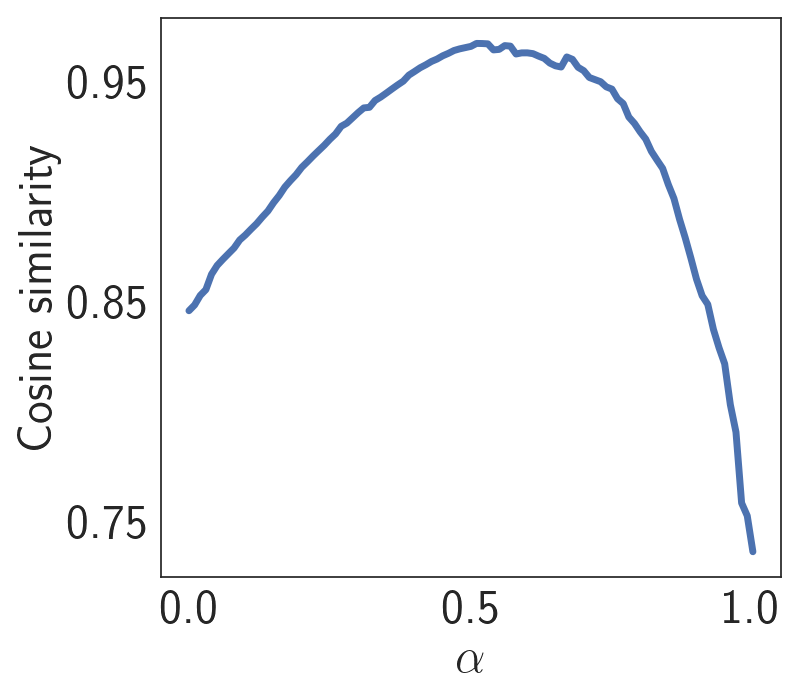}}%
    \subfloat[\centering $E$=10, $B$=10, $R$=20]{\includegraphics[width=0.23\textwidth]{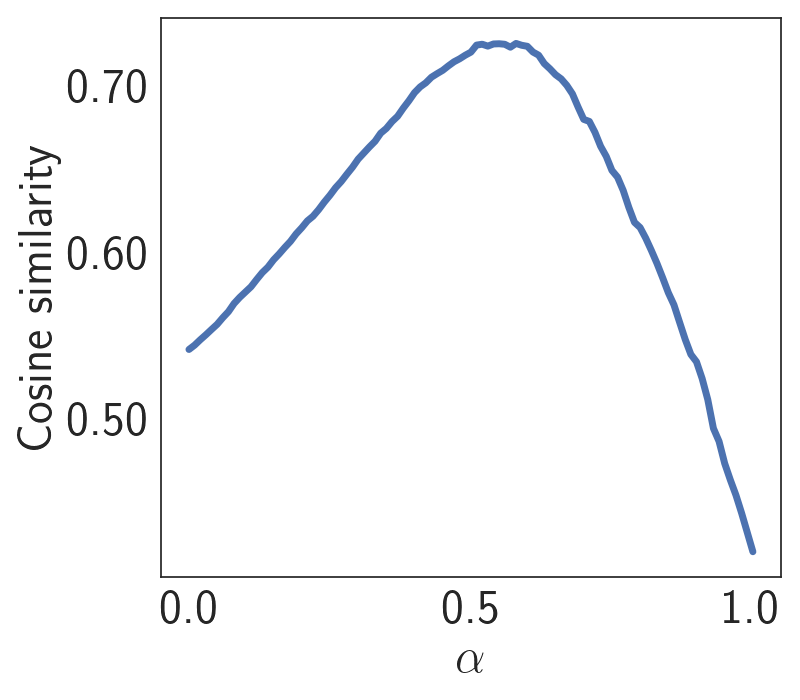}}%
    \caption{Cosine similarity between $\bm w_0 - \bm w_T$ and $\nabla\widehat \w$ vs.\ $\alpha$. The settings are aligned with \Cref{fig:g2_trained}, please refer to the caption above.}
    \label{fig:cos_trained}
\end{figure*}

Next, we investigate the influence of the network's training state. During FL, the low-rank property of the gradients may change. Additionally, the loss reduction $\ell(\w_0) - \ell(\w_T')$ is expected to be smaller as $\ell(\w_0)$ decreases. To study this, we train a network with FL as described in \Cref{sec:app:experiment}, and conduct attacks in a challenging setting with 50 steps of SGD. 
\Cref{fig:g2_trained} shows that as the communication round increases, the gradient norm ratio decreases, i.e.\ $G_2$ becomes larger. \Cref{fig:cos_trained} shows that in the middle of the connected line, the surrogate gradient still has a higher cosine similarity. However, the highest cosine similarity that can be reached becomes smaller. {Statistical evaluations of multiple samplings are presented in \Cref{tab:stats_trained}.}
We also provide the reconstruction performance on trained networks in \Cref{sec:app:trained_network}.
\begin{table*}[h!]
\begin{center}
\begin{small}
\begin{tabular}{ccccc}
\toprule
     &(E=10, B=10, R=1) 
&(E=10, B=10, R=5)  &(E=10, B=10, R=10)& (E=10, B=10, R=20)\\
\hhline{~----}
$\maE [\min_t \frac{\|P_2\nabla \bm w_t\|}{\|\bm w_t\|}]$
&.977$\pm$.008 &.937$\pm$.036 &.813$\pm$.087 &.406$\pm$.176\\
\midrule
$\maE[\Cosim|\alpha=0]$
&.872$\pm$.004 &.871$\pm$.044 &.764$\pm$.061 &.542$\pm$.074\\
\midrule
$\maE[\max_{\alpha\in[0,1]}\Cosim]$
&.979$\pm$.007 &.962$\pm$.019 &.902$\pm$.052 &.602$\pm$.010\\
\bottomrule
\end{tabular}
\end{small}
\caption{{Statistical evaluations corresponding to \Cref{fig:g2_trained,fig:cos_trained}. We repeat the experiment 100 times to compute the mean and standard deviation. $\Cosim$ is a shorthand for the cosine similarity between $\w_0 - \w_T$ and $\nabla \hw$}.}
\label{tab:stats_trained}
\end{center}
\end{table*}

\clearpage
\newpage
\section{Reconstruction Performance on Trained Networks}
\label{sec:app:trained_network}
During FL, the low-rank property of gradients may be changed, also the loss reduction of local training could be narrowed. Both of these could have a negative impact on the performance of \verb|SME| according to our \Cref{thm:GD} and \Cref{thm:SGD}. The empirical evidence provided in \Cref{sec:app:evidence} reflects this phenomenon. As the number of communication rounds grows, the gradients are less concentrated in a 2D subspace (see \Cref{fig:g2_trained}), and the highest cosine similarity achieved by the surrogate model also becomes smaller (see \Cref{fig:cos_trained}).

In this section, we present the results of reconstruction performance w.r.t.\ the influence of the training state of $\w_0$. We give the details of FL in \Cref{sec:app:experiment}. The attack results are shown in \Cref{tab:trained_psnr}. As we can see, for larger communication round $R$, the difference of PSNR between \verb|SME| and \verb|IG| becomes smaller, but we note that \verb|SME| still consistently outperforms \verb|IG|. \verb|DLFA| performs better than \verb|SME| when $R$ is large. However, we emphasize that \verb|DLFA| generally has a significantly higher demand for computational resources as discussed in \Cref{sec:experiments:computation}, it may not be able to execute \verb|DLFA| under some circumstances.

\begin{figure*}[h]
\begin{center}
    \subfloat[\centering $E$=10, $N$=50, $R$=1]{\includegraphics[width=0.95\textwidth]{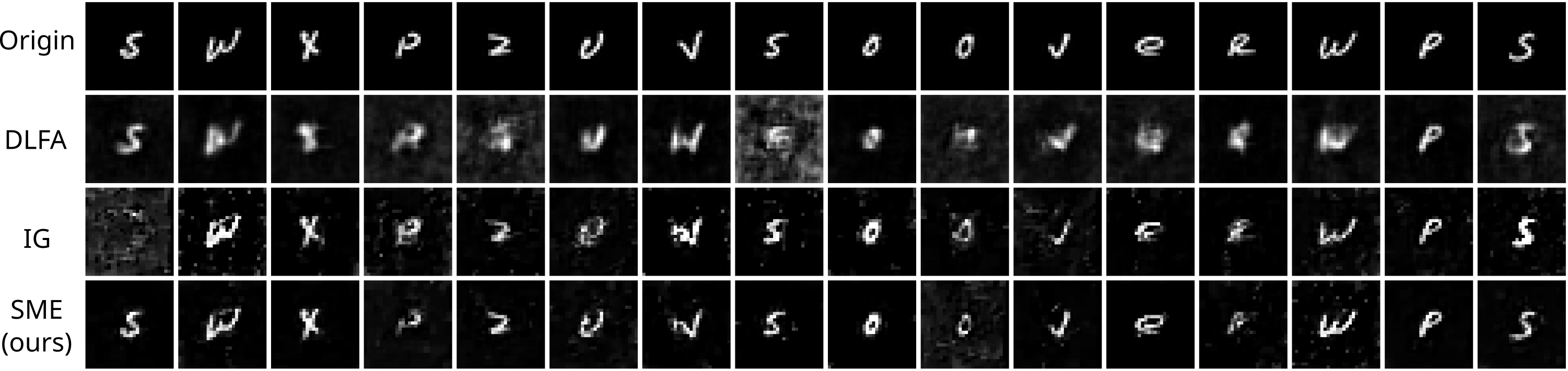}}%

    \subfloat[\centering $E$=10, $N$=50, $R$=5]{\includegraphics[width=0.95\textwidth]{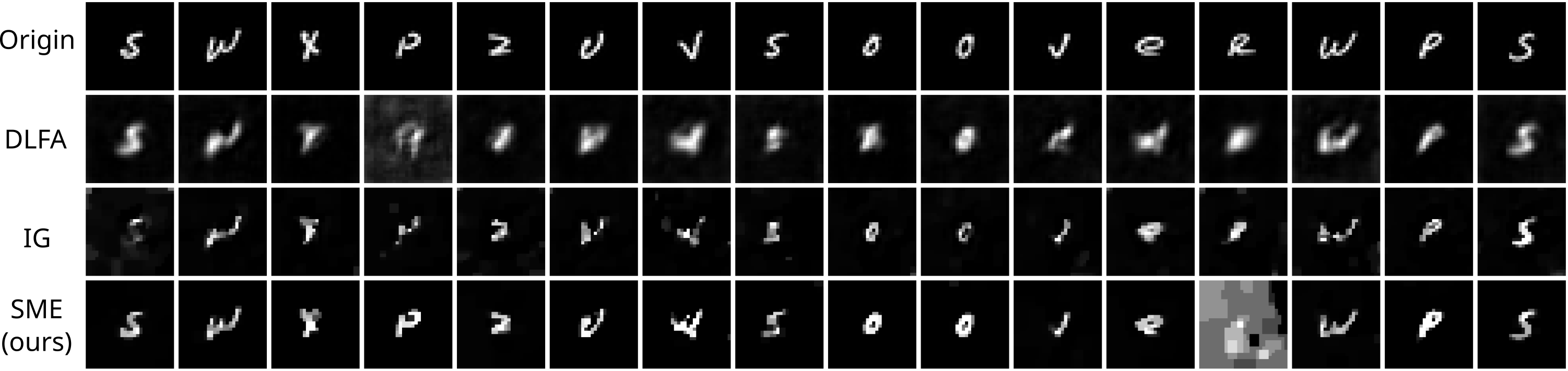}}%

    \subfloat[\centering $E$=10, $N$=50, $R$=10]{\includegraphics[width=0.95\textwidth]{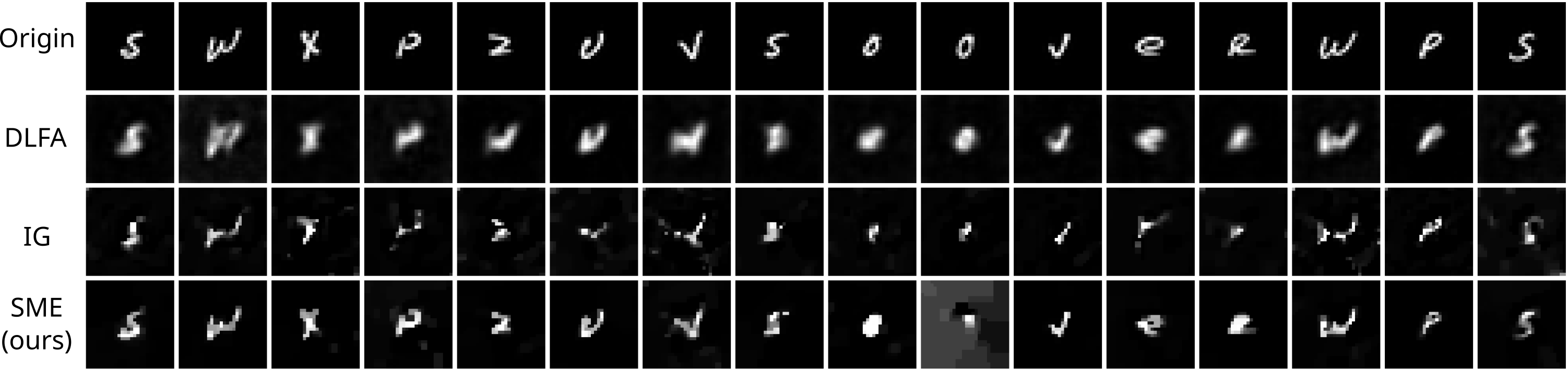}}%

    \subfloat[\centering $E$=10, $N$=50, $R$=20]{\includegraphics[width=0.95\textwidth]{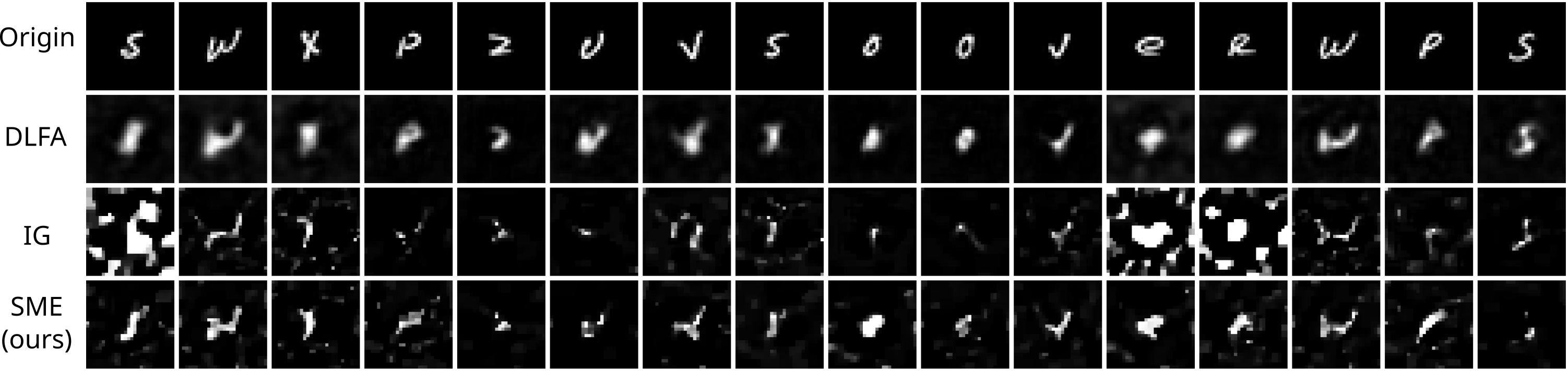}}%

\caption{Visualization of the reconstructed images. The results are drawn from the settings ($E=10$, $N=50$, $R=\{1, 5, 10, 20\}$) in \Cref{tab:trained_psnr}. The reconstructed images are paired with the original images through \textit{linear sum assignment}. We randomly sample 16 out of 50 images of one reconstruction.}
\label{fig:visual_trained}
\end{center}
\end{figure*}

We also observe that the reconstruction performance of all attack approaches becomes worse as the number of communication rounds $R$ grows. Seemingly, clients joining FL after multiple rounds of communication would have better privacy. However, this also leads to less contribution to the collaboration, which is an issue that needs to be considered in FL. We leave the study of the trade-off between privacy and collaboration contribution for future work. Additionally, the results indicate that using a pre-trained network in FL may be preferable not only in terms of network performance but also privacy. We also provide a visualization of the reconstruction in \Cref{fig:visual_trained}.
\begin{table*}[h]
\begin{center}
\begin{small}
\begin{tabular}{cccchahaha|d}
\toprule
     & & & &\multicolumn{2}{c}{DLFA} 
&\multicolumn{2}{c}{IG}  &\multicolumn{2}{c}{SME (ours)}&\multicolumn{1}{c}{}\\
\hhline{~~~~--||--||---}
$R$  &$E$ &$N$ &$T$ &$\mathcal{L}_{sim}\downarrow$     &PSNR $\uparrow$     &$\mathcal{L}_{sim}\downarrow$     &PSNR $\uparrow$ &$\mathcal{L}_{sim}\downarrow$     &PSNR $\uparrow$&$\bm \Delta$ PSNR\\
\midrule
\multirow{4}{*}{1} & 10 &10 &10 &$.021\pm.001$ &$24.9\pm0.2$ &$.044\pm.002$ &$27.8\pm0.3$ &$.019\pm.001$ &\bftab 30.3 $\pm$ \bftab0.3& $+2.5$\\
& 20 &10 &20 &$.019\pm.001$ &$25.4\pm0.2$ &$.090\pm.003$ &$24.2\pm0.3$ &$.019\pm.001$ &\bftab 28.6 $\pm$ \bftab0.3& $+3.2$\\
& 10 &50 &50 &$.015\pm.001$ &$21.7\pm0.1$ &$.091\pm.003$ &$19.1\pm0.2$ &$.027\pm.001$ &\bftab 22.2 $\pm$ \bftab 0.2&$+0.5$\\
& 20 &50 &100 &$.014\pm.001$ &\bftab 21.7 $\pm$ \bftab 0.2 &$.176\pm.011$ &$17.0\pm0.2$ &$.037\pm.001$ &$21.5\pm0.2$&$-0.2$\\
\midrule
\multirow{4}{*}{5} & 10 &10 &10 &$.003\pm.000$ &$22.1\pm0.2$ &$.026\pm.001$ &$22.7\pm0.3$ &$.009\pm.001$ &\bftab 24.8 $\pm$ \bftab0.3& $+2.7$\\
& 20 &10 &20 &$.003\pm.000$ &$23.0\pm0.1$ &$.064\pm.003$ &$20.9\pm0.2$ &$.018\pm.001$ &\bftab 24.7 $\pm$ \bftab0.3& $+1.7$\\
& 10 &50 &50 &$.005\pm.000$ &$20.8\pm0.2$ &$.062\pm.004$ &$18.7\pm0.2$ &$.020\pm.002$ &\bftab 21.4 $\pm$ \bftab 0.3&$+0.6$\\
& 20 &50 &100 &$.005\pm.000$ &\bftab 20.9 $\pm$ \bftab 0.2 &$.131\pm.007$ &$17.2\pm0.2$ &$.042\pm.003$ &$20.3\pm0.3$&$-0.6$\\
\midrule
\multirow{4}{*}{10} & 10 &10 &10 &$.003\pm.000$ &$22.5\pm0.1$ &$.034\pm.001$ &$21.4\pm0.3$ &$.011\pm.001$ &\bftab 23.8 $\pm$ \bftab0.3& $+1.3$\\
& 20 &10 &20 &$.004\pm.000$ &$23.1\pm0.2$ &$.079\pm.003$ &$19.5\pm0.2$ &$.019\pm.001$ &\bftab 23.2 $\pm$ \bftab0.4& $+0.1$\\
& 10 &50 &50 &$.006\pm.000$ &\bftab 20.7 $\pm$ \bftab 0.2 &$.107\pm.016$ &$17.7\pm0.2$ &$.027\pm.001$ &$19.8\pm0.3$&$-0.9$\\
& 20 &50 &100 &$.008\pm.000$ &\bftab 20.5 $\pm$ \bftab 0.2 &$.298\pm.035$ &$16.7\pm0.2$ &$.060\pm.004$ &$18.4\pm0.3$&$-2.1$\\
\midrule
\multirow{4}{*}{20} & 10 &10 &10 &$.008\pm.001$ &\bftab 21.3 $\pm$ \bftab 0.2 &$.053\pm.002$ &$19.0\pm0.2$ &$.019\pm.001$ &$20.7\pm0.3$& $-0.6$\\
& 20 &10 &20 &$.013\pm.001$ &\bftab 20.8 $\pm$ 0.2 &$.109\pm.013$ &$17.9\pm0.2$ &$.044\pm.003$ &$19.6\pm0.3$& $-1.2$\\
& 10 &50 &50 &$.020\pm.001$ &\bftab 19.4 $\pm$ 0.2 &$.208\pm.003$ &$16.2\pm0.2$ &$.152\pm.030$ &$17.3\pm0.2$&$-2.1$\\
& 20 &50 &100 &$.020\pm.001$ &\bftab 19.0 $\pm$ \bftab 0.2 &$.374\pm.043$ &$15.4\pm0.2$ &$.306\pm.043$ &$16.2\pm0.2$&$-2.8$\\
\bottomrule
\end{tabular}
\end{small}
\caption{Average reconstructed image quality measured by PSNR and similarity loss of the reconstruction objective $\mathcal{L}_{sim}$ on FEMNIST. We set the batch size $B=10$ and consider different communication round $R$, local data sizes $N$, and epochs $E$. Local steps $T=E \times (N / B)$. The best reconstruction results are bold. The difference of PSNRs between SME and the best baseline is given in the last column. We restate part of the results in \Cref{tab:psnr} for $R=1$. Also refer to \Cref{fig:visual_trained} for visualization.
}
\label{tab:trained_psnr}
\end{center}
\end{table*}

\clearpage
\newpage
\section{{Reconstruction Performance on other Network Architectures}}
\label{sec:app:other_network}
We provide the reconstruction performance results on other network architectures. In particular, we consider: (1) A MLP with two hidden layers of shape $(1000, 1000)$ equipped with ReLU activation function. (2) A LeNet~\cite{LeNet} equipped with Tanh activation function. (3) ResNet~\citep{resnet}, a convolutional neural network with skip connections. In particular, we adopt a relatively small ResNet8 with three building blocks in favor of \verb|DLFA| and follow \citet{geiping_inverting_gradient} to freeze the running statistics of the batch normalization layers during training. Updating the running statistics might disable reconstruction as reported by \citet{huang2021evaluating}. (4) ViT~\citep{vit}, a non-convolutional network with multi-head attention building blocks and layer normalization. We follow \citet{april_lu_cvpr} and adopt the architecture of type A. We set the learning rate in FL to 0.4 for LeNet and 0.004 for the other architectures. All attack approaches adopt the hyperparameters described in \Cref{sec:app:experiment}. The results are given in \Cref{tab:network_psnr}. 

As we can see, in general \verb|SME| achieves the best performance in terms of reconstruction. \verb|DLFA|, on the other hand, fails in the attacks of LeNet and ResNet8.
We also notice that \verb|DLFA| works when the learning rate of LeNet is lower. However, when we increase the learning rate to
0.4, the reconstruction of \verb|DLFA| converges in the first few iterations and then diverges dramatically (we record the best
PSNR during reconstruction). This may indicate that \verb|DLFA| is prone to failure or sensitive to configuration, while the gradient inversion methods are more robust on different architectures.
\begin{table*}[h]
\begin{center}
\begin{small}
\begin{tabular}{cccchahaha|d}
\toprule
     & & & &\multicolumn{2}{c}{DLFA} 
&\multicolumn{2}{c}{IG}  &\multicolumn{2}{c}{SME (ours)}&\multicolumn{1}{c}{}\\
\hhline{~~~~--||--||---}
Network  &$E$ &$N$ &$T$ &$\mathcal{L}_{sim}\downarrow$     &PSNR $\uparrow$     &$\mathcal{L}_{sim}\downarrow$     &PSNR $\uparrow$ &$\mathcal{L}_{sim}\downarrow$     &PSNR $\uparrow$&$\bm \Delta$ PSNR\\
\midrule
\multirow{6}{*}{MLP} & 10 &10 &10 &$.018\pm.000$  &\bftab 36.0 $\pm$ 0.3 &$.043\pm.006$ &$31.7\pm0.3$ &$.009\pm.000$ &$35.9\pm0.4$& $-0.1$\\
& 20 &10 &20 &$.025\pm.000$  & $31.3\pm0.2$&$.041\pm.002$ &$29.5\pm0.3$ &$.013\pm.000$ &\bftab 34.1 $\pm$ \bftab0.4& $+2.8$\\
& 50 &10 &50 &$.051\pm.001$  & $28.0\pm0.2$&$.097\pm.002$ &$25.1\pm0.3$ &$.023\pm.001$ &\bftab 30.1 $\pm$ \bftab0.3& $+2.1$\\
& 10 &50 &50 &$.030\pm.001$  & $27.0\pm0.2$ &$.056\pm.002$ &$25.1\pm0.3$ &$.009\pm.000$ &\bftab 31.4 $\pm$ \bftab 0.3&$+4.4$\\
& 20 &50 &100&$.039\pm.001$  &$24.6\pm0.2$ &$.090\pm.002$ &$23.3\pm0.3$ &$.017\pm.001$ &\bftab 28.5 $\pm$ \bftab 0.3&$+3.9$\\
& 50 &50 &250&$.058\pm.002$  &$22.0\pm0.2$ &$.173\pm.003$ &$20.6\pm0.2$ &$.054\pm.001$ &\bftab 22.8 $\pm$ \bftab 0.3&$+0.8$\\
\midrule
\multirow{6}{*}{LeNet} & 10 &10 &10 &$.315\pm.022$  &$16.3\pm0.1$ &$.000\pm.000$ &$29.1\pm0.3$ &$.000\pm.000$ &\bftab 29.5 $\pm$ \bftab0.3& $+0.4$\\
& 20 &10 &20 &$.555\pm.016$  &$16.0\pm0.1$ &$.002\pm.000$ &$27.8\pm0.3$ &$.000\pm.000$ &\bftab 29.4 $\pm$ \bftab0.3& $+1.6$\\
& 50 &10 &50 &$.690\pm.040$  &$16.1\pm0.1$ &$.015\pm.002$ &$23.0\pm0.3$ &$.000\pm.000$ &\bftab 29.0 $\pm$ \bftab0.3& $+6.0$\\
& 10 &50 &50 &$.663\pm.012$  &$15.9\pm0.1$ &$.003\pm.000$ &$23.0\pm0.4$ &$.000\pm.000$ &\bftab 24.6 $\pm$ \bftab 0.3&$+1.6$\\
& 20 &50 &100 &$.553\pm.023$  &$16.1\pm0.1$ &$.017\pm.002$ &$19.8\pm0.4$ &$.000\pm.000$ &\bftab 24.2 $\pm$ \bftab 0.3&$+4.4$\\
& 50 &50 &250 &$.690\pm.040$  &$16.4\pm0.2$ &$.159\pm.010$ &$15.0\pm0.2$ &$.019\pm.002$ &\bftab 19.4 $\pm$ \bftab 0.5&$+3.0$\\
\midrule
\multirow{6}{*}{ResNet8} & 10 &10 &10 &$.148\pm.009$  &$16.5\pm0.1$ &$.004\pm.000$ &$32.7\pm0.5$ &$.001\pm.000$ &\bftab 36.8 $\pm$ \bftab0.5& $+4.1$\\
 & 20 &10 &20 &$.163\pm.011$  &$16.3\pm0.1$ &$.011\pm.000$ &$26.7\pm0.4$ &$.001\pm.000$ &\bftab 36.6 $\pm$ \bftab0.5& $+9.9$\\
  & 50 &10 &50 &$.125\pm.010$  &$16.2\pm0.2$ &$.059\pm.003$ &$18.4\pm0.3$ &$.001\pm.000$ &\bftab 33.2 $\pm$ \bftab0.5& $+14.8$\\
& 10 &50 &50 &$.114\pm.001$  &$15.9\pm0.1$ &$.015\pm.001$ &$21.2\pm0.5$ &$.002\pm.000$ &\bftab 28.2 $\pm$ \bftab0.5& $+7.0$\\
& 20 &50 &100 &$.116\pm.005$  &$16.1\pm0.2$ &$.061\pm.004$ &$17.1\pm0.4$ &$.002\pm.000$ &\bftab 27.3 $\pm$ \bftab0.6& $+10.2$\\
& 50 &50 &250 &$.206\pm.007$  &$16.2\pm0.2$ &$.163\pm.006$ &$14.0\pm0.1$ &$.028\pm.003$ &\bftab 20.8 $\pm$ \bftab0.6& $+4.6$\\
\midrule
\multirow{6}{*}{ViT} & 10 &10 &10 &$.008\pm.001$  &$25.6\pm0.5$ &$.037\pm.001$ &$25.1\pm0.2$ &$.012\pm.000$ &\bftab 30.3 $\pm$ \bftab0.2& $+4.7$\\
& 20 &10 &20 &$.007\pm.000$  &$21.1\pm0.2$ &$.079\pm.001$ &$21.1\pm0.1$ &$.028\pm.000$ &\bftab 26.4 $\pm$ \bftab0.2& $+5.3$\\
& 50 &10 &50 &$.013\pm.004$  &$18.9 \pm 0.1$ &$.138\pm.001$ &$17.7\pm0.1$ &$.053\pm.001$ &\bftab 21.3 $\pm$ \bftab 0.1& $+1.4$\\
& 10 &50 &50 &$.017\pm.009$  &$20.6\pm0.2$ &$.067\pm.002$ &$19.1\pm0.2$ &$.022\pm.001$ &\bftab 23.3 $\pm$ \bftab0.3& $+2.7$\\
& 20 &50 &100 &$.015\pm.004$  &$18.3\pm0.1$ &$.114\pm.002$ &$16.8\pm0.1$ &$.044\pm.001$ &\bftab 20.0 $\pm$ \bftab0.1& $+1.7$\\
& 50 &50 &250 &N/A  &N/A &$.157\pm.001$ &$14.9\pm0.1$ &$.070\pm.002$ &\bftab 17.2 $\pm$ \bftab0.1& N/A\\
\bottomrule
\end{tabular}
\end{small}
\caption{Average reconstructed image quality measured by PSNR and similarity loss of the reconstruction objective $\mathcal{L}_{sim}$ on FEMNIST. We set the batch size $B=10$ and change the local data sizes $N$ and epochs $E$. Local steps $T=E \times (N / B)$. The experiments are conducted at the first communication round. The best reconstruction results are bold. The difference of PSNRs between SME and the best baseline is given in the last column. Also refer to \Cref{fig:visual_network} for visualization.
Results of DLFA in the last row is not available, as \textit{JAX} raises unknown compilation error.
}
\label{tab:network_psnr}
\end{center}
\end{table*}
\clearpage
\newpage
\section{Reconstruction Visualization}
\label{sec:app:visual}
\begin{figure*}[h!]
    \centering
    \subfloat[\centering $E$=20, $B$=10, $T$=20]{\includegraphics[width=0.59\textwidth]{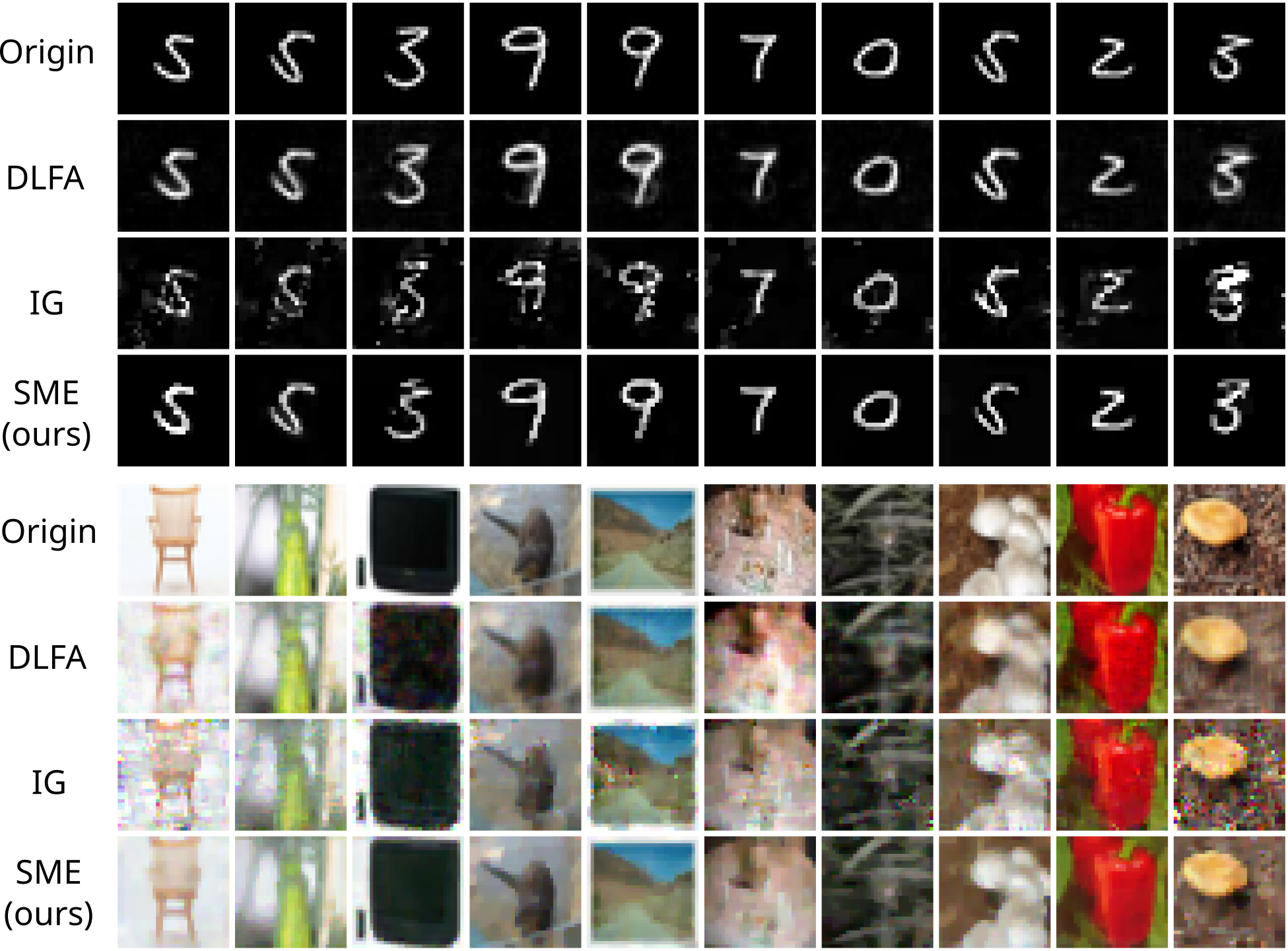}}%

    \subfloat[\centering $E$=50, $B$=10, $T$=50]
    {\includegraphics[width=0.59\textwidth]{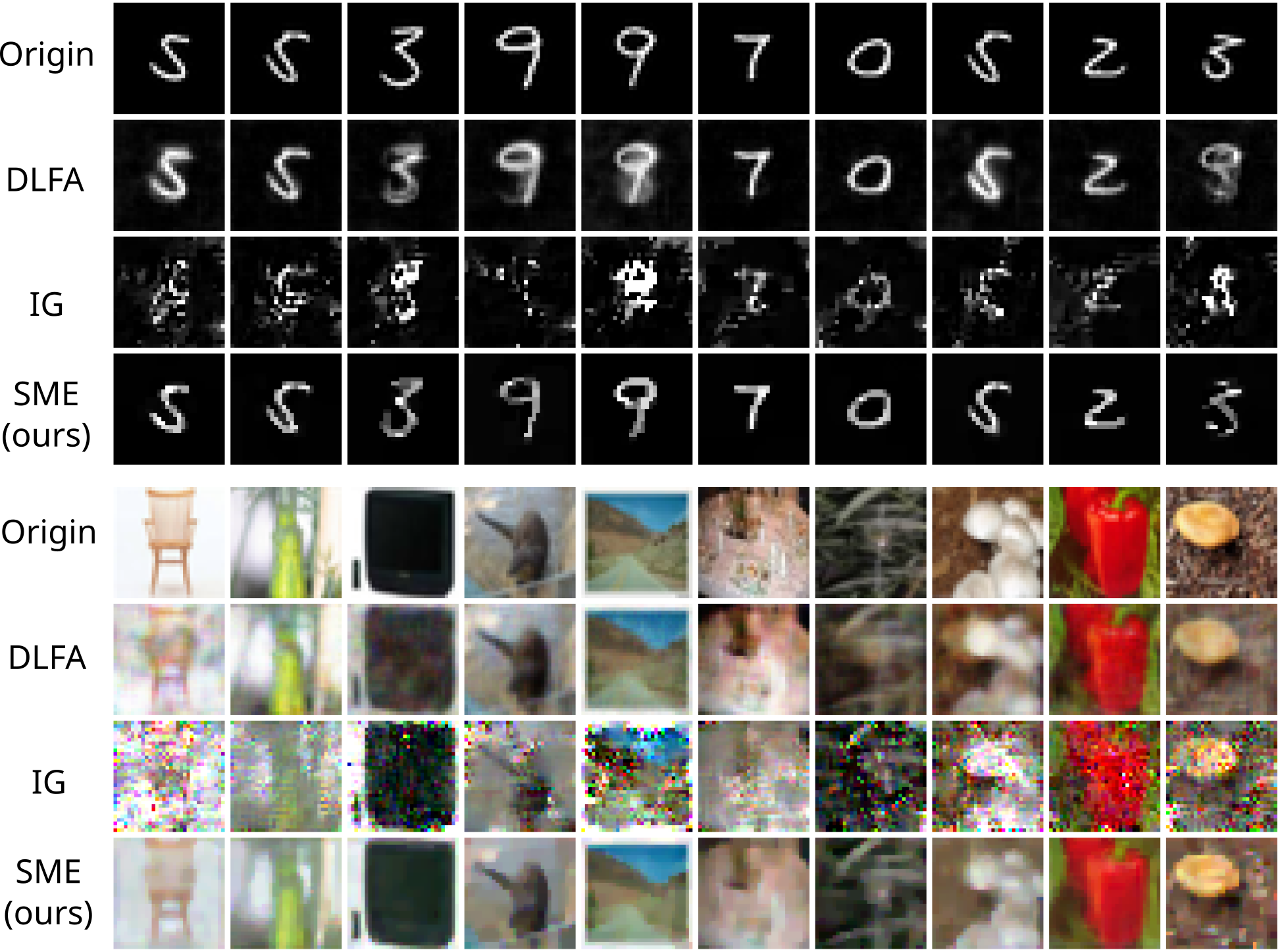}}%
    \caption{Visualization of the reconstructed images. The results belong to one reconstruction of the settings ($E\in\{20,50\}$, $N=10$) in \Cref{tab:psnr}. The reconstructed images are paired with the original images through \textit{linear sum assignment}.}
    \label{fig:visual_10k}
\end{figure*}

\begin{figure*}[h]
\begin{center}
    \subfloat[\centering $E$=10, $B$=10, $T$=50]{\includegraphics[width=0.95\textwidth]{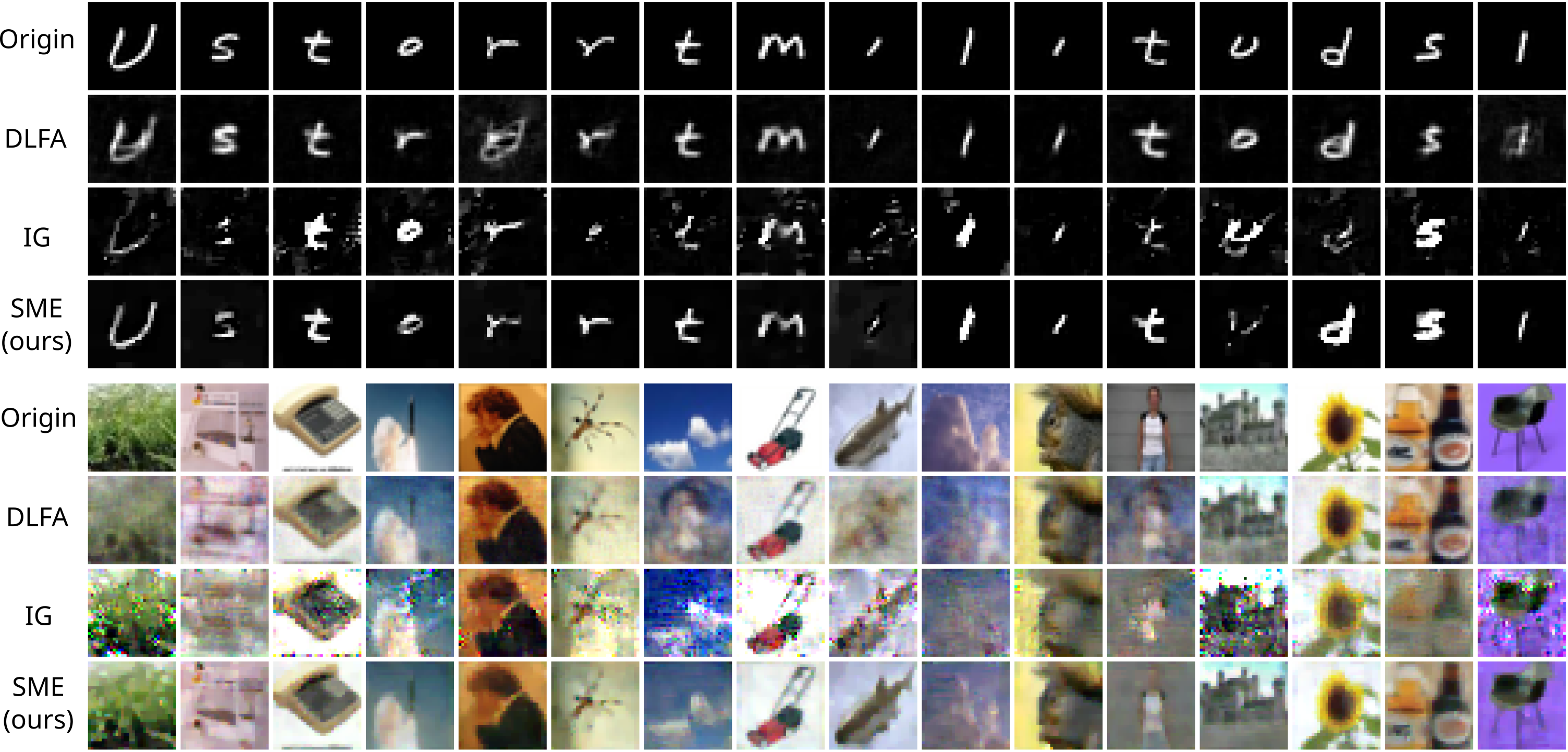}}%

    \subfloat[\centering $E$=20, $B$=10, $T$=100]
    {\includegraphics[width=0.95\textwidth]{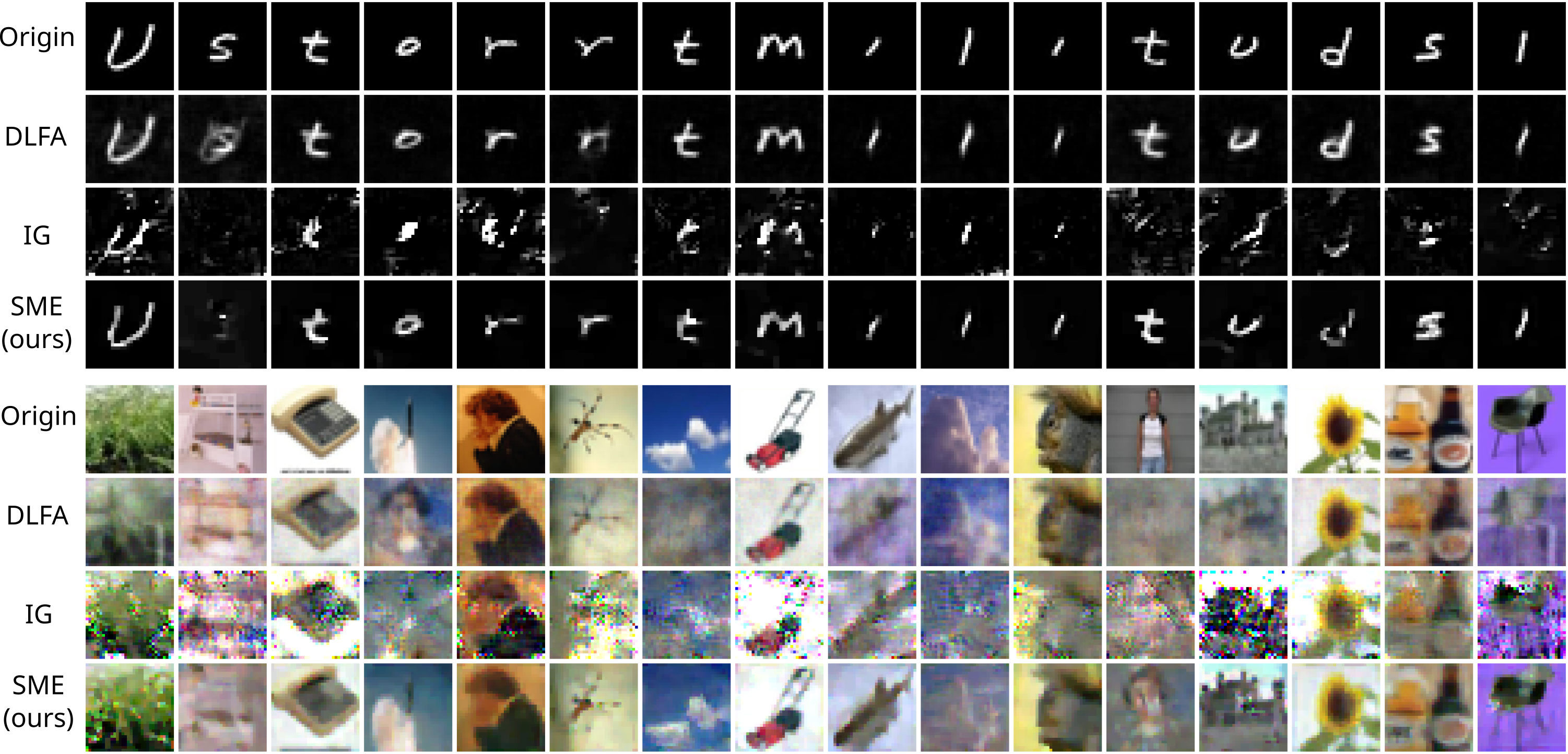}}%
\caption{Visualization of the reconstructed images. The results are drawn from the settings ($E\in\{10, 20\}$, $N=50$) in \Cref{tab:psnr}. The reconstructed images are paired with the original images through \textit{linear sum assignment}. We randomly sample 16 out of 50 images of one reconstruction.}
\label{fig:visual_10e10b50k}
\end{center}
\end{figure*}

\begin{figure*}[h]
\begin{center}
    \subfloat[\centering MLP, $E$=20, $N$=50, $T=100$]{\includegraphics[width=0.95\textwidth]{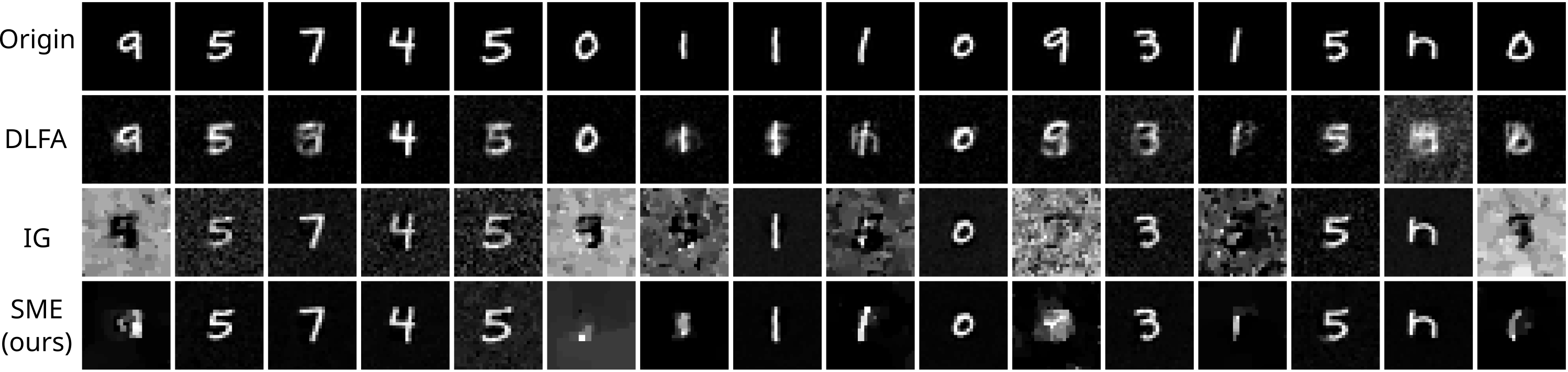}}%

    \subfloat[\centering LeNet, $E$=20, $N$=50, $T=100$]{\includegraphics[width=0.95\textwidth]{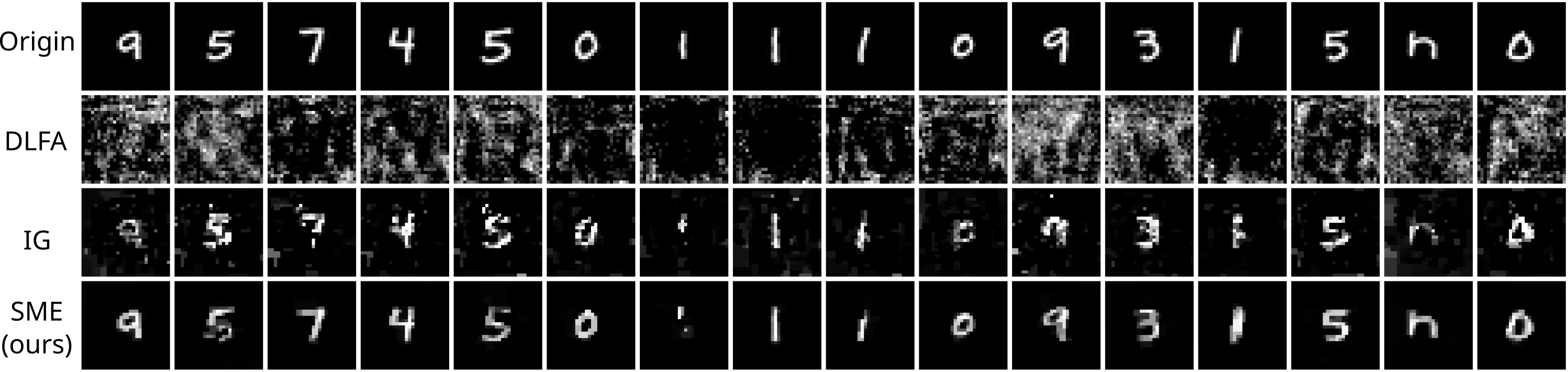}}%

    \subfloat[\centering ResNet8, $E$=20, $N$=50, $T=100$]{\includegraphics[width=0.95\textwidth]{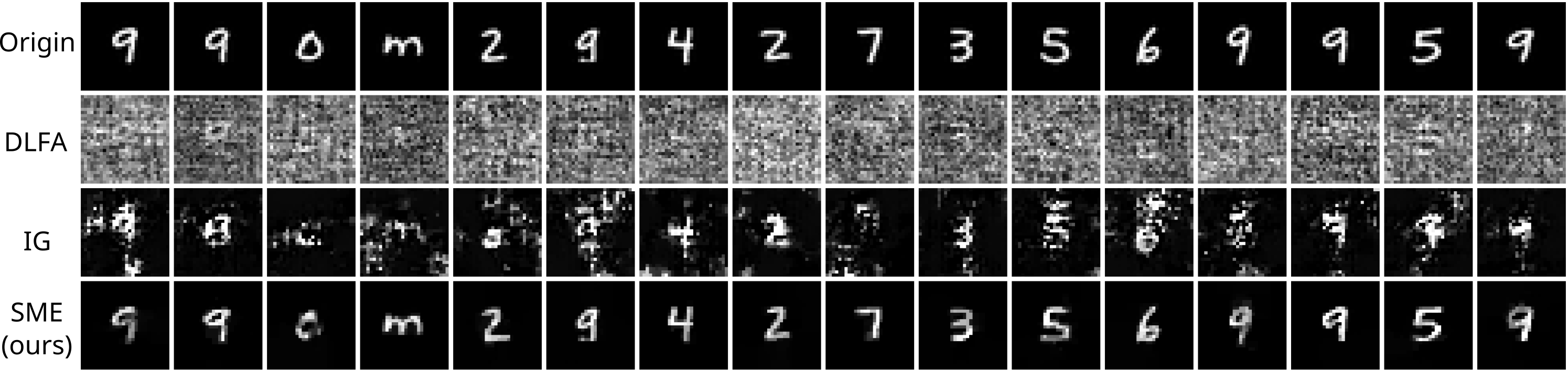}}%

    \subfloat[\centering ViT, $E$=20, $N$=50, $T=100$]{\includegraphics[width=0.95\textwidth]{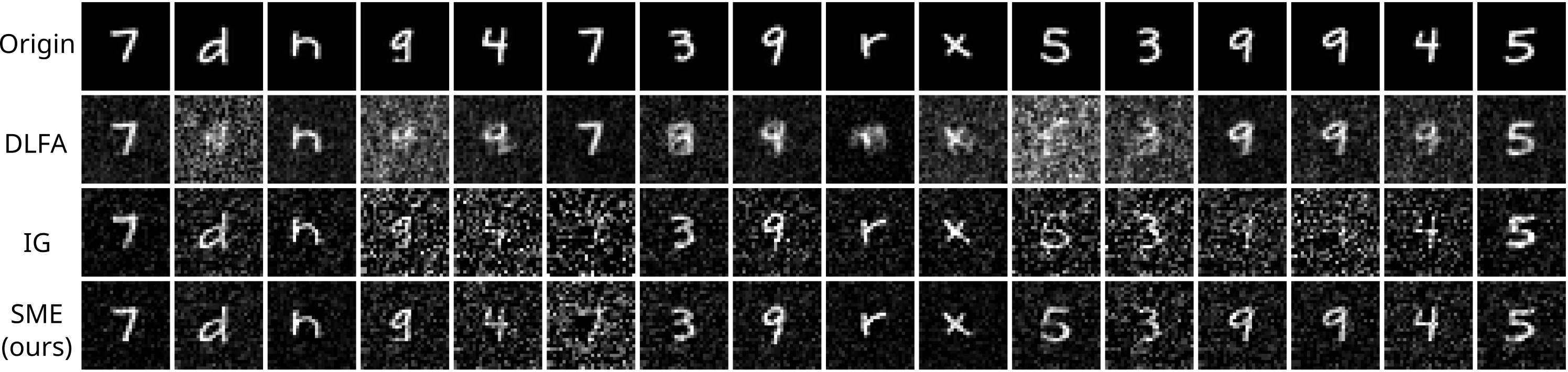}}%
\caption{Visualization of the reconstructed images. The results are drawn from the setting ($E=20$, $N=50$, $T=100$) on the network architectures MLP, LeNet, ResNet8 and ViT in \Cref{tab:network_psnr}. The reconstructed images are paired with the original images through \textit{linear sum assignment}. We randomly sample 16 out of 50 images of one reconstruction.}
\label{fig:visual_network}
\end{center}
\end{figure*}

\end{document}